\documentclass[english,listtotoc, chapterprefix, tablecaptionabove, bibtotoc, parskip=full, headings=small, numbers=noenddot]{article}
\usepackage[T1]{fontenc}
\usepackage[utf8]{inputenc}
\usepackage[a4paper]{geometry}
\usepackage{showlabels}
\usepackage{color}
\usepackage{babel}
\usepackage{float}
\usepackage{amsmath}
\usepackage{amssymb}
\usepackage{amsfonts}
\usepackage{amsthm}
\usepackage{graphicx}
\usepackage{esint}
\usepackage{bm}
\usepackage{hyperref}
\usepackage{breakcites}
\hypersetup{
 pdfauthor={Miguel},
 linkcolor=blue, citecolor=red, urlcolor=blue, pdfstartview=XYZ, plainpages=false, pdfpagelabels}

\newcommand{\1}{\mathbb I}

\def\beq{\begin{equation}}
\def\eeq{\end{equation}}
\def\barray{\begin{eqnarray}}
\def\earray{\end{eqnarray}}
\def\<{\langle}
\def\>{\rangle}
\newcommand{\matriz}[4]{\left(\begin{array}{cc} #1 & #2 \\ #3 & #4 \end{array}\right)}

\def\d{{\rm d}}

\def\-{\setminus}

\def\L{{\cal L}}

\def\T{\bm{\theta}}
\def\x{{\bf x}}

\def\h{{\bf h}}
\def\f{{\bf f}}

\def\sx{{\sf x}}
\def\sy{{\sf y}}
\def\c{{\sf c}}
\def\nl{{n}}
\def\Ns{{n_{\rm s}}}
\def\D{{\vec\Delta}}
\def\DD{{\bm\Delta}}
\def\r{{\vec r}}
\def\Jx{{J^{(\sx\sx)}}}
\def\Jy{{J^{(\sy\sy)}}}
\def\Jxy{{J^{(\sx\sy)}}}
\def\Cx{{C^{(\sx\sx)}}}
\def\Cy{{C^{(\sy\sy)}}}
\def\Cxy{{C^{(\sx\sy)}}}

\def\tP{{\tilde P}}

\makeatletter
\@ifundefined{lettrine}{\usepackage{lettrine}}{}
\numberwithin{equation}{section}
\numberwithin{figure}{section}
\numberwithin{table}{section}

\usepackage{multicol}
\usepackage{fancyhdr}
\usepackage{lettrine}
\usepackage[backend=biber,style=numeric,sorting=none,citestyle=numeric,maxcitenames=2]{biblatex}
\date{}

\makeatother

\bibliography{main_unformatted}

\title{Unsupervised inference approach to facial attractiveness}

\author{\small  Miguel Ib{\'a}{\~n}ez-Berganza$^{1,*}$, Ambra Amico$^{2}$, Gian Luca Lancia$^{1}$, \\ \small Federico Maggiore$^1$, Bernardo Monechi$^3$, Vittorio Loreto$^{1,3,4}$ \\ 
\small $^1$ Sapienza University of Rome, Physics Department, Piazzale Aldo Moro 2, 00185 Rome, Italy. \\ 
\small $^2$ ETH Zurich, Chair of Systems Design, WEV G 212 Weinbergstrasse 56/58, 8092 Zurich. \\ 
\small $^3$ Sony Computer Science Laboratories, Paris, 6, rue Amyot, 75005, Paris, France. \\ 
\small $^4$ Complexity Science Hub, Josefst\"adter Strasse 39, A 1080 Vienna, Austria. \\
{\small $^*${\sf  miguel.berganza@roma1.infn.it }}
} %
\begin{document}

\maketitle
\tableofcontents 

\begin{abstract}
	
The perception of facial beauty is a complex phenomenon depending on many, detailed and global facial features influencing each other. In the machine learning community this problem is typically tackled as a problem of supervised inference. However, it has been conjectured that this approach does not capture the complexity of the phenomenon. A recent original experiment (Ib{\'a}{\~n}ez-Berganza et al., Scientific Reports {\textbf 9}, 8364, 2019) allowed different human subjects to navigate the face-space and ``sculpt'' their preferred modification of a reference facial portrait. Here we present an unsupervised inference study of the set of sculpted facial vectors in that experiment. We first infer minimal, interpretable, and faithful probabilistic models (through Maximum Entropy and artificial neural networks) of the preferred facial variations, that capture the origin of the observed inter-subject diversity in the sculpted faces. The application of such generative models to the supervised classification of the gender of the sculpting subjects, reveals an astonishingly high prediction accuracy. This result suggests that much relevant information regarding the subjects may influence (and be elicited from) her/his facial preference criteria, in agreement with the multiple motive theory of attractiveness proposed in previous works.

\end{abstract}



\section{Introduction}

Human facial perception (of identity, emotions, personality dimensions, attractiveness \cite{walker2016,little2011,leopold2010}) has been the subject of an intense and multidisciplinary research in the last decades. In particular, facial attractiveness is a research topic in many different disciplines, from evolutionary biology and psychology to neuroscience \cite{bzdok2011,hahn2014,laurentini2014,little2014,thornhill1999}. Furthermore, it is an interesting case of study in the machine learning research community, as a paradigm of a complex cognitive phenomenon, ruled by complex and difficult to infer criteria. Indeed, the rules according to which a facial image will probably result pleasant to an individual or in average, are poorly known \cite{little2011,laurentini2014}. The most relevant face-space variables in terms of which such rules should be inferred remain elusive as well \cite{laurentini2014}. 

In the context of evolutionary biology, on the one hand, many works have discussed the validity of the so called {\it natural selection hypothesis} \cite{little2011}. 
Despite the success of the natural selection hypothesis, it is believed that it does not take into account the phenomenon in its various complex facets. Indeed, important cultural and inter-person differences, beyond the species-typical criterion, are known to influence facial attractiveness \cite{little2014}. On the other hand, the main goal of the machine learning approach is the automatic rating of facial images, as a supervised inference problem \cite{laurentini2014}. The facial image is parametrised in a face-space vector $\f$, the inference goal consists in inferring the model $R(\f)$ that reproduces at best the subject-averaged ratings $\< R_s\>_s$ of a database $\{\f_s,R_s\}$.\footnote{In the case of deep, hierarchical networks, which automatically perform feature selection, the raw image is used as an input to the learning algorithm instead of a face-space parametrisation $\f$. The resulting relevant features are, however, not immediately accessible.}

From a methodological point of view, most of the works in facial attractiveness draw their conclusions from the average rating assigned to several natural facial images by a pool of subjects (although computer-modified facial images have also been used \cite{laurentini2014,ibanez2019}). Such a strategy may present important limitations. It has been argued that the analysis of average ratings assigned to natural faces may suffer, as an experimental technique, the curse of dimensionality and, consequently, it may hinder the complexity and subjectivity of the phenomenon \cite{laurentini2014,valentine2016,ibanez2019}. As an alternative measure to the average rating, it has been proposed the estimation of the single subject's {\it preferred region in (a subspace of) the face-space} \cite{ibanez2019}.\footnote{The alternative experimental technique allows a given subject to seek her/his preferred {variation of a reference facial portrait}. Such variations differ only in {a low-dimensional face-space of essential facial features}. It is arguably the introduction of these two ingredients: the reduction of facial degrees of freedom and the possibility to efficiently {\it explore} the face-space (rather than {\it rating} facial images differing in many facial dimensions) that allows for a significant experimental distinction of different subject's criteria.} 
 Within a sufficiently high precision, different subjects would systematically reveal distinguishable preferred regions in the face-(sub)space. Such a {\it complete subjectivity} picture is compatible with previous studies \cite{cunningham1995,edler2001,little2014,galantucci2014,oosterhof2008,todorov2011,walker2016,abir2017} arguing that complex psychological mechanisms influence the single subject preferences in the face-space (the {\it multiple motive hypothesis}).

According to this idea, the single subject preferred modifications, elicited with high accuracy in \cite{ibanez2019}, are expected to reflect relevant information regarding the subject. We here investigate this concept by means of an inference study of the set of facial variations sculpted by different subjects in \cite{ibanez2019}. In particular, we infer a probabilistic generative model, $\L({\f}|\T)$, from the database of sculpted facial vectors ${\cal S}=\{\f^{(s)}\}_{s=1}^S$ (where $s$ is the subject index). $\L({\f}|\T)$ represents the probability density of a facial image with face-space vector $\f$ to be sculpted by any subject (given the reference facial portrait and the sculpture protocol). We have considered three generative models of unsupervised learning: two Maximum Entropy (MaxEnt) models, with linear and non-linear interactions among the facial coordinates, and the Gaussian Restricted Boltzmann Machine (GRBM) model of Artificial Neural Network (ANN). 

The generative models account for the inter-subject fluctuations around the most probable facial vector. Such fluctuations are expected to reflect and encode meaningful differences among experimental subjects. 
To highlight this fact, we apply our models to the supervised classification of the facial variations according to the subject's gender. This allows to predict the gender of test subjects with at least $95\%$ of accuracy.

The models presented here are interpretable, as the model parameters $\T$ provide information regarding the relative importance of the various facial distances and their mutual influence in the cognitive process of face perception. These are fundamental questions in the specific litterature \cite{laurentini2014}. In particular, a comparison among the various models' efficiency highlights the relevance of the {\it nonlinear mutual influence} (hence beyond {\it proportions}, or pairwise influence) of facial distances. Finally, this work provides a novel case of study, in the field of cognitive science, for techniques and methods in unsupervised inference and, in particular, a further application of the MaxEnt method \cite{jaynes1957,berg2017,nguyen2017,demartino2018},  otherwise extensively used in physics, systems neuroscience and systems biology \cite{lezon2006,schneidman2006,shlens2006,bialek2007,tang2008,weigt2009,roudi2009,tkacik2009,stephens2010,mora2010,morcos2011,bialek2012,bialek2012}. 

The inference models of the data in reference \cite{ibanez2019}  will be first presented in sec. \ref{sec:methods}, along with some key methodological details (see the Supplementary Information (SI) document for in-depth methodological descriptions). In sec. \ref{sec:results} we will assess the quality of our models as generative model of the set of sculpted vectors. We will draw our conclusions in sec. \ref{sec:discussion}.

\section{Materials and methods}
\label{sec:methods}

We have considered of the dataset ${\cal S}$ described in \cite{ibanez2019}. In such experiments, each subject was allowed to sculpt her/his favorite deformation of a reference portrait (through the interaction with an software which combines image deformation techniques with a genetic algorithm for the efficient search in the face-space). The set of selected images are, hence, artificial, though realistic, {\it variations} of a common {\it reference portrait} (corresponding to a real person). In such a way, only the geometric positions of the landmarks are allowed to vary, the texture degrees of freedom are fixed (and correspond to the reference portrait RP1 in \cite{ibanez2019}, see fig. \ref{fig:key}). 

\begin{figure}[h!]                        
\begin{center} 
\includegraphics[width=0.425\columnwidth]{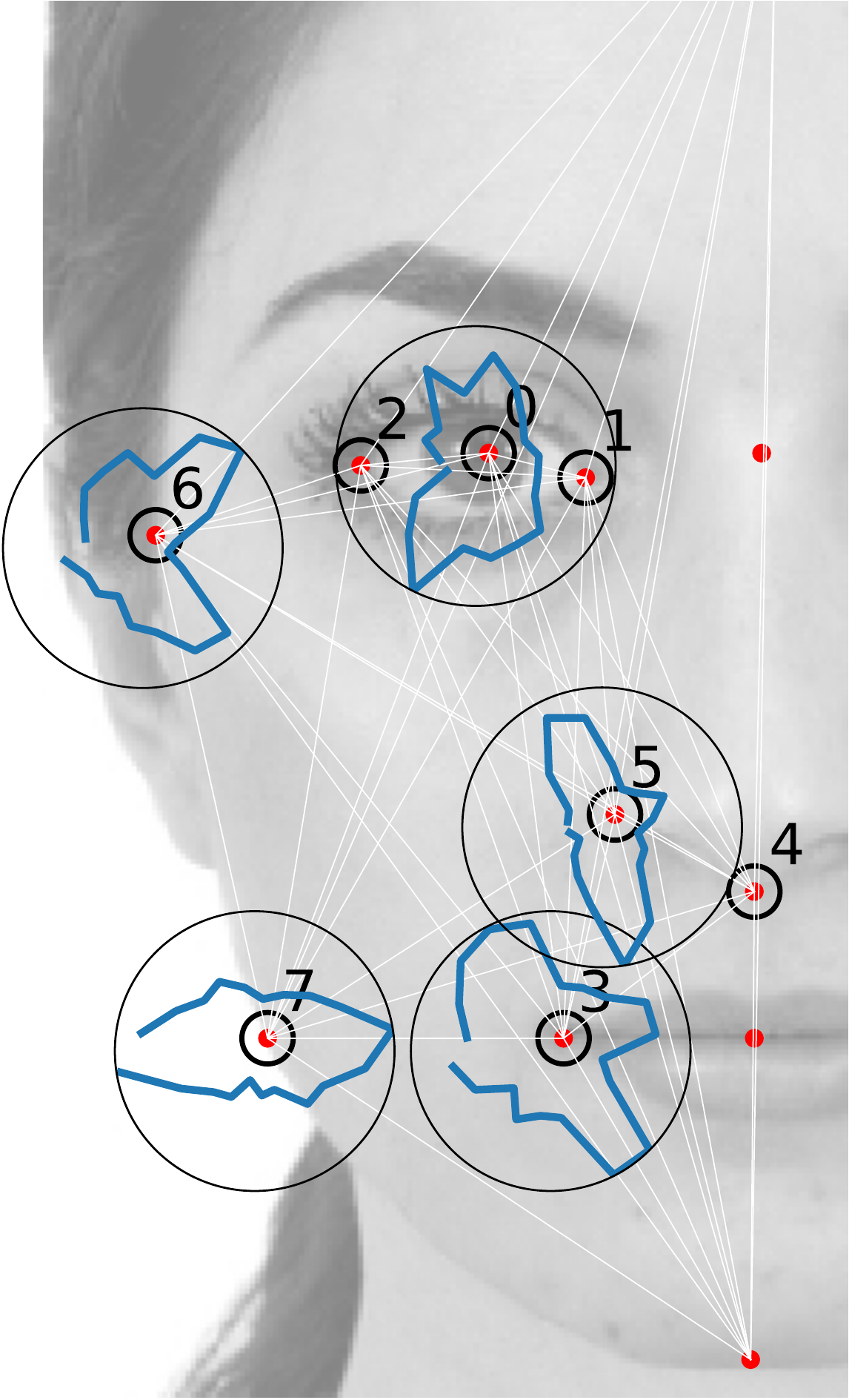}  
	\caption{Facial landmarks $i=0,\ldots,7$ whose 2D coordinates ${\vec r}_i$ constitute the face space (signaled with black circles). Their position in the figure correspond to the average position, $\<{\vec r}_i\>$. The background image corresponds to the texture degrees of freedom of the reference portrai, see \cite{ibanez2019}. The blue lines are polar hystograms ${\sf h}(\phi)$ (the radius is proportional to ${\sf h}(\phi)$) corresponding to the experimental distribution of angle landmark fluctuations around their average position. }
\label{fig:key}
\end{center}   
\end{figure}

The database consists in the set of landmark geometric coordinates ${\cal S}=\{{\bf r}^{(s)}\}_{s=1}^{S}$, where $s$ is the facial vector index corresponding to the $\cal N$ vectors sculpted by each of the $\Ns$ experimental subjects (hence: $S=\Ns {\cal N}$).\footnote{Indeed, the database ${\cal S}=\{{\bf r}^{(v,i)}\}$ is composed by $S=\Ns \times {\cal N}$ facial vectors labelled by a single index $s=1,\ldots,S$ or, alternatively, by a tuple of indices $(v,i)$ ($v=1,\ldots,{\Ns}$, $i=1,\ldots,{\cal N}$, $\Ns=95$, ${\cal N}=28$) referring to the $i$-th facial vector sculpted by the $v$-th subject (in a single genetic experiment, see \cite{ibanez2019}). In the SI we present a detailed analysis of the error estimation over the dataset, distinguishing inter- and intra-subject fluctuations. Similarly, the models may be concieved to account for intra- and inter-subject, or only for inter-subject correlations (see the SI).}   We will call ${\bf r}^{(s)}=(r^{(s)}_{(\sx,1)},\ldots,r^{(s)}_{(\sx,\nl)},r^{(s)}_{(\sy,1)},\ldots,r^{(s)}_{(\sy,\nl)})$ the vector whose  $2\nl$ components are the $(\sx,\sy)$ Cartesian coordinates of a set of $\nl=8$ landmarks, in units of the facial height. The landmarks (signaled with an empty circle in figure \ref{fig:key}) are a subset of the set of landmarks used for the image deformation in the \cite{ibanez2019} experiment (signaled with red points in  figure \ref{fig:key}). We will also refer to the 2D Cartesian vector of the $i$-th landmark as $\r_i=(\sx_i,\sy_i)$, and define the fluctuations of the landmark positions with respect to their average value as $\D_i=\r_i-\<\r_i\>$, where $\<\cdot\>$ denotes the experimental average, $\<\cdot\>=(1/S)\sum_s \cdot$. An important aspect of the dataset is that even the coordinates of the restricted set of $\nl=8$ landmarks, ${\bf r}^{(s)}$, are redundant and depend on 10 coordinates only, due to the presence of $2\nl-10=6$ constraints that result from the very definition of the face-space. Such constraints are described in detail in the SI.

\subsection{The Maximum Entropy models}

We propose two probabilistic generative models of the set of selected faces, inferred from the dataset $\cal S$. They result form the {\it Maximum Entropy (MaxEnt)} method \cite{jaynes1957,berg2017,nguyen2017,demartino2018}, which provides the maximum entropy probability distribution $\cal L(\cdot|\T)$ being consistent with the average experimental value of some observables of the data, $\<{\Sigma}\>$, that will be called {\it sufficient statistics}. $\cal L$ must satisfy $\<\Sigma\>_{\cal L}=\<\Sigma\>$, where $\<\cdot\>_{\cal L}$ refers to the theoretical average according to the distribution $\cal L$ (see a more precise definition in the SI). In the case of the {\it Gaussian} or {\it 2-MaxEnt model}, the sufficient statistics is given by the $2n$ averages $\<\Delta_\mu\>$ and by the  $2\nl\times 2\nl$ matrix of horizontal, vertical and oblique correlations among couples of vertical and horizontal landmark coordinates, whose components are $C_{\mu\nu}=\<\Delta_{\mu}\Delta_{\nu}\>$. In these equations, the $2\nl$ Greek indices $\mu=i,\c_i$ denote the $\c_i=\sx,\sy$ coordinates of the $i$-th landmark. 
The 2-MaxEnt model probability distribution takes the form (see the SI) of a Maxwell-Boltzmann distribution, 
${\cal L}({{\bm \Delta}}|\T)= \frac{1}{Z} \exp\left(-H[{{\bm \Delta}}|\T] \right)$. In this equation, $Z$ is a normalising constant (the partition function, in the language of statistical physics) depending on $\T$, and $H=H_2$ (the Hamiltonian) is the function:

\begin{eqnarray}
	\label{eq:H2}
	H_2[{{\bm \Delta}}|\T]&=&  \frac{1}{2} {\bm \Delta}^\dag \cdot  J \cdot {\bm \Delta} + {\bf h}^\dag\cdot {{\bm \Delta}}
.
\end{eqnarray}
The model depends on the parameters $\T=\{J,{\bf h}\}$, or the $2\nl\times 2\nl$ {\it matrix of effective interactions} $J$ and the $2\nl$ vector of {\it effective fields}, $\bf h$. Due to the symmetry of matrix $J$, the number of independent parameters in the 2-MaxEnt model is $D +D(D+1)/2$, where $D=2\nl$ is the dimension of the vectors of landmark coordinates $\bm \Delta$. The value of these parameters is such that the equations $\<{\bm \Delta}\>=\<{\bm \Delta}\>_{\cal L}$ and $\<\Delta_{\mu}\Delta_{\nu}\>_{\cal L}=C_{\mu\nu}$ are satisfied. This is equivalent to require that $\T$ are those that maximise the likelihood of the joint $\cal L$ over the database $\cal S$ (the Maximum Likelihood condition). The solution of such an inverse problem is (see SI): $J=C^{-1}$, ${\bf h}=J\cdot \<{\bm \Delta} \>$, and $Z={(2\pi)^\nl}\exp({\bf h}^\dag\cdot J^{-1}\cdot{\bf h}/2){(\det J)^{-1/2}}$, where the $-1$ power in equation $J=C^{-1}$ denotes the pseudo-inverse operation, or the inverse matrix disregarding the null eigenvalues induced by the database constraints (see SI). 

We will define as well the {\it non-linear}, or {\it 3-MaxEnt model}. In this case, the sufficient statistics is given by averages, pairwise correlations and correlations among 3-landmark coordinates, $C^{(3)}_{\mu\nu\kappa}=\<\Delta_\mu\Delta_\nu\Delta_\kappa\>$. The 3-MaxEnt model probability distribution ${\cal L}(\cdot|{\bf h},J,Q)$ assumes the Maxwell-Boltzmann form, with Hamiltonian $H=H_2+H_3$, where $H_3$ is:

\begin{eqnarray}
	H_3[{\vec{\bm \Delta}}|Q]&=&  \frac{1}{6} \sum_{\mu\nu\kappa} \Delta_\mu\Delta_\nu\Delta_\kappa Q_{\mu\nu\kappa} \label{eq:H3}
\end{eqnarray}
Besides ${\bf h}$ and $J$, the non-linear MaxEnt model depends on a further tensor of three-wise interaction constants among triplets of landmark coordinates. Consequently, the number of independent parameters is $D+D(D+1)/2+D(D-1)(D-2)/6$. The solution of the inverse problem for the non-linear MaxEnt model does not take a closed analytic form. The maximum likelihood value of the parameters $\T=({\bf h},J,Q)$ is numerically estimated by gradient ascent (see section \ref{sec:methods} and the SI).

\subsection{Learning in the non-linear models.} 

 In the case of the 3-MaxEnt model, we have numerically the maximum likelihood value of the parameters $\T^*$ by means of deterministic gradient ascent, using an algorithm that will be presented in a dedicated publication. A detailed explanation of the learning protocol may be found in the SI. Before inferring the data with the non-linear models (3-MaxEnt and GRBM) we have eliminated a subset of redundant $6$ coordinates from the original $2\nl$ coordinates. The data has been standardised in order to favor the convergence of the likelihood maximisation.

\subsection {The Restricted Boltzmann Machine model for unsupervised inference. } 
We have learned the data with the (Gaussian-Binary) Restricted Boltzmann Machine (GRBM) model of unsupervised inference \cite{wang2012,melchior2014}. 
We have employed the open-source software \cite{melchior2017pydeep} for the efficient learning of GRBM. The learning protocol and parameters are described in detail in the SI. 

\section{Results}
\label{sec:results}

We will now present an assessment of the description of the database according to the inference models described in the precedent section. In sec. \ref{sec:quality} we will argue that the Harmonic MaxEnt model is a faithful representation of the dataset. 
Finally, in subsection \ref{sec:Janalysis}, we  will argue that the matrix of effective interactions $J$ provides meaningful information, beyond the raw information present in the raw experimental measure $C$.

\subsection{Quality of the MaxEnt models as generative models \label{sec:quality}}

\subsubsection{Histograms of single landmark-angle fluctuations \label{sec:anglehistograms}}

\begin{figure}[t!]                        
\begin{center} 
\includegraphics[width=.45\columnwidth]{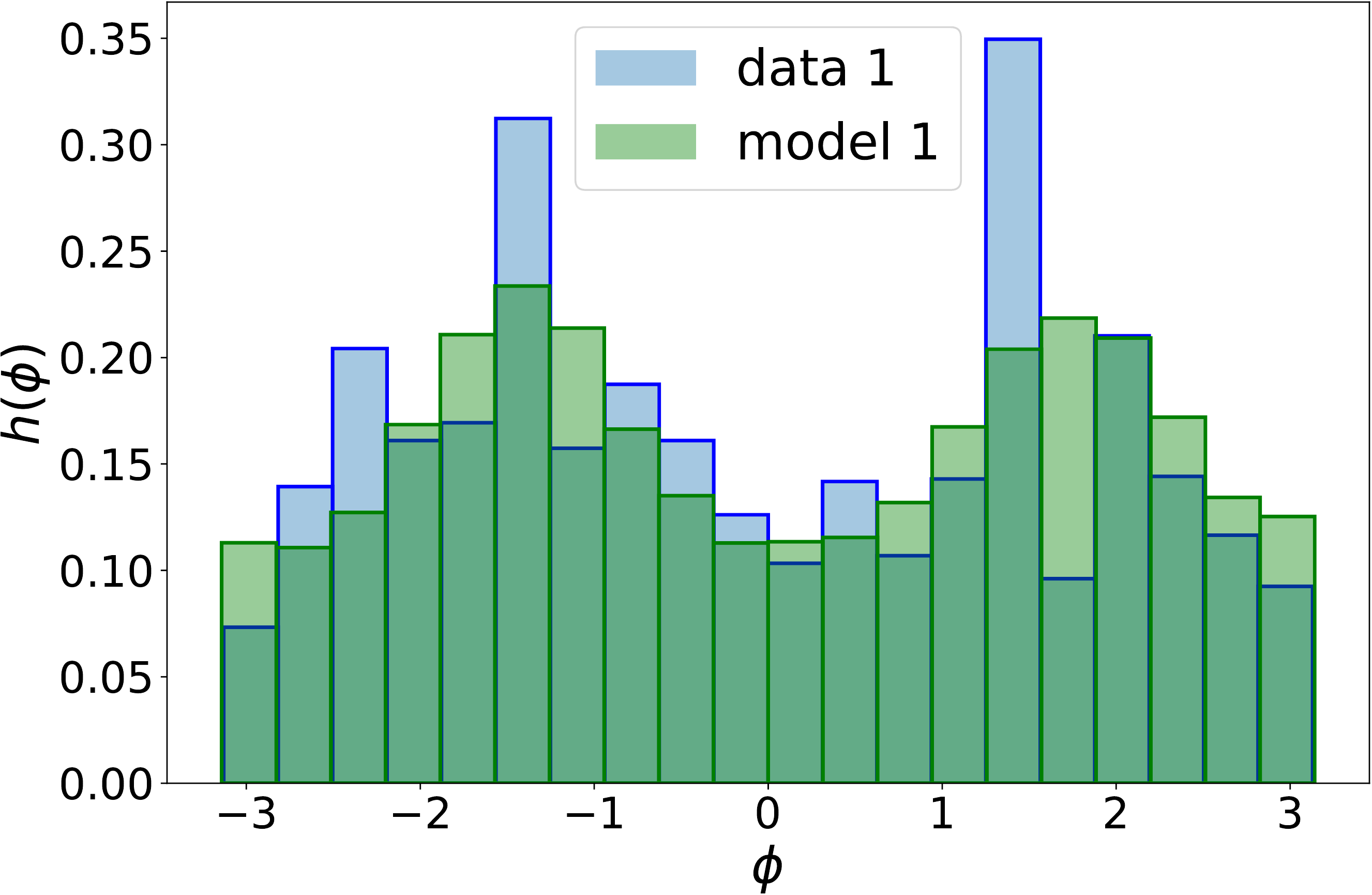}  
\includegraphics[width=.45\columnwidth]{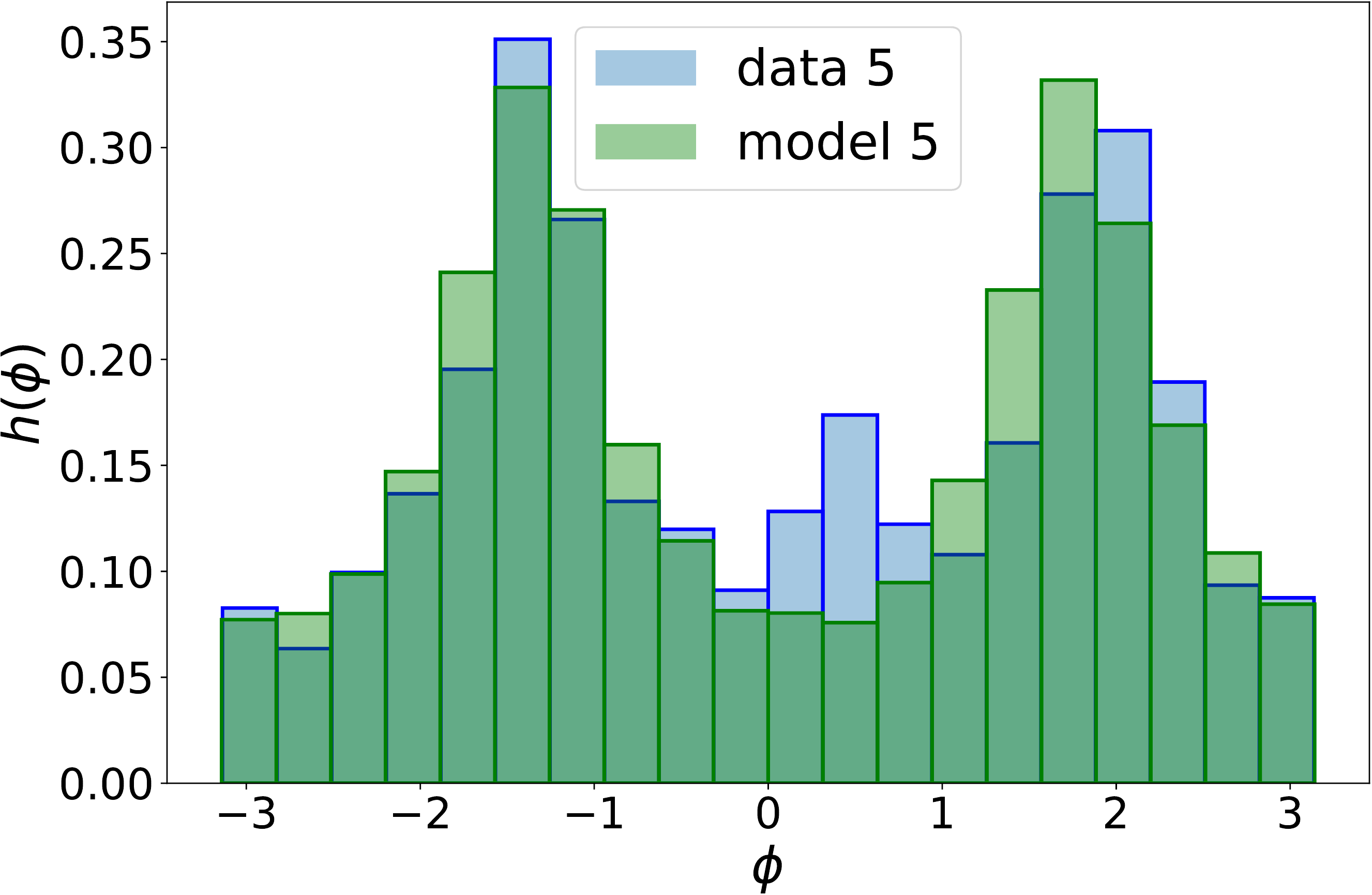} \\  
\includegraphics[width=.45\columnwidth]{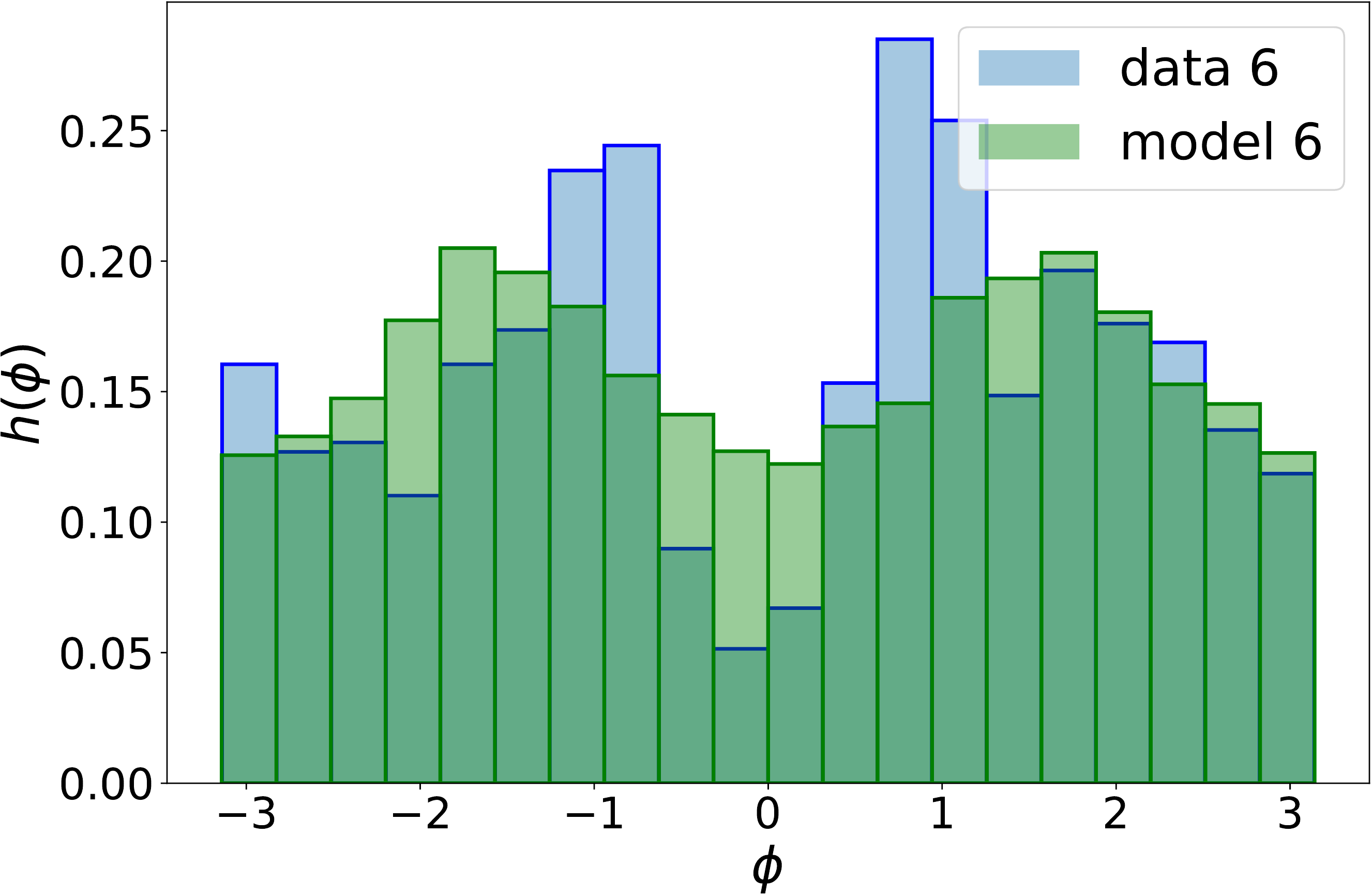}  
\includegraphics[width=.45\columnwidth]{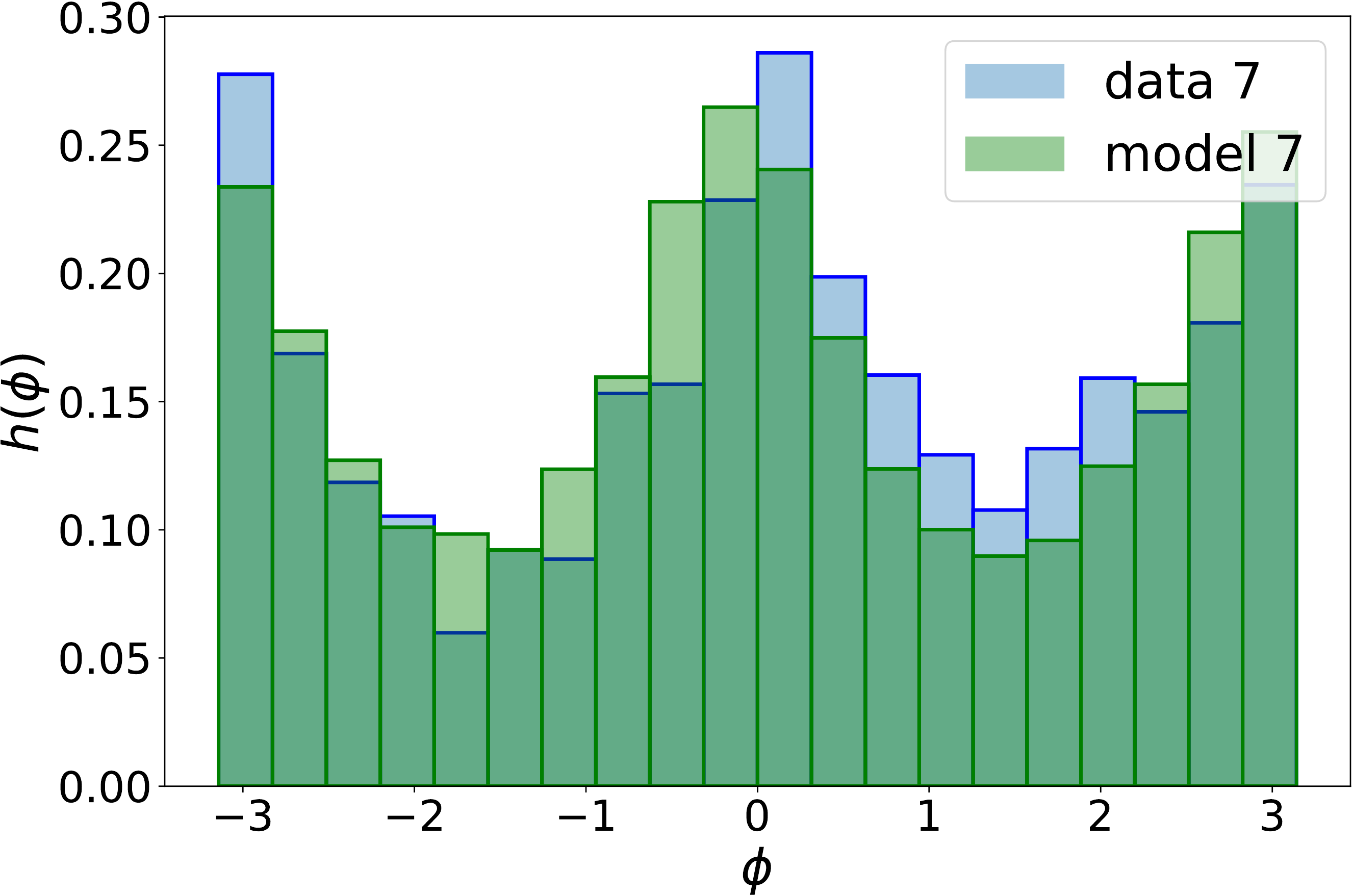}  
	\caption{Experimental ${\sf h}^{(i)}(\phi)$ versus theoretical, ${\sf h}^{(i)}_{\sf t}(\phi)$ distributions of angle landmark fluctuations, for several landmarks, $i=1,5,6,7$, see fig. \ref{fig:key} (from left to right, from top to bottom).  }
\label{fig:histogramsphi}
\end{center}   
\end{figure}

\begin{figure}[t!]                        
\begin{center} 
\includegraphics[width=.9\columnwidth]{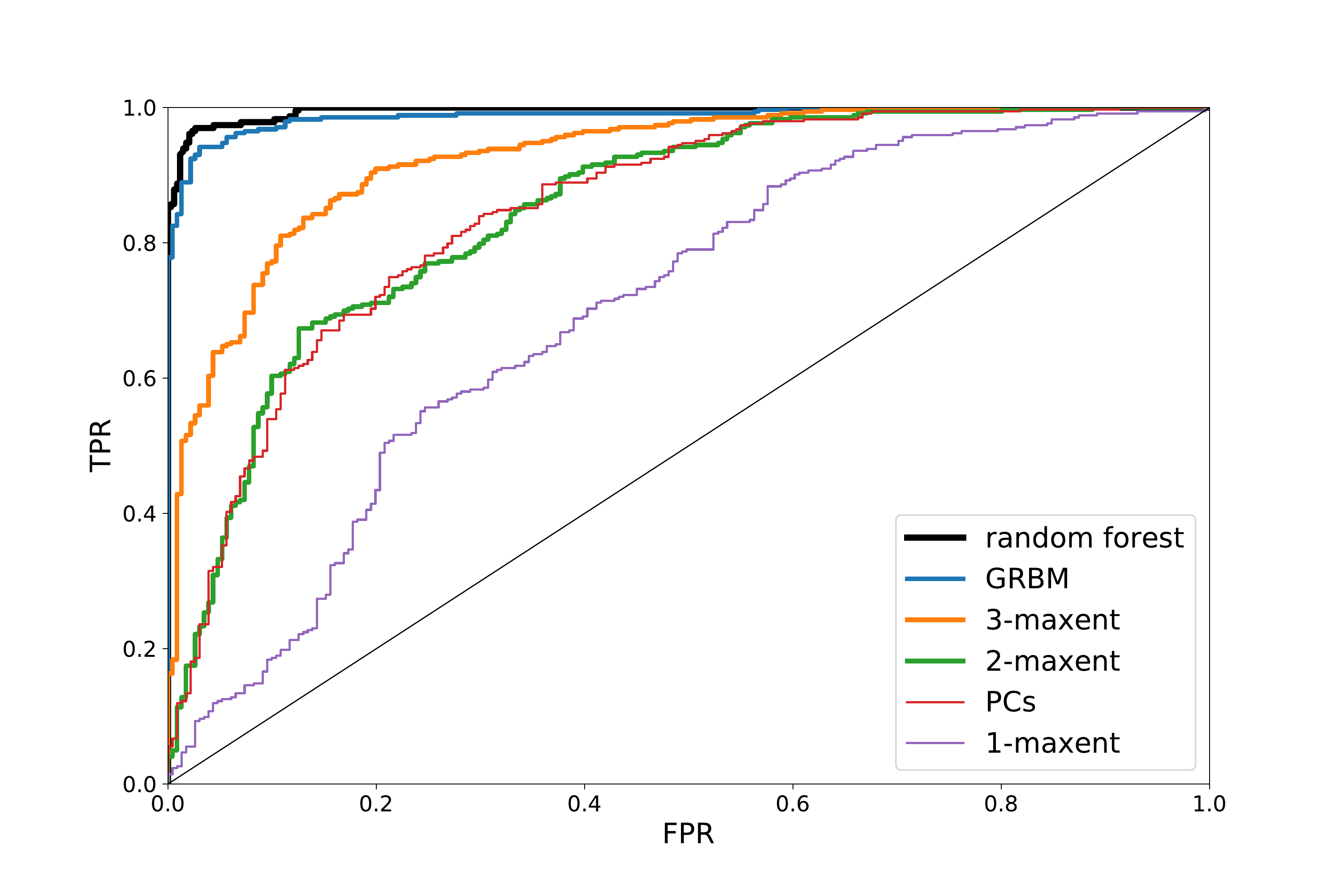}  
	\caption{ROC curves corresponding to the geneder classification. Different curves correspond to different algorithms. PC's refers to a $\sf t$-Student test of the difference in the principal components of a vector with respect to their average value in the $A$, $B$ sets.}
\label{fig:ROC}
\end{center}   
\end{figure}

The quality of the 2-MaxEnt generative model as a faithful description of the database may be evaluated by the extent to what the model ${\cal L}$ reproduces observables $O$ that it is not required to reproduce by construction. In other words, observables that cannot be written in terms of couples and triplets of coordinates $\Delta_\alpha$. The model is faithful in the extent to what $\<O\>\simeq\<O\>_{\cal L}$. 

The $i$-th landmark coordinates $\vec\Delta_i$ tend to fluctuate in the database with respect their average position $\<\vec\Delta_i\>=\vec 0$. As a nonlinear observable $O$ we will consider the {\it angle} that the $i$-th landmark fluctuation $\vec \Delta_i$ forms with the $\sf x$-axis. This quantity will be referred to as $\phi_i^{(s)}=\arctan(\Delta_{i,{\sf y}}^{(s)}/\Delta_{i,{\sf x}}^{(s)})$. In figs. \ref{fig:key},\ref{fig:histogramsphi}, we report the empirical histogram of angles, ${\sf h}(\phi_i)$ for some landmarks $i$. Remarkably, some landmarks' angle distribution exhibit local maxima,  probably reflecting their tendency to follow the direction of some inter-landmark segments (as it is apparent for the $3$-rd and $6$-th landmark's in figure \ref{fig:key}).\footnote{Interestingly, such local maxima are oriented along inter-landmark segments eventually joining such landmarks to other landmarks which {\it are not described in the facial vectors} $\D$: the landmarks $\ell_0$ and $\ell_{18}$, see the SI.} We have compared the empirical histograms with the theoretical $\phi_i$ distributions according to the model. These have been obtained as the angle histograms of a set of $S$ vectors $\bm \Delta$ sampled from the inferred distribution ${\cal L}(\cdot | \T)$ (see fig. \ref{fig:histogramsphi}). The 2-MaxEnt model satisfactorily reproduces most of the landmark angle distributions. The empirical angle distribution, in other words, is reasonably well reproduced by the theoretical distribution ${\sf h_t}(\varphi)=\int \d{\DD}\, {\cal L}(\DD|\T) \delta(\phi(\DD)-\varphi)$. 
It is important to remark that the model-data agreement on ${\sf h}(\phi_i)$ is observed also for large values of $\phi_i\in(-\pi/2,\pi/2)$ (see fig. \ref{fig:histogramsphi}), and not only for small values of $\phi_i$, for which it approximately becomes $\Delta_{i,\sy}/\Delta_{i,\sx}$ (whose average is related to the correlation $C_{\alpha\beta}$, see SI). 

We conclude that, very remarkably, a highly non-linear observable as $\phi$ is well described by the 2-MaxEnt model, albeit it has been inferred from linear (pairwise) correlations only. In this sense, the 2-MaxEnt model is a faithful and economic description of the dataset. This picture is confirmed by the results of the following section which suggest, however, that a description of the gender differences in the dataset require taking into account effective interactions of order $p>2$.

\subsubsection{Performance of the MaxEnt model in a classification task \label{sec:classification}}

We now further evaluate the quality of the 2- and 3-MaxEnt models by assessing their efficiency to classify a test database of vectors in two disjoint subsets ${\cal S}={\cal S}_A\cup {\cal S}_B$ corresponding to {\it the gender of the subject that sculpted the facial vector} in \cite{ibanez2019}. We compare such efficiency with that of the GRBM model of ANN (see \cite{wang2012,melchior2014}, sec. \ref{sec:methods} and the SI). This comparison allows to assess the relative relevance of products of $p$-facial coordinates $\Delta_\alpha$ in the classification task: averages ($p=1$), pairwise correlations ($p=2$), and non-linear correlations of higher, $p>2$ order (modelled by the 3-MaxEnt and GRBM models only). 

The dataset is divided in two disjoint classes ${\cal S}_A,{\cal S}_B$. Afterwards, both ${\cal S}_{A,B}$ are divided in training- and test- sets ($20\%$ and $80\%$ of the elements of ${\cal S}_{A,B}$, respectively), and inferred the $A$ and $B$ training sets separately, with the MaxEnt and GRBM models. This results in six ($\{2,3,G\}\times\{A,B\}$) sets of parameters  $\T^{2,3,G}_{A,B}$, where the super-index refers to the model. Given a vector $\DD$ belonging to the $A$ or $B$ test set, the {\it score} ${\sf s}(\DD)=\ln \L(\DD|\T_A)-\ln \L(\DD|\T_B)$ is taken as the estimation of the model prediction for $\Delta\in{\cal S}_A$. The resulting Receiver Operating Characteristic (ROC) curves \cite{murphy2012} are shown in fig. \ref{fig:ROC} for the various models considered.\footnote{ These consist in a scatter plot with the fraction of true positive classifications (${\rm TPR}$) in the ${\cal S}_A$ test-set versus the fraction of false positive classifications (${\rm FPR}$) in the ${\cal S}_B$ test-set, where each point corresponds to a different soil $\delta$ over the estimator ${\sf s}(\DD) \lessgtr \delta$ that we use to assign whether the model predicts that $\DD$ belongs to $A$ or $B$. The curve is invariant under reparametrizations of ${\sf s}\to f({\sf s})$ defined by any monotone function $f$.}

Considering only the averages $\<{\bm \Delta}\>$ as sufficient statistics  (or, equivalently, inferring only the fields $\bf h$ (and setting $J_{ij}=\sigma^{-2}_i\delta_{ij}$ in equation \ref{eq:H2}) results in a poor, near-casual classification (specially in the most interesting region of the ROC curve, for small ${\rm FPR}$ and large ${\rm TPR}$), see figure \ref{fig:ROC}. The 2-MaxEnt model allows,  indeed, for a more efficient classification. Rather remarkably, the 3-MaxEnt and GRBM models gradually improves the classification. We interpret this as an indication of the fact that non-linear effective interactions at least of fourth order are necessary for a complete description of the database. For completeness, we have included a comparison with the Random Forest (RF) algorithm \cite{murphy2012}. As shown in fig. \ref{fig:ROC}, RF achieves the highest classification accuracy (auROC$=0.995$, see SI). We notice that this does not imply that the unsupervised models are less accurate: the RF algorithm is advantaged, being a specific model trained to classify at best the $A,B$ partitions, not to provide a generative model of the $A$ and $B$ partitions separately. 

We report the maximal accuracy scores for all the algorithms: RF ($0.971$); GRBM ($0.952$); 3-MaxEnt ($0.865$); 2-MaxEnt ($0.764$); 1-MaxEnt ($0.680$). The 2-MaxEnt model efficiency is, as expected, compatible with that of a $\sf t$-Student test regarding the differences in the principal component values of $A$ and $B$ vectors, see fig. \ref{fig:ROC}. See the auROC scores of all the algorithms in the SI.

We conclude that, on the one hand, the subjects' gender strikingly determines her/his preferred set of faces, to such an extent that it may be predicted from the sculpted facial modification  with an impressively high accuracy score \cite{murphy2012}: a $97.1\%$ of correct classifications. On the other hand, the relative efficiency of various models highlights the necessity of non-linear interactions for a description of the differences among male and female facial preference criterion in the database. Arguably, such nonlinear functions play also a role in the cognitive process of facial perception. 
The criterion with which the subjects evaluate and discriminate facial images seems to involve not only {\it proportions} $r_\alpha/r_\beta$ (related to the pairwise correlations $C_{\alpha\beta}$, see SI), but also triplets and quadruplets of facial coordinates influencing each other (yet, see the SI for an alternative explanation)\footnote{As we explain in the SI, the non-Gaussian correlations of order 3 present in the dataset are, at least partially, not of cognitive origin, but due to an artifact of the numerical algorithm allowing subjects to sculpt their preferred facial vectors. However, we believe that the non-linear effective interactions that we infer do reflect the existence of non-linear operations playing a role in the cognitive process of facial evaluation. This is suggested by the fact that the introduction of non-linear effective interactions drastically  improves the gender classification.}.

\subsection{Analysis of the matrix of effective interactions \label{sec:Janalysis}}
We now show that the generative models may provide directly interpretable information. This is an advantage of the MaxEnt method, whose parameters, the effective interaction constants, may exhibit an interpretable significance.

The 2-MaxEnt model admits an immediate interpretation. The associated probability density ${\cal L}({{\bm \Delta}}|\T)= \exp\left(-H_2[{{\bm \Delta}}|\T] \right)/Z$ formally coincides with a Maxwell-Boltzmann probability distribution of a set of $\nl$ interacting particles in the plane (with positions $\vec\Delta_i$, $i=1,\ldots,n$), subject to the influence of a thermal bath at constant temperature. Each couple $i$,$j$ of such fictitious set of particles interacts through an harmonic coupling that corresponds to a set of three effective, virtual springs with non-isotropic elastic constants, $\Jx_{ij},\Jy_{ij},\Jxy_{ij}$  corresponding (see equation \ref{eq:H2}) to horizontal, vertical and oblique displacements, $\Delta_{i,\sx}-\Delta_{j,\sx}$, $\Delta_{i,\sy}-\Delta_{j,\sy}$, and $\Delta_{i,\sx}-\Delta_{j,\sy}$, respectively. 

The inferred effective interactions are more easily interpretable if one considers, rather than their $\sx\sx$, $\sy\sy$ and $\sx\sy$ components, the {\it longitudinal} and {\it torsion effective interactions}, $J_{ij}^{\parallel}$ and $J_{ij}^{\perp}$, respectively. The longitudinal coupling $|J_{ij}^{\parallel}|$ may be understood (see the SI for a precise definition) as the elastic constant corresponding to the virtual spring that anchors the inter-$ij$ landmark distance  to its average value, $\<r_{ij}\>$ (where $\vec r_{ij}=\vec r_{j}-\vec r_i$). In its turn, the torsion interaction $|J_{ij}^{\perp}|$ is the elastic constant related to fluctuations of $\vec r_{ij}$  along the direction normal to $\vec r_{ij}$ or, equivalently, to fluctuations of the {\it $ij$-segment angle}, with respect to its average value that we will call $\alpha_{ij}=\arctan(\<r_{ij,\sy}\>/\<r_{ij,\sx}\>)$.

In fig. \ref{fig:Jslongtran}-A,B we show the quantities $|J_{ij}^{\parallel}|$ and $|J_{ij}^{\perp}|$ for those couples $i,j$ presenting a statistically significant value (for which the $\sf t-$value $t_{ij}=|J_{ij}|/\sigma_{J_{ij}}>1$, see sec. \ref{sec:methods}). The width of the colored arrow over the $i,j$ segment is proportional to $|J_{ij}^{\parallel}|$ (blue arrows in fig. \ref{fig:Jslongtran}-A) and $|J_{ij}^{\perp}|$ (red arrows in fig. \ref{fig:Jslongtran}-B). We notice that there exist inter-landmark segments for which $|J_{ij}^\parallel|$ is significant while $|J_{ij}^\perp|$ is not (as the $0,4$ or the $5,6$ segments) and vice-versa (as the $6,7$ and $2,5$). This suggests that $|J_{ij}^{\parallel}|$, $|J_{ij}^{\perp}|$ actually capture the cognitive relative relevance of distance fluctuations around  $\<r_{ij}\>$, and of angle fluctuations around $\alpha_{ij}$.  

In the SI we explain in more detail the analogy with the system of particles. We also analyse the dependence of the torsion and longitudinal effective interactions, $|J_{ij}^{\parallel}|$ and $|J_{ij}^{\perp}|$, with the average distance and angle of the $ij$ inter-landmark segment, showing that there is a moderate decreasing trend of $|J_{ij}^{\parallel}|$ with $\<r_{ij}\>$

We remark that the prominent importance of the inter-segment angles $ij$ highlighted in fig. \ref{fig:Jslongtran}-B is fully compatible with the analysis presented in ref. \cite{ibanez2019} at the level of the oblique correlation matrix $\Cxy$, and it goes beyond, as far as it quantitatively assess their relative relevance. As we will see before, such information cannot be retrieved from the experimental matrix $C$ only.

\subsubsection{Extra information retrieved with effective interactions} A relevant question is to what extent the inferred effective interactions $J$ provide interpretable information, inaccessible from the raw experimental correlations $C$. In the general case, couples of variables may be statistically correlated through spurious correlations, even in the absence of a causal relation among them (see the SI). 
In the present case, the main source of spurious correlations is the presence of the constraints among various landmark coordinates. The MaxEnt inference eventually subtracts (through the pseudo-inverse operation) the influence of the constraints from matrix $J$, which describes the essential effective mutual influence among pairs of coordinates of prominent relative importance (see the SI). 

\begin{figure}[t!]                        
\begin{center} 
\includegraphics[width=.325\columnwidth]{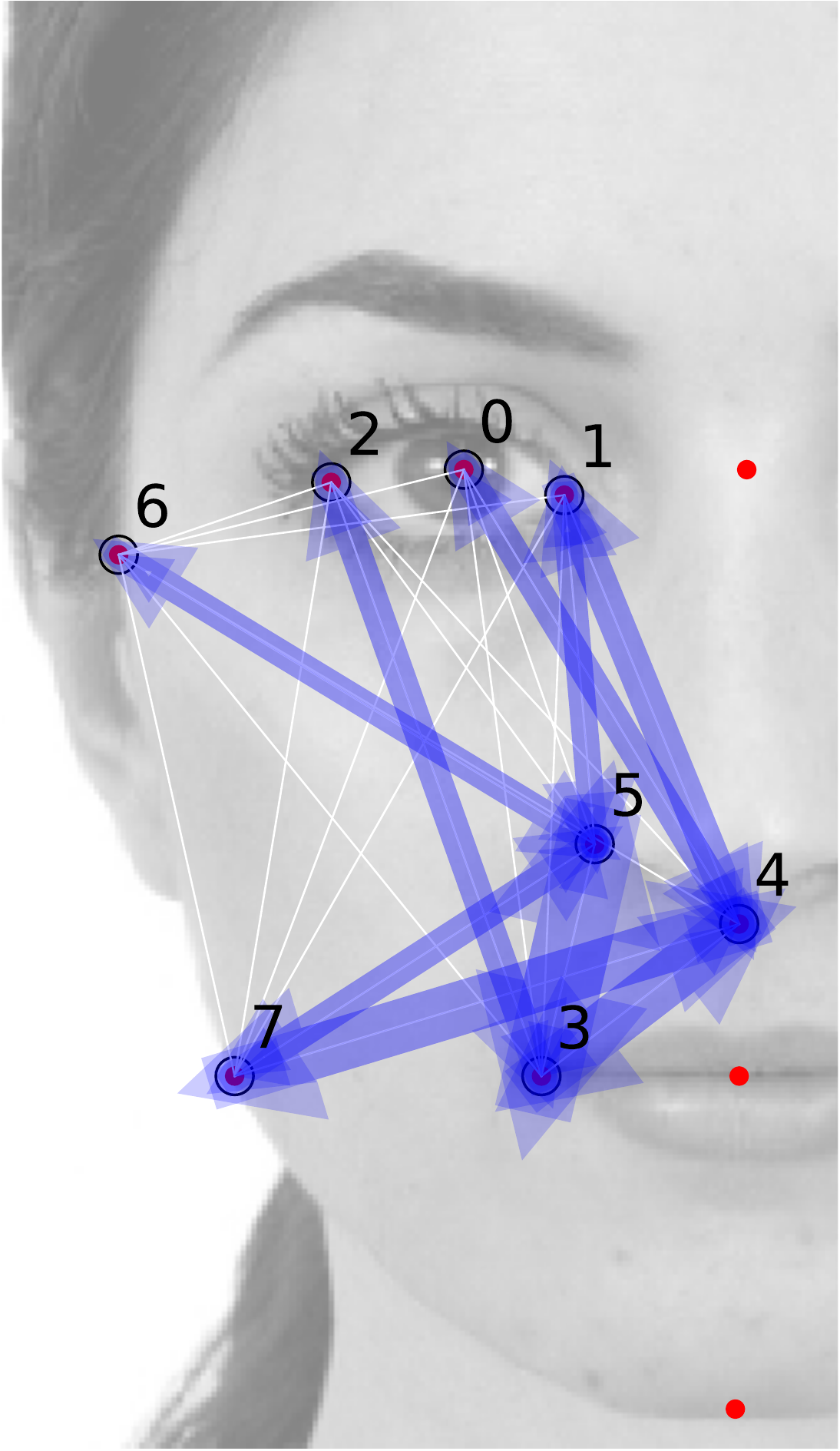}  
\includegraphics[width=.325\columnwidth]{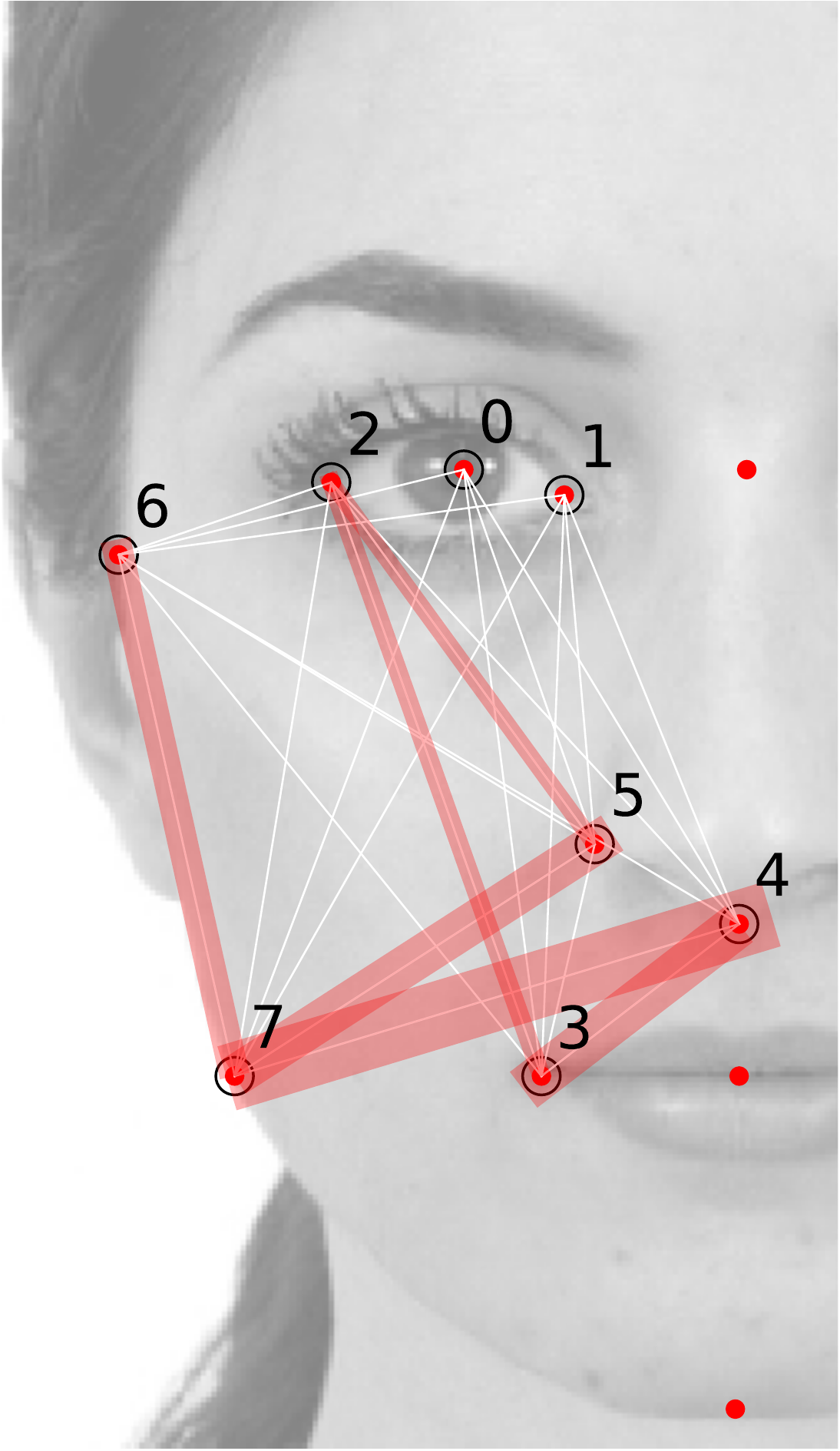}  
\includegraphics[width=.325\columnwidth]{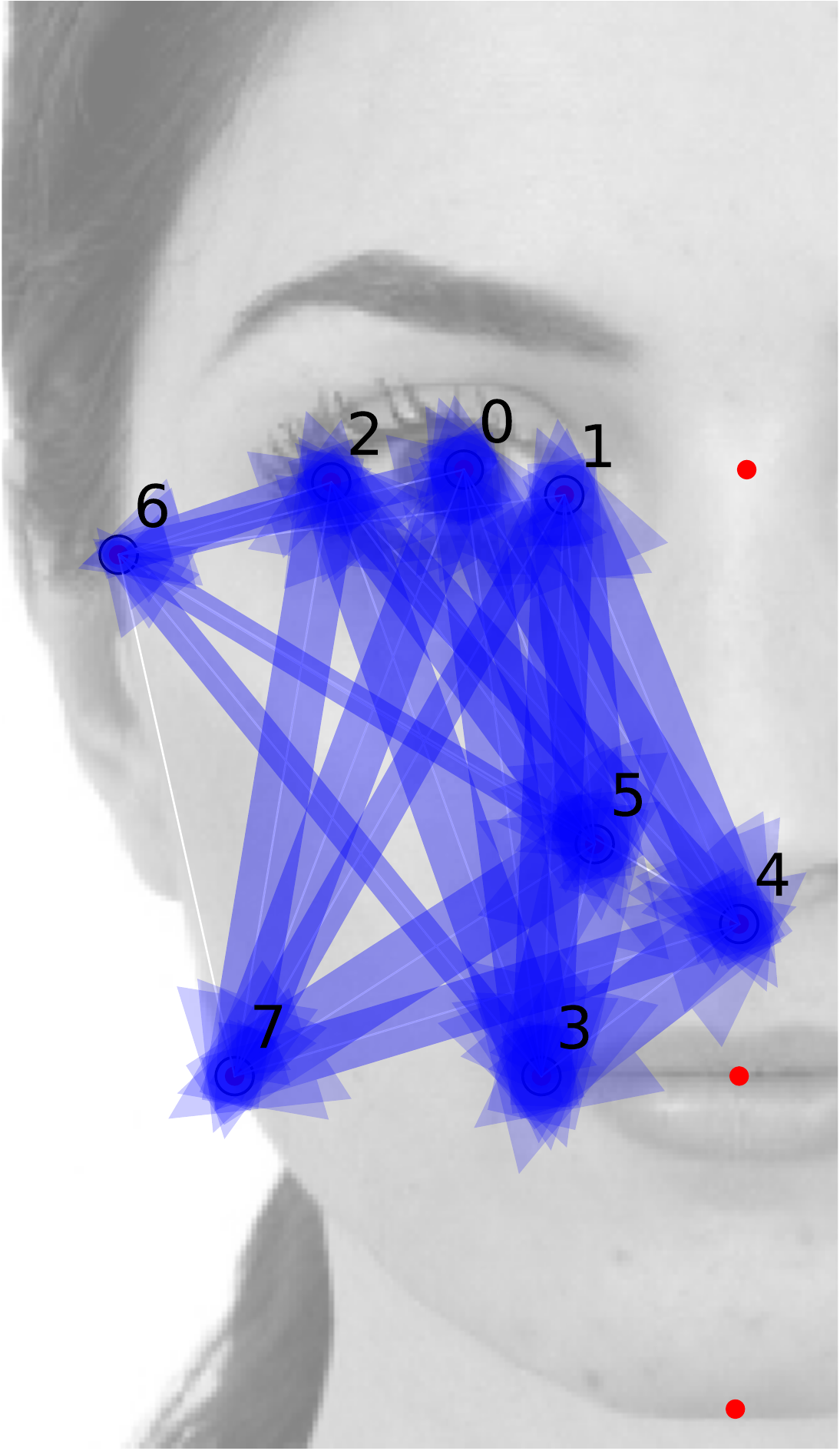}  
	\caption{Modulus of the matrices $J^\parallel$ (left), $J^\perp$ (center), $C^{\parallel}$ (right). The width of the arrow joining the $i$-th and $j$-th landmarks is proportional to $|A_{ij}|$, where $A$ is the corresponding matrix. Only significant matrix elements have been plotted: only those exhibiting a $t\sf $-value larger than one: $t_{ij}=|A_{ij}|/\sigma_{A_{ij}}>1$. }
\label{fig:Jslongtran}
\end{center}   
\end{figure}

The differences among $C$ and $J$ matrices are shown in fig. \ref{fig:Jslongtran}-C, where the arrows represent the absolute value of the raw experimental matrix elements $C^{\parallel}_{ij}$. Remarkably, all but two of the matrix elements result statistically significant ($t$-value $>1$): matrix $C^{\parallel}$ can be hardly used to assess the relative relevance of various inter-landmark segments. 
The effective interaction matrix $J$ disambiguates the correlations propagated by the constraints, attributing them to the effect of a reduced set of   elastic constants, in the particle analogy. Such attribution is not unambiguous, but the result of an inference procedure. 

An in-depth comparison among $C$ and $J$ is presented in the SI, where we consider also the alternative method of avoiding the constraints, consisting in inferring from a non-redundant set of coordinates. We conclude that, in the general case, and for the sake of the interpretation of the effective interactions, it may convenient to infer from a database of redundant variables, eliminating {\it a posteriori} the influence of the null modes associated to the constraints, or the $C$-eigenvectors associated to non-linear constraints. 

\section{Discussion}
\label{sec:discussion}

We have presented an application of the Maximum Entropy (MaxEnt) method to the inference of a database of preferred facial modifications. Much work has been devoted to the supervised inference of the rating assigned to a set of face-space vectors --a technological, more than scientific aim. Nevertheless, such studies indirectly allow for an assessment of the relative impact of various facial traits on perceived beauty \cite{laurentini2014}. These remain, along with the nature of inter-subject differences in preferred faces, controversial questions. As a further tool to address them, we here propose an alternative inference scheme in which the variability to be inferred is not the average rating assigned to {\it different natural faces}, but {\it the inter-subject variability} of preferred modifications in a subspace of the face-space. 

To our knowledge, the present work is the first unsupervised inference approach in facial preference research. It presents at least three innovative, distinguishing traits: the inference is performed over a set of preferred facial modifications, each one corresponding to a different subject; it avoids the use of ratings, a limited quantity \cite{laurentini2014,ibanez2019}, and the curse of dimensionality associated to the rating of natural faces; it accounts for the mutual influence of couples and $p-$plets of facial features, $p>2$, hence beyond a principal component analysis. At the same time, the inference is done in terms of interpretable quantities involving ``physical'' facial coordinates only, hence overcoming the black-box issue.

The present approach allows to clarify several aspects regarding facial preference. First, that the cognitive mechanisms related to facial discrimination in the brain definitely involves the mutual influence of couples of landmarks, more than the positions of single landmarks. Moreover, the results suggest as well that non-linear operators of the facial coordinates are involved in the cognitive process (see secs. \ref{sec:anglehistograms},\ref{sec:classification}). The recent de-codification of the neural code for facial recognition in the primate brain \cite{chang2017} has revealed that recognition is based on linear operations (or projections in the geometric and texture principal axes) in the face-space. The evaluation of attractiveness (probably related to the higher-level inference of abstract personality attributes \cite{cunningham1995,edler2001,little2014,oosterhof2008,todorov2011,walker2016,abir2017}) may be a more complex process, requiring the evaluation of non-linear operations in the face-space. 

Second, and rather remarkably, the introduction non-linear effective interactions allows for an astonishingly high classification efficiency of the facial vectors according the subject's gender: a $97.1\%$ of correct classifications, for the random forest algorithm. This implies that the subject's gender strikingly determines her/his facial preference criteria. Our results strongly suggest that such an influence is not reflected in {\it differences in the position of single landmarks}.  Moreover, the MaxEnt method provides the set of proportions (see fig. \ref{fig:Jslongtran}) and triplets of landmarks which male and female subjects tend to evaluate more differently. 

In summary, the MaxEnt method provides a faithful probabilistic generative model of the database of preferred facial variations of ref. \cite{ibanez2019}, describing the inter-subject fluctuations around of the preferred facial vector (given a reference background portrait). We demonstrate that such fluctuations encode, and may accurately reveal, meaningful information regarding the subject that selected the facial vector, such as her/his gender. As also suggested by the {\it multiple motive hypothesis}, many other subjects' attributes and distinguishing traits may influence her/his personal preferences in the face-space and, hence, could be retrieved from their sculpted facial vectors. By sculpting a set of preferred facial modifications the subject reveals a large amount of information, arguably more than the rating of natural images would do. 

According to this idea, the analysis of a subjects' sculpted facial vectors could be used for an assessment, on a voluntary basis, of her/his abstract personality dimensions most influencing attractiveness, as the subjects' ``traits of desired personality'' \cite{little2006}. In the same way, and as far as the {\it complete subjectivity picture} proposed in \cite{ibanez2019} is valid, an accurate enough experiment of preferred variation sculpture could be used as a voluntary fingerprint identity test.

Furthermore, the present study represents a novel case of study for the application of  the  MaxEnt method for unsupervised inference, in particular for the assessment of the relevant order of interaction by comparison with an ANN model and for the comparison among various strategies of inference in the presence of constraints. 

Possible extensions of this work are the generalisation to different datasets and facial codification methods (see the SI for a discussion on the generality of the present approach); the classification of different subject's features from her/his set of sculpted faces; the analysis of medical imaging data characterised by the 3D position of landmark points.



\section{Acknowledgements}

We acknowledge Andrea Gabrielli, Irene Giardina, Carlo Lucibello, Giorgio Parisi and Massimiliano Viale for inspiring discussions. Particular thanks to Andrea Cavagna for his suggestions and comments to the draft.

\section{Supplementary Information}

\subsection{Introduction to the Maximum Entropy principle: Correlations vs effective interactions}

Consider an $n$-dimensional space of vectors, ${\bf x}=(x_i)_{i=1}^n\in \chi$, along with a set of $K$ observables, $O_k:\chi\to\mathbb{R}$, $k=1,\ldots,K$. The {\it maximum entropy} approach \cite{jaynes1957,berg2017,nguyen2017,demartino2018} provides the {\it most probable} probability distribution $P({\bf x}|{\bm \lambda})$, ${\bf x}\in \chi$, which is consistent with a fixed value of the operators, in the sense that their average according to $P$, $\<O_k\>_P$ is constraint to assume a fixed value, 

\begin{equation}
\<O_k\>_P=o_k
\label{eq:maximumentropyconstraint}
\end{equation}
(where $\<O_k\>_P=\int\d \x\, O_k(\x) P(\x|{\bm \lambda})$). In other words the {\it maximum entropy} probability distribution  $P_{\rm me}$ is the one exhibiting highest entropy (i.e., the most random, or less structured distribution) subject to the constraint (\ref{eq:maximumentropyconstraint}), and to no other constraint. It assumes the form:

\begin{equation}
P_{\rm me}(\x) = \frac{1}{Z({\bm \lambda})} \exp \left[\sum_{k=1}^K \lambda_k O_k(\x) \right]
\label{eq:maenconstraint}
\end{equation}
$Z({\bm \lambda})$ being a normalizing constant. The maximum entropy probability distribution is, hence, formally identical to a Maxwell-Boltzmann distribution in the canonical ensemble at temperature $=1$, with effective Hamiltonian ${\cal H}=-\sum_k \lambda_k O_k$. It is important to remark that no assumption at all has been done about thermal equilibrium, ergodicity, nor about the existence of an effective interaction in energy units: the Maxwell-Boltzmann form is a consequence of the maximum entropy assumption --reflecting, rather, {\it absence} of hypothesis-- of a probability distribution subject to constraints on the average of some operators. The values of the Lagrange multipliers $\lambda$'s in (\ref{eq:maenconstraint}) are such that the  constraints in (\ref{eq:maximumentropyconstraint}) are satisfied. 

In the context of unsupervised statistical inference, one infers from a finite number $M$ of experimental measures of the observables $O_k$, to which correspond the values $o_k^{(m)}$, $m=1,\ldots,M$. The maximum entropy distribution provides a generative probabilistic model for the data, that is aimed to be a faithful representation of the experimental dataset and, at the same time, a {\it generalisation} of the dataset, not too dependent on the specific realisation of the database that is being inferred. For this reason, $P$ is chosen to reproduce the experimental value of {\it a limited} set of observables, depending on the dataset. Ideally, a faithful and general model should be consistent with the minimum set of experimental averages that allow to reproduce some essential database properties and, at the same time, that may be significantly inferred given the database finiteness. Once the observables have been selected, a possible choice for {\it their value} $o_k$ (determining the value of the parameters $\lambda_k$) in \ref{eq:maximumentropyconstraint} is the experimental average, $\<O_k\>=(1/M)\sum_{m=1}^M o_k^{(m)}$. This choice $o_k=\<O_k\>$ is equivalent to the Maximum Likelihood prescription of the whole experimental database:

\begin{eqnarray}
\{\lambda^*_k\}_k = \arg\max_{\{\lambda_k\}_k}  \sum_{m=1}^M \ln P(\x^{(m)}| {\bm \lambda}) 
\label{eq:generalizedentropy}
\end{eqnarray}
where $\x^{(m)}$ is the $m$-th experimental configuration, and $o^{(m)}_k=O_k(\x^{(m)})$. The parameters $\lambda$ are called effective interactions, in the language of statistical physics. In the case that the observables to be reproduced by $P$ are the data correlations of order  $n\le p$ (where the correlations of order $n$ are defined as $C^{(n)}_{i_1, \cdots ,i_n}=\<x_{i_1} \cdots  x_{i_n}\>$), the effective interactions assume the form of $n$-th order tensors $J^{(n)}$ coupling $n$-plets of vector coordinates, with $n=1,\ldots,p$. 

A self-consistency criterion for the choice of the sufficient statistics $O_k$ is that of calculating different nontrivial observables according to $P$ (different from the sufficient statistics, i.e., observables that $P$ is not required to reproduce by construction, and that cannot be expressed in terms of the sufficient statistics), and comparing them with their experimental counterparts. In particular, a criterion is that of choosing the $n\le p$-th order correlations as sufficient statistics, with $p$ being the minimum value such that the $p+1$-th order experimental correlations are satisfactorily reproduced by $P_{\rm me}$ (i.e., $\<x_{i_1} \cdots  x_{i_{p+1}}\>\simeq \<x_{i_1} \cdots  x_{i_{p+1}}\>_P$), and such that all the parameters corresponding to such sufficient statistics may be significantly inferred from the data.

{\bf Correlations and effective interactions.} The effective interaction tensors $J^{(n)}$ may admit, in certain circumstances, an interpretation regarding the mutual effective influence among variables, beyond the statistical correlation among them (whose experimental value is $C^{(n)}$). Correlations and effective interactions are actually different. Focusing for simplicity in $p=2$, the pairwise correlations are but the statistical  consequence of the effective interactions among couples of landmarks causing them. This is the case in the direct problem (the calculation of $C^{(2)}$ from $J^{(2)}$): in this case, it may happen that the matrix $J^{(2)}$ is sparser than $C^{(2)}$: there are couples of variables not influencing each other that, nevertheless, result statistically correlated. In the direct problem, the Maximum Entropy method may allow for a discrimination of the spurious correlations of couples of components that are correlated although they do not influence each other (but are, instead, commonly and mutually influenced by other components). This a frequent phenomenon in biological data \cite{schneidman2006,cavagna2015}, with an obvious interpretation in statistical physical terms: in the general case, the mutual influence among a sparse set of couples of bodies propagates statistically, leading to {\it emergent, collective phenomena}. A paradigmatic and extreme case of this general phenomenology is critical behaviour \cite{kadanoff2000}, in which microscopic interactions lead to  macroscopic correlations: long-range and high-order body correlations originate from short-range, sparse and pairwise interactions \cite{schneidman2006}). 

\begin{figure}[h!]                        
\begin{center} 
\includegraphics[width=.33\columnwidth]{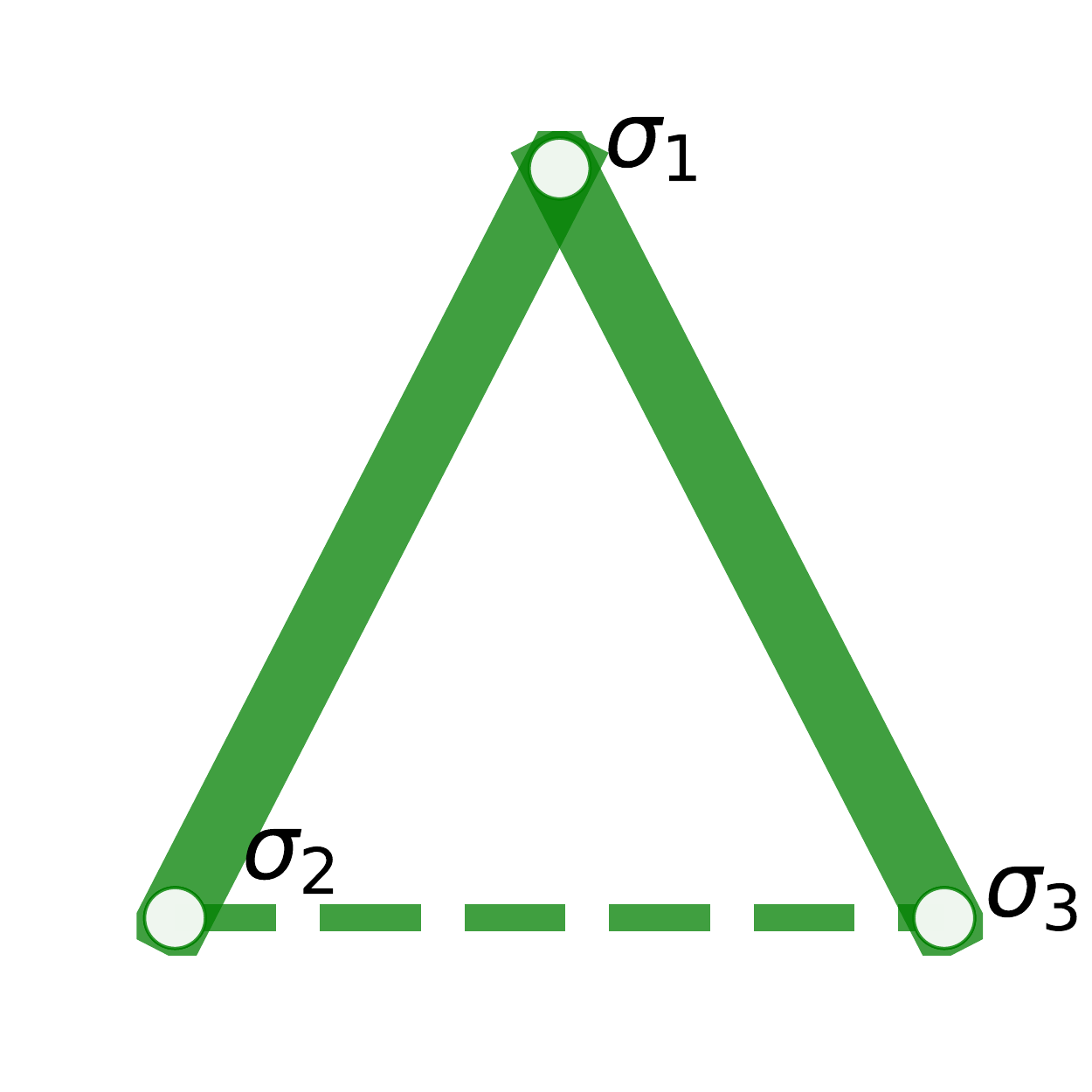}  
\includegraphics[width=.33\columnwidth]{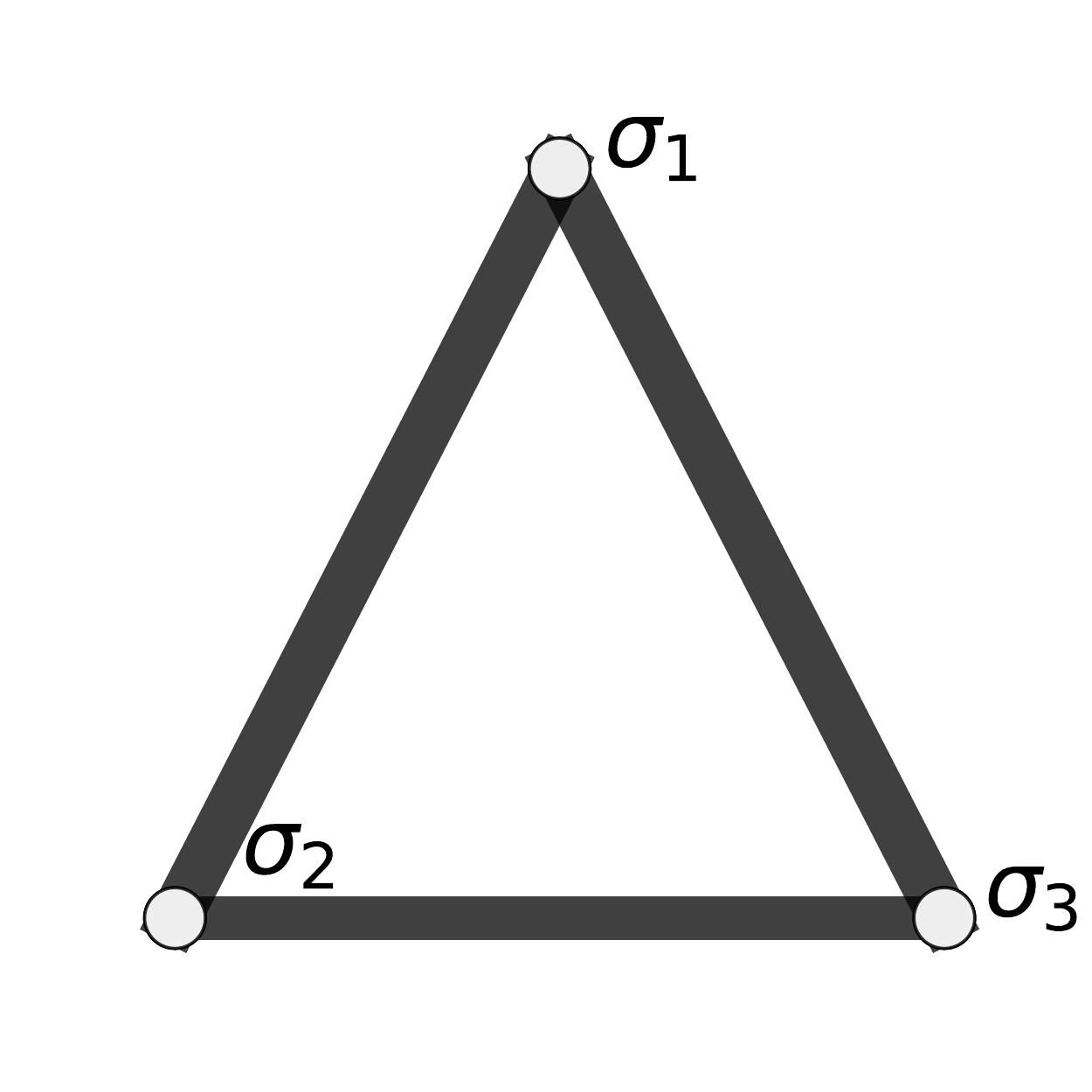}  
	\caption{Effective interactions (left) versus the emerging statistical correlations (right) among three binary variables, $\sigma_i=\pm 1$, $i=1,2,3$. The effective interactions are given by the symmetric matrix $J_{12}=J_{13}=2$, $J_{23}=-3/4$ (i.e., 2,3 are negatively coupled). The correlations $\<\sigma_i \sigma_j\>$ are given by their expectation value according to the Maxwell-Boltzmann probability distribution corresponding to a pairwise interaction given by matrix $J$: $\<\sigma_i \sigma_j\>=\sum_{\bm\sigma} \sigma_i\sigma_j P({\bm\sigma})$, with $P({\bm\sigma})=\exp(+\sum_{i<j}\sigma_i\sigma_j J_{ij})/Z$ where $Z=\sum_{\bm\sigma}\exp(+\sum_{i<j}\sigma_i\sigma_j J_{ij})$. The line width is proportional to the absolute value  $|A_{ij}|$ of the corresponding matrix element. The dashed line in the $J$ triangle indicates that  $J_{23}$ is negative (i.e., there is a tendency of $2$ to decrease when $3$ increases and vice-versa). Such tendency is, however, not reflected in the correlation matrix, which presents all positive elements.}
\label{fig:spuriouscorrelations}
\end{center}   
\end{figure}

Probably the simplest illustration of the emergence of spurious statistical correlations is that of three variables ($x_1,x_2,x_3$ in fig. \ref{fig:spuriouscorrelations}, of which only two of them are strongly interacting (in the figure $J_{12}=J_{13}=2$), while the second and third are moderately interacting (or even negatively interacting, as in the figure: $J_{23}=-1$). Such an information is not accessible from the emerging correlations $\<x_i x_j\>_{P(\cdot|J)}$ (in the direct problem), revealing a strong, positive correlation among all the variables. Conversely, in the inverse problem (i.e., when an empirical correlation matrix is given, resulting from an average of a sufficiently high number of measures), the Maximum Entropy method may provide not only a generative model, $P(\cdot |J_*^{(n)})$, but also the most probable interaction matrices $J_*^{(n)}$ suggesting that, indeed, the correlation among $2$ and $3$ is (most likely, given the data and the sufficient statistics, and within the Maximum Likelihood hypothesis) a statistical consequence of the mutual influences of $1,2$ and of $1,3$. This information is not  unambiguously elicited from the data, but the result of an inference procedure: the most probable guess given the inference model and the ambiguity induced by the data finiteness.

\subsection{Maximum Entropy inference from pairwise correlations with a priori constraints {\label{sec:constraints}}}

In this section we solve the problem of the Maximum Entropy (MaxEnt) inference from pairwise correlations (i.e., $p=2$), in the presence of linear constraints involving the coordinates. 

In the absence of constraints, the Maximum Likelihood solution to the problem, equation (\ref{eq:generalizedentropy}) is analytic and straightforward. Suppose that one infers from a database composed by $S$ experimental realisations $\{\x^{(s)}\}_{s=1}^S$ of a real, $D$-dimensional vector, $\x=(x_i)_{i=1}^D \in \mathbb{R}^D$. The sufficient statistics to infer from is by hypothesis the correlation matrix (supposing null-average vectors): $C_{ij}=\<x_ix_j\>$ where $\<\cdot\>$ represents, as before, the experimental average: a symmetric, positive definite matrix. The MaxEnt probability distribution is, consequently, the multi-variate normal distribution:

\begin{equation}
	{P}(\x|J) = \left[\frac{\det J}{(2\pi)^n}\right]^{1/2}\exp[- \frac{1}{2} \x^\dag J \x]
\label{eq:normaldist}
.
\end{equation}
The Maximum Likelihood solution for the matrix $J$, $J^*$, is that satisfying that the theoretical pairwise correlations $\<x_ix_j\>_{P}={J^{-1}}_{ij}$ coincide with the experimental correlations $C_{ij}$. This is satisfied whenever $J^*=C^{-1}$. 

We now consider the presence of linear constraints involving the coordinates, $x_i$. Each linear constraint may be expressed in the form ${\bm a}_j^\dag \x=c_j$, being ${\bm a}_j$ a real $D$-dimensional vector and $c_j$ a real constant, for the $j$-th constraint. If all the vectors in the database $\{\x^{(s)}\}_{s=1}^S$, are subject to the constraints, each constraint induces a null mode (a zero eigenvalue) in the experimental covariance matrix. In this case, the Maximum Likelihood solution to the problem, i.e., the probability distribution $P(\cdot|J)$ such that $\<x_ix_j\>_P = C_{ij}$ cannot simply be $J^*=C^{-1}$, since matrix $C$ actually exhibits a vanishing determinant.

We will see that Maximum Likelihood solution in this case is $J^*=C^{-1}$, where the $-1$ exponent means the pseudo-inverse operation, a generalisation of the matrix inverse operation in which the null eigenvalues are avoided. We define the pseudo-inverse of the real, square matrix $A$ as:

\begin{equation}
	{A^{-1}}_{ij}  = \sum_{k | \epsilon_k\ne 0}  \epsilon_k^{-1} \eta^{(k)}_i \eta^{(k)}_j
\end{equation}
where $\epsilon_k$, $\eta^{(k)}_j$ are the $k$-th eigenvalue and the $j$-th component of the $k$-th eigenvector of $A$, respectively.

We first consider the solution of the direct problem, $\<x_ix_j\>_P$ from $J$, in a situation in which the interaction matrix $J$ is such that ${\rm rank}(J)=r<D$. In other words, $J$ exhibits $D-r$ null eigenvalues. Suppose that the eigenvalues $\lambda_j$ of $J$ are ordered in decreasing order, so that $\lambda_j=0$ for $j=r+1,\ldots,D$. In this case, the probability distribution $P(\x)$ in equation \ref{eq:normaldist} is, trivially, constantly zero since the determinant of matrix $J$ vanishes. However, we can define a real function in the space of the $D$-dimensional variables $\x$:

\begin{equation}
\tilde P(\x) = \frac{1}{\tilde Z}\exp[- \x^\dag J \x]\label{eq:tildeP} \\
\end{equation}
where $\tilde Z$ is a normalising factor involving the non-zero eigenvalues of $J$ only:

\begin{equation}
	\tilde Z = \left[ \frac{\tilde\det J}{(2\pi)^r}\right]^{1/2} \qquad  \tilde\det J = \prod_{k=1}^r \lambda_k
\end{equation}

The function $\tilde P$ may be considered as a normalised probability distribution, but only over the $r$-dimensional subspace of ${\mathbb R}^D$ expanded by the first $r$ eigenvectors of $J$: ${\mathbb S}_+={\rm span} \{{\bf e}^{(k)}\}_{k=1}^r$ with $1\le k\le r$. In other words, $\tilde P$ is a probability distribution on the subspace of ${\mathbb R}^D$, ${\mathbb S}_+$, defined by the vectors that are already subject to the constraints, for any value $c_j$'s of the constants associated to the constraints.  

One can easily define a proper, normalised probability distribution $P$ defined in ${\mathbb R}^D$, by regularising the null modes associated to the constraints:

\begin{equation}
	P(\x) = \tP(\x) \prod_{j=r+1}^D \delta(x'_j - {\sf c}_j)
	\label{eq:myP}
\end{equation}
where the $D-r$-dimensional vector $(x'_{r+1},\ldots,x'_{D})$ is a vector of the projection of $\x$ in a basis of vectors expanding the space of the constraints (as the vectors ${\bf a}_j$ defining the constraints, before, $x'_{j+r}={\bf a}_j^\dag \x$). On the other hand, we define the $r$-dimensional vector $\x'=(x'_1,\ldots,x'_r)$ as the projection of $\x$ over the first $r$ eigenvectors of $J$ (associated to a non-null eigenvalue): $\x'=E\x$, where $E$ is the $r\times D$ matrix defined as the row-disposed eigenvectors, $E_{ij}=e^{(i)}_j$\footnote{We make notice that $E E^\dag=\1_r$ but $E^\dag E\ne \1_D$ (where $\1_d$ is the identity matrix in $d$ dimensions).}.

To each of the null eigenvalues corresponding to a constraint ${\bf a}\cdot \x=c$, is associated an {\it invariance} of $\tP$ with respect to the linear operator $G(c)$ that changes the value of the constraint, i.e., such that ${\bf a}\cdot(G(c)\x)=c$:

\begin{equation}
\tP(\x|J)=\tP(G(c)\x|J) \label{eq:symmetry}
.
\end{equation}
Indeed, $G$ acts on the subspace ${\mathbb S}_0$ only, while it is the identity in the subspace ${\mathbb S}_+$ (where ${\mathbb S}_0$ is defined as the complement of $\mathbb{S}_+$, i.e., $\mathbb{R}^D={\mathbb S_+}\times {\mathbb S}_0$). In the physical language, each eigenvector corresponding to a constraint is called a {\it null mode}, and represents a {\it symmetry} reflected in the invariance of the function $\tP$ with respect to the symmetry. The function $P$, in its turn, represents vectors for which the symmetry is broken, as the value of the constraint has been fixed. For example, if the vectors are constrained to have a constant sum of its components, $\sum_{i=1}^Dx_i = c$, the corresponding eigenvector, or null mode ${\bf e}$, has all its components equal to $e_i=D^{-1/2}$. Consequently, the function $\tP$ is invariant under scale transformations. 

We are now interested in the calculation of a general $n$-order cumulant  $\<\<x_{s_1}\cdots x_{s_n}\>\>_{P}$ according to the distribution $P$ in (\ref{eq:myP}), with $s_j=1,\ldots,D$. As it can be seen immediately, the $n$-th order cumulant is related to the $n$-th order derivative of the generating function through the standard cumulant expansion equation:

\begin{equation}
	\<\<x_{s_{1}} x_{s_{2}}... x_{s_{n}}  \>\>_P =  \left.\frac{\partial^{n} \ln \tilde Z[\h']}{\partial h'_{s_{1}} \partial h'_{s_{2}} ... \partial h'_{s_{n}}}\right|_{\h={\bf 0}}
\label{npointcorrelationtilde}
\end{equation}
where the generating function $\tilde Z[\h]$ has the form:

\begin{equation}
\tilde Z[\h]= \left[\prod_{k=1}^{r} \int_{-\infty}^{+\infty} dz'_{k} \right] e^{-\frac{1}{2}\x^\dag J \x + {\bf h}^\dag \x}
\label{eq:Ztilde}
\end{equation}

We notice that $\tilde Z[{\bf 0}]=\tilde Z$. We would like an analytical expression for $\tilde Z[\h]$. Using the relations $\x'=E\x$ and $J=E^\dag \Lambda E$, where $\Lambda$ is the $r\times r$ diagonal matrix whose diagonal is $\lambda_1,\ldots,\lambda_r$, one obtains: 

\begin{equation}
	\tilde Z[\h]=\prod_{k=1}^{r}  \int_{-\infty}^{+\infty} dx'_{k}  \exp\left[{-\frac{1}{2} {x'}_k^2\lambda_k  + x'_k h'_k}\right] 
\end{equation}
where ${\bf h}'=E{\bf h}$. Using Gaussian integration rules, one finds:

\begin{equation}
	\tilde Z[\h]=\tilde Z  e^{\frac{1}{2}\h^\dag J^{-1} \h}
\end{equation}
where $J^{-1}$ is the pseudo-inverse of matrix $J$, ${J^{-1}}_{ij} =  \sum_{k\le r}  \lambda_k^{-1} e^{(k)}_i e^{(k)}_j$. This equation is the generalisation of the standard expression for the generating function of the multi-variate normal distribution, to the case in which matrix $J$ is be non-invertible. As it is evident from the expression of $\tilde Z[\h]$ and from the cumulant expansion (\ref{npointcorrelationtilde}), and as it happens with the normal distribution, the only non-zero cumulant is the second-order cumulant $\<\<x_ix_j\>\>_P$, equal to the correlation $\<x_ix_j\>_P$ in the case of null-averaged vectors. Its form is, from equation \ref{npointcorrelationtilde}:

\begin{equation}
	\< x_i x_j \>_P = {J^{-1}}_{ij}
\label{2pointcorrelationtilde}
\end{equation}
where $J^{-1}$ is the pseudo-inverse of matrix $J$.

The theoretical correlation matrix whose elements are $\<x_ix_j\>_P$ exhibits, as matrix $J$ does, rank equal to $r$. Hence, the theoretical correlation matrix $M_{ij}=\<x_i x_j\>$ as a function of $J$ (i.e., the direct problem) is $M=J^{-1}$. Consequently, the $J^{*}$ that is needed to satisfy $C_{ij}=\<x_i x_j\>_P$ given an experimental correlation matrix $C$ (i.e., the inverse problem) is $J^{*}=C^{-1}$, where the $-1$ power means the pseudo-inverse operation.

\subsection{Constraints in the database of facial modifications {\label{sec:constraints_faces}}}

\begin{figure}[h!]                        
\begin{center} 
\includegraphics[width=.33\columnwidth]{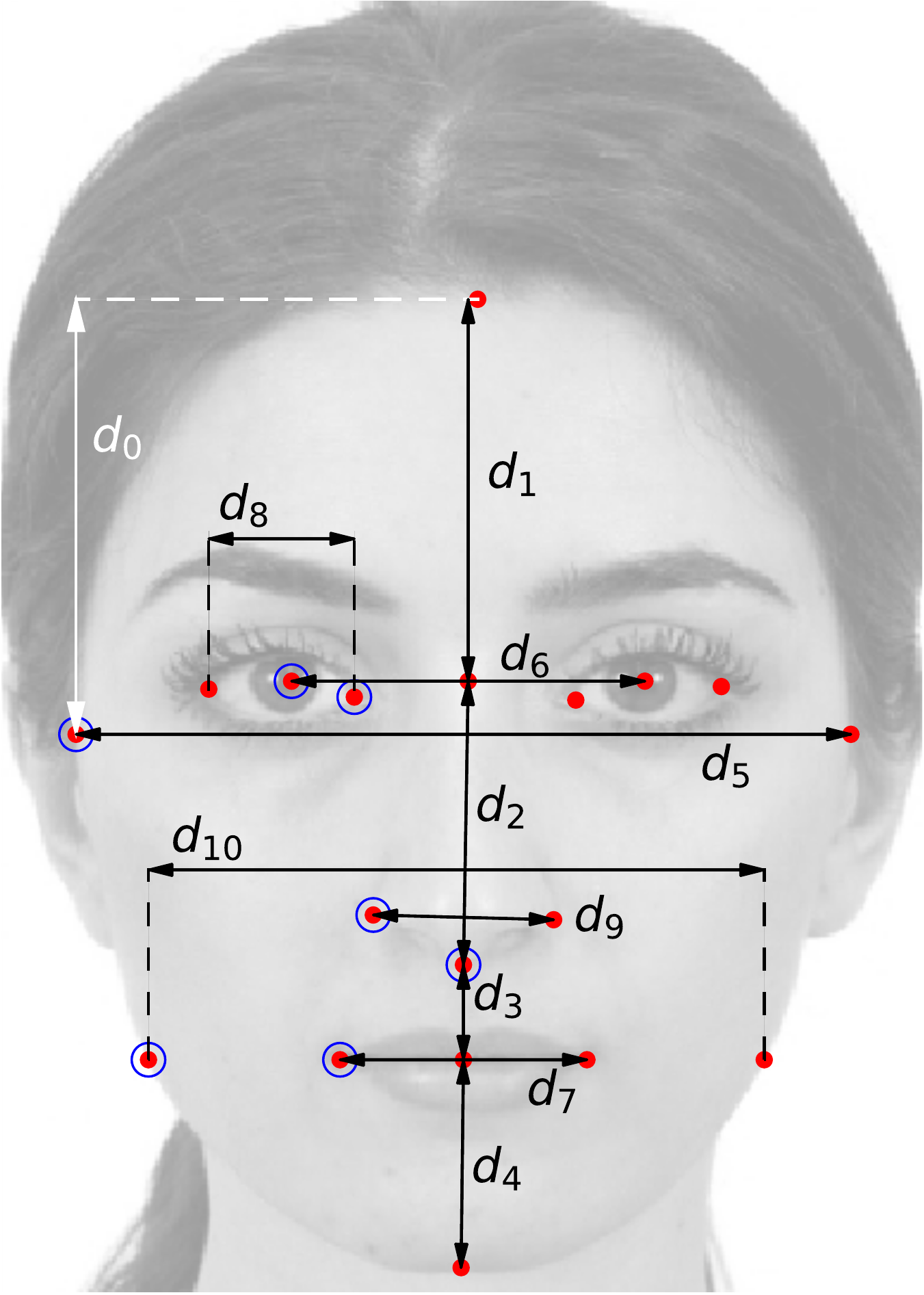}  
\includegraphics[width=.33\columnwidth]{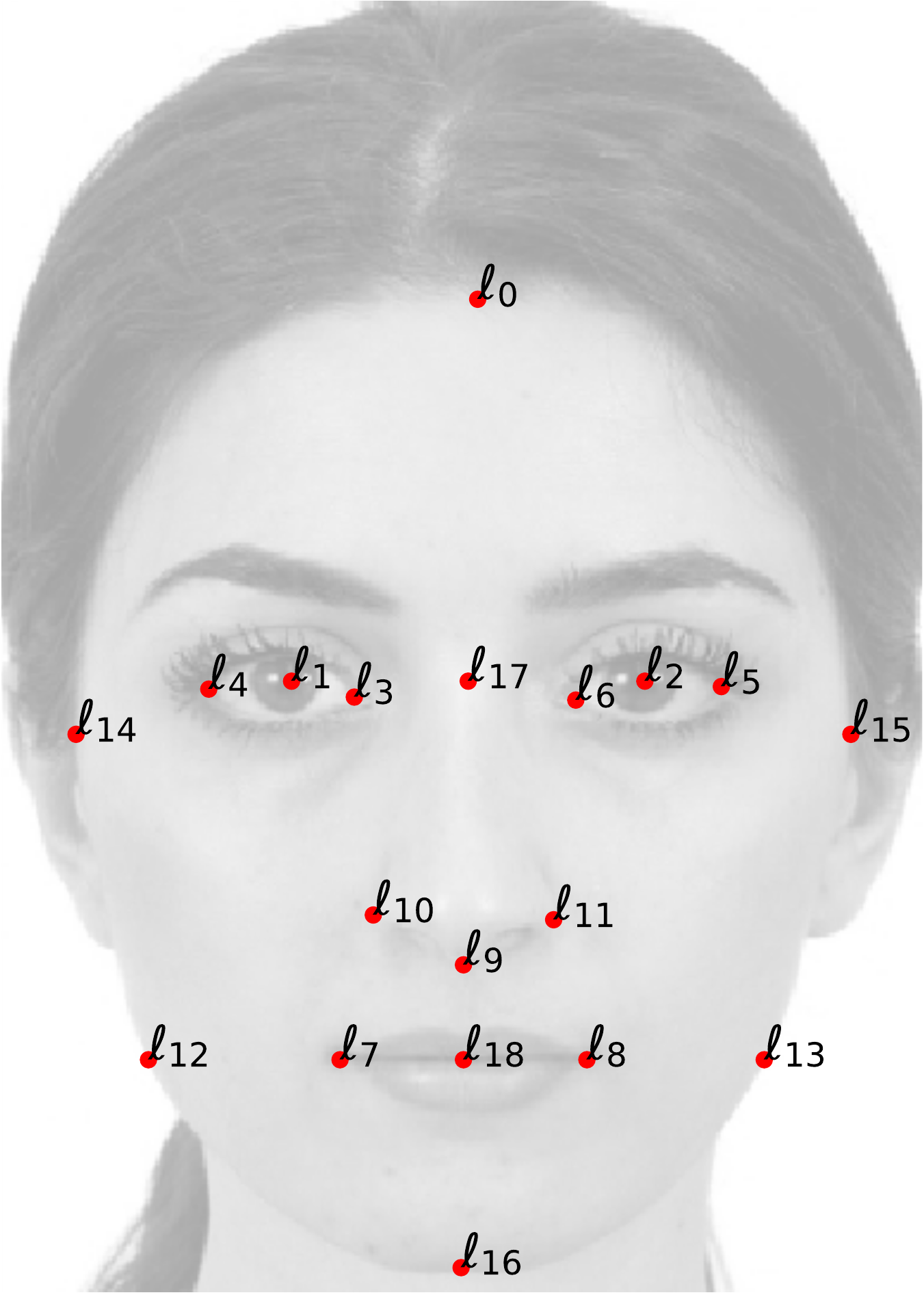}  
	\caption{Left: definition of the face space in terms of inter-landmark distances $\bf d$. The landmark coordinates are, instead, the $\sx,\sy$ 2D coordinates ${\vec r}_i$ of  a subset of seven landmarks, marked with blue circles. Right: all the landmarks used for the facial deformation algorithm described in \cite{ibanez2019}.}
\label{fig:key}
\end{center}   
\end{figure}

As we have explained in the main text, the facial modifications in the 2017 experiment are defined in terms of a set of 17 2D landmarks which are redundant in the sense that the positions of some of them may be deduced in terms of 10 coordinates only. The facial landmarks are, in particular, symmetric by construction, and hence the coordinates of the right-side landmark are determined given those of the left side. We have, hence, considered a subset of $\nl=8$ landmarks only (see figure \ref{fig:key} Furthermore, the landmark coordinates $r_{{\sf c},i}$ (where ${\sf c=\sx,\sy}$) in the database ${\cal S}$ are still subject to 6 constraints: indeed, $2\times \nl-6=10$, the numer of degrees of freedom. For instance, the nose endpoint abscissa is constrained to lie in the center of the image, $\Delta_{\sx,9} = 0$, and the jaw landmark is defined to be at the same heigth of the mouth, $\Delta_{\sy,3} = \Delta_{\sy,7}$. We have intentionally kept such redundant information in the inferred training database. Indeed, the redundant information turns to be necessary for the correct {\it interpretation} of the inference parameters, as we will see in section \ref{sec:inferring_constraints}.

The database ${\cal S}=\{{\bm \Delta}^{(s)}\}_{s=1}^S$ of facial displacements is, hence, highly constrained in the various facial coordinates. The following constraints hold, for all the vectors in the database:

\begin{eqnarray} 
	\Delta_{4,\sx} &=& 0  					\label{eq:constraint1} \\
\Delta_{4,\sy} - \Delta_{5,\sy} &=& {\rm constant} 			\label{eq:constraint2} \\
\Delta_{7,\sy} - \Delta_{3,\sy}  &=&  0				\label{eq:constraint3} \\
\frac{\Delta_{0,\sy}-\Delta_{1,\sy}}{\Delta_{0,\sx}-\Delta_{1,\sx}}&=&{\rm constant} 	\label{eq:constraint4}  \\
\frac{\Delta_{0,\sy}-\Delta_{2,\sy}}{\Delta_{0,\sx}-\Delta_{2,\sx}}&=&{\rm constant}  	\label{eq:constraint5}\\
\frac{\Delta_{1,\sx}-\Delta_{0,\sx}}{\Delta_{0,\sx}-\Delta_{2,\sx}}&=&{\rm constant} 	\label{eq:constraint6} 
\end{eqnarray} 
The last three constraints ensure that the eye aspect ratio remains unchanged with respect to the average facial vector ${\bm \Delta}={\bf 0}$ (otherwise, the image deformation algorithm corresponding to the landmark deformation ${\bf 0}\to{\bm \Delta}$ could lead to an ellipse-like shaped eye). Indeed, the constants in the right-hand side of each equation correspond to the value that assumes the left-hand side quantity in the average facial vector. 

Each of these constraints induces a null mode in one of the correlation matrices $\Cx$, $\Cy$, $\Cxy$ (those involving only $\sx$'s coordinates, in $\Cx$; those involving only $\sy$'s, in $\Cy$; those involving a $\sx$ and a $\sy$ coordinate, in $\Cxy$). As we have shown before, the inverse problem in this case is solved through the matrix pseudo-inverse operation. In these circumstances, the probability distribution ${\cal L}(\cdot|J,{\bf h})$ described in the main article refers to a probability distribution in the $10$-dimensional sub-space of coordinates that are invariant under the symmetries associated to the constraints ($\tP$, in the notation of section \ref{sec:constraints}). Strictly speaking, to become a proper probability distribution in the space of facial modification vectors $\bm \Delta$  it has to be regularised as in section \ref{sec:constraints}:

\begin{equation}
P({\bm \Delta}|J,{\bf h})= \left( \prod_{\mu} \delta(\Delta'_\mu-{\sf c}_\mu) \right) \tP ({\bm \Delta}|J,{\bf h})
\end{equation}
where $\tP$ is the distribution that in the main article is called ${\cal L}=\exp(-H)/Z$, the product is over the ${\bm \Delta}$ components over the $6$ eigenvectors of the global correlation matrix with a null eigenvalue, $\lambda_\mu=0$, and ${\bm \Delta}'=E{\bm\Delta}$, with  $E$ being the matrix of column-eigenvectors of $C$ (and of $J$). 

\subsection{Correlation vs interaction matrices \label{sec:CvsJ}}

In the particular case of our database, the main source of spurious correlations is not collective behaviour but the presence of the {\it a priori} constraints  among various landmark coordinates, which are imposed in the experimental construction of the face space vectors (see sec. \ref{sec:constraints_faces} and \cite{ibanez2019} for a precise description of the constraints), and that play the role of the strong interaction $1,2$ in fig. \ref{fig:spuriouscorrelations}. The MaxEnt method {\it subtracts} the effect of such constraints and provides a sparser  interaction matrix. Our MaxEnt inference scheme discounts the effect of constraints since we eliminate the matrix $C$ eigenvectors corresponding to the constraints (through the pseudo-inverse operation $C^{-1}$), see sec \ref{sec:inferring_constraints} for an in-depth discussion.

\begin{figure}[t!]                        
\begin{center} 
\includegraphics[width=.4\columnwidth]{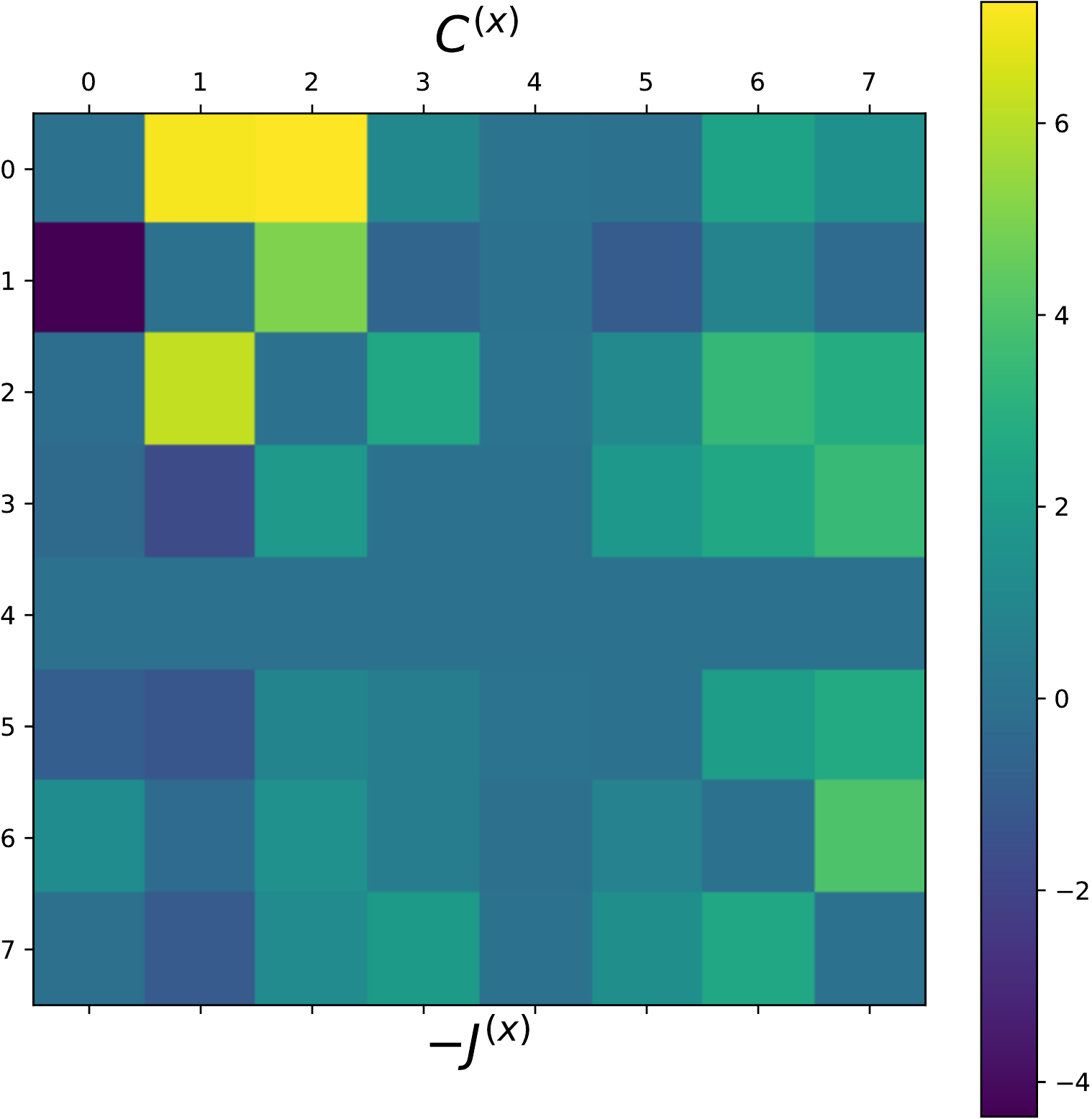}  
\includegraphics[width=.4\columnwidth]{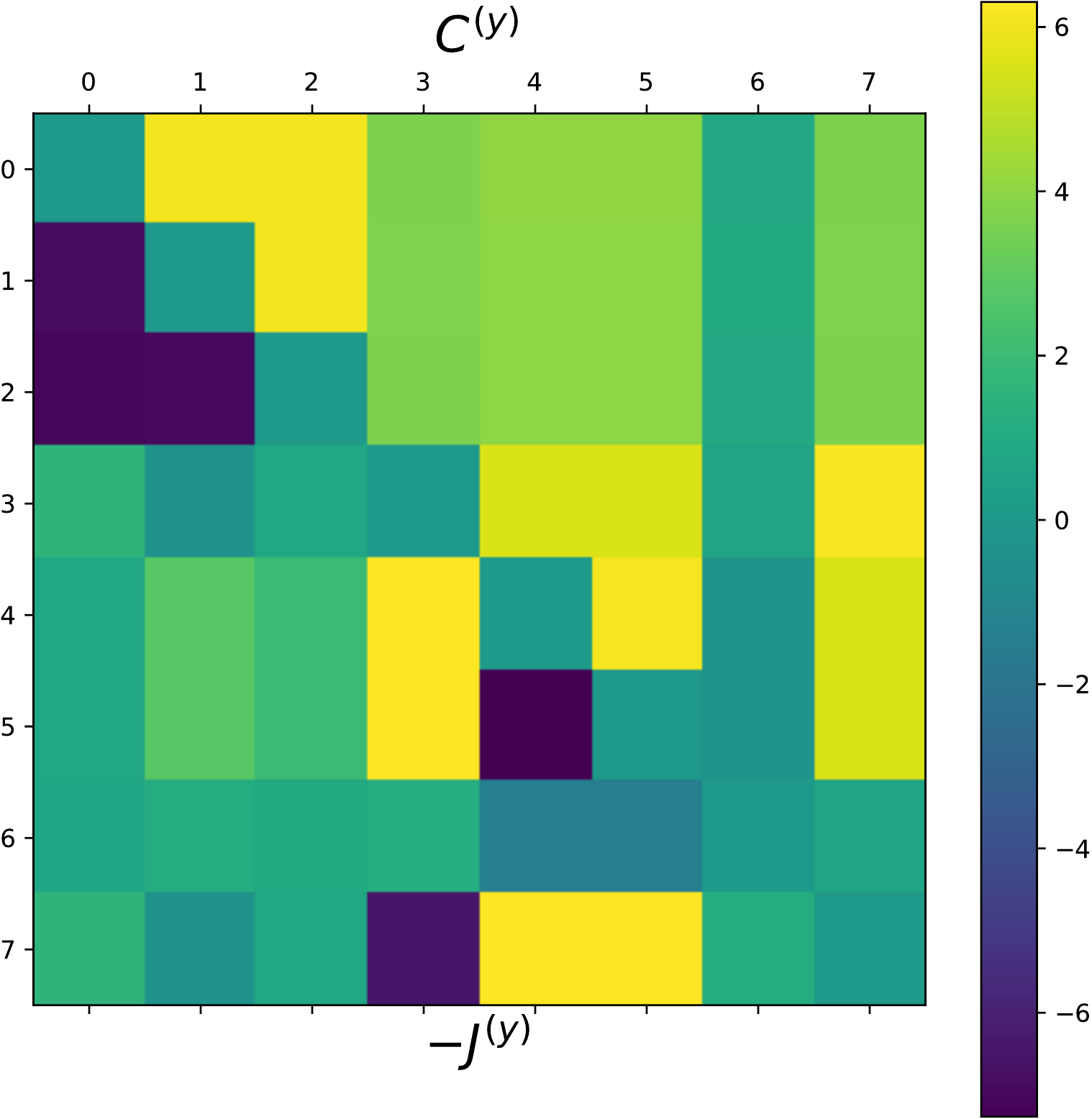}  
\includegraphics[width=.8\columnwidth]{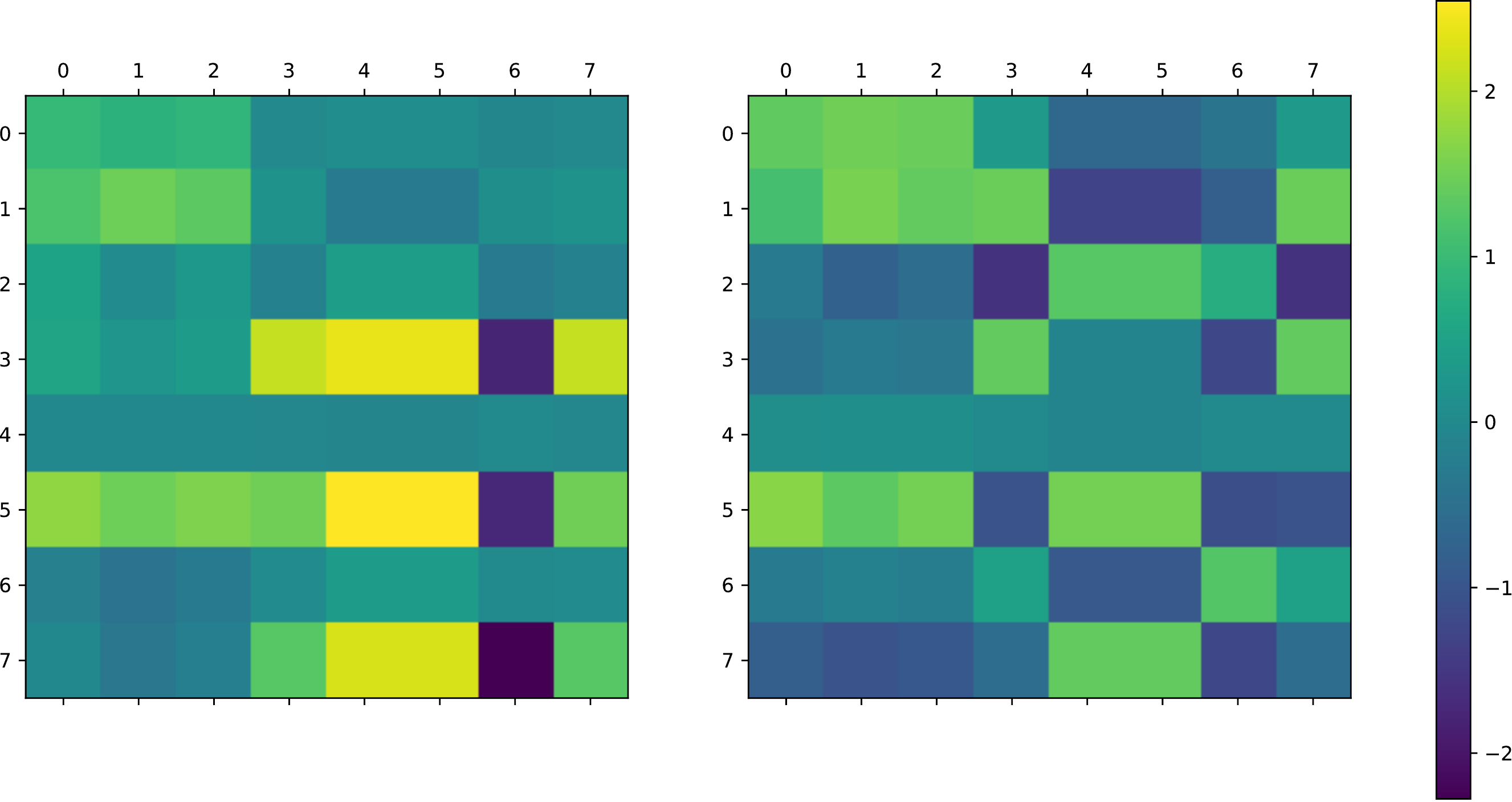}  
	\caption{Comparison between matrices $C$ and $J$. Top left: the $\Cx$ box of matrix $C$ (upper right triangle) versus the $\Jx$ box of matrix $-J$. Top right: idem, but for $\Cy$ and $\Jy$. Bottom left: $\Cxy$. Bottom right: $\Jxy$. Mind that the $J$ matrix is such that negative matrix elements represent ferromagnetic couplings, or affine interactions. As a general trend, matrix $J$ is sparser than matrix $C$, as expected. The matrices $C$ and $J$ exhibit similar matrix elements, except in couples of coordinates involved in the same constraint. In the top row, the diagonal has been set to zero. }
\label{fig:CxJxCyJy}
\end{center}   
\end{figure}

In figure \ref{fig:CxJxCyJy} we present a comparison among the matrices $\Cx$ and $\Jx$, $\Cy$ and $\Jy$, $\Cxy$ and $\Jxy$. As a general observation, the effective matrices are sparser than the correlation matrices, as expected. In particular, while both $\Cx_{6,3}$ and $\Cx_{7,3}$ are statistically significant, only $J_{7,3}$ is (the effective interaction attributes the $6-3$ correlation to the $\Delta_{7,{\sx}}-\Delta_{3,{\sx}}=0$ constraint). The same happens, for instance with $\Cy_{1,3}$ and $\Cy_{0,3}$, statistically significant, while only $J_{0,3}$ is (the $1-3$ correlation is attributed to the $0-1$ constraint). 

We conclude that the effective interaction coupling matrix $J$ provides information beyond the experimental correlations, since it disambiguates the correlations propagated by the constraints, attributing them to the effect of a reduced set of couplings. In section \ref{sec:inferring_constraints} we illustrate the fact that an alternative method of avoiding the constraints, consisting in fitting a dataset in which the redundant variables are eliminated (instead of keeping them and avoiding the influence of the constraint-eigenvectors), may lead to $J$ matrices whose interpretation is misleading.

\subsection{Longitudinal and Torsion interaction strengths}


The $\nl\times\nl$ vertical, horizontal and oblique correlation matrices are defined as the corresponding correlations among landmark fluctuations: $\Cx_{ij}=\<\Delta_{i,\sx}\Delta_{j,\sy}\>$, and the same for $\Cxy$, $\Cy$. The whole $2\nl\times 2\nl$ correlation matrix $C$ is defined as $C_{\mu\nu}=\<\Delta_\mu\Delta_\nu\>$, where the $2\nl$ Greek indices $\mu=i,\c_i$ denote the $\c_i=\sx,\sy$ coordinates of the $i$-th landmark. We define analogously the vertical, horizontal and oblique interaction matrices. The relation among these matrices is given by:

\begin{equation}
C = \matriz{\Cx}{\Cxy}{\Cxy^\dag}{\Cy}, \qquad J = \matriz{\Jx}{\Jxy}{\Jxy^\dag}{\Jy}, \qquad C=J^{-1}
\end{equation}
where the $-1$ power means the pseudo-inverse operation.

In their turn, the longitudinal and torsion interaction matrices, $J^{\parallel}$, $J^\perp$, correspond to the displacements along, and normal to, the segment joining the landmarks $i$ and $j$, called $\hat e_{ij}=\<\vec r_{ij}\>/r_{ij}$, where $\vec{r_{ij}}=\vec r_j-\vec r_i$ and $r_{ij}=|\<\vec r_{ij}\>|$. These are defined so that the matrix elements $J_{ij}^{\parallel}$, $J_{ij}^{\perp}$ are the $\Jx_{ij}$ and $\Jy_{ij}$ couplings, but in a ($ij-$dependent) rotated basis such that the $\sx$-axis coincides with the $i,j$ inter-landmark segment versor, ${\hat e}_{ij}$. Henceforth, the $J^{\parallel}$ and $J^{\perp}$ matrices are not obtained by a rotation of the original matrices $\Jx$ and $\Jy$. Instead, each $J^{\parallel}_{ij}$ element results from a whole inference procedure in a different basis depending on the couple $ij$. In particular, $J_{ij}^{\parallel}=\Jx_{ij}(\hat e_{ij})$, where $\Jx(\hat e_{ij})$ is the inferred matrix obtained from the pseudo-inverse of matrix $C(\hat e_{ij})$ in a coordinate system in which the $\sx$ axis coincides with the $ij$-segment (in other words, $C(\hat e_{ij})={\cal R}_{ij}^{\dag}C {\cal R}_{ij}$, where ${\cal R}_{ij}$ is the 2D rotation matrix by the angle $-\alpha_{ij}$, and the matrix product is over the $\sx$ and $\sy$ blocks of matrix $C$). 

We remark that there is less information in $J_{ij}^{\parallel}$, $J_{ij}^{\perp}$ (for all $i,j$) than in the whole effective interaction matrix $J$ (since the matrix whose matrix elements are $\Jxy_{ij}(\hat e_{ij})$ is not $J^{\parallel}$, nor $J^{\perp}$). 

We now provide a clearer interpretation of the longitudinal and torsion effective interaction matrices.  $J_{ij}^{\parallel}$, $J_{ij}^{\perp}$ capture the relative relevance of the fluctuations around the average distance $\<r_{ij}\>$, and of angle fluctuations around $\alpha_{ij}$, respectively. Large values of $J_{ij}^{\parallel}$ imply that the distance among $i$ and $j$ in the direction of its average axis is highly ``locked'', i.e., it tends to exhibit small fluctuations, from sample to sample, around its most probable value. For instance (see figure 4 in the main text), a small fluctuation $\delta_{4,7}^{\parallel}=|(\vec\Delta_4-\vec\Delta_7)\cdot \hat e_{4,7}|$ of the $4,7$-segment distance with respect to the average face $\DD= {\bf 0}$ implies a large energy increment, $J^{\parallel}_{4,7} {\delta_{4,7}^{\parallel}}^2$ and, consequently, a large decrement of the probability density $\cal L$, proportional to $\exp(-J^{\parallel}_{4,7} {\delta_{4,7}^{\parallel}}^2)$. Conversely, a fluctuation of the $6,7$ segment distance $\delta_{6,7}^{\parallel}$ will give rise to a small or non-significant decrement of the probability of the resulting facial vector since the longitudinal coupling constant $J^{\parallel}_{6,7}$ is small, a fact that highlights the prominent importance of the inter-landmark distance $r_{4,7}$ over $r_{6,7}$ in the process of facial discrimination. In the same way, fluctuations in the transversal components of both segments, $\delta_{4,7}^{\perp}$ and $\delta_{6,7}^{\perp}$ (and consequent fluctuations of the inter-landmark segment angles around $\alpha_{4,7}$ and $\alpha_{6,7}$), have a strong impact in their probability of being sculpted (i.e., in their perceived attractiveness), since both torsion coupling constants $J^{\perp}_{4,7}$ and $J^{\perp}_{6,7}$ are large in absolute value.

\subsection{Dependence of $J$ on inter-landmark distances and angles} A different, interesting aspect of the matrix of effective interactions $J$ is the dependence of the interaction strengths among landmarks $i,j$ as a function of their average distance $\<r_{ij}\>$ and average segment angle $\alpha_{ij}$. We stress that $\<r_{ij}\>$ and $\alpha_{ij}$ are meta-parameters in the sense that that they are not codified in the database $\cal S$ and, hence, are not inferred (the facial vectors $\DD$ are actually fluctuations around the average single-landmark positions). From a cognitive point of view, one would expect that the interaction strengths $|\Jx_{ij}|$, $|\Jy_{ij}|$, $|\Jxy_{ij}|$ among couples of nearby landmarks should tend to be stronger for smaller values of $r_{ij}$ or, at most, that they do not present an increasing trend (which would mean that farther away landmarks influence each other more than closer landmarks). In its turn, if the $ij$ coupling absolute value decreases with $\alpha_{ij}$, this would indicate the prominence of horizontal over vertical inter-landmark segments, and vice versa. 

The data does not allow for sharp conclusions at these regards. However, and although the absolute value of the $J$ matrix elements do not show a clear trend with $r_{ij}$ nor with $\alpha_{ij}$, some interesting information can be retrieved from such analysis. Indeed, a moderate decreasing trend is observed in $|J^{\parallel}_{ij}|$ vs. $r_{ij}$, signifying that nearer landmarks tend to influence each other more than farther away landmarks, but {\it only along the inter-$ij$ landmark segment}, in the sense that only the longitudinal coupling presents such trend. Interestingly, the trend is lost when the $\sx$, $\sy$, $\sx\sy$ components of $J$ are plotted vs. $r_{ij}$. The absence of a clear trend with $\alpha_{ij}$ indicates lack of prominent importance of horizontal versus vertical inter-landmark segments. 

\begin{figure}[t!]                        
\begin{center} 
\includegraphics[width=.8\columnwidth]{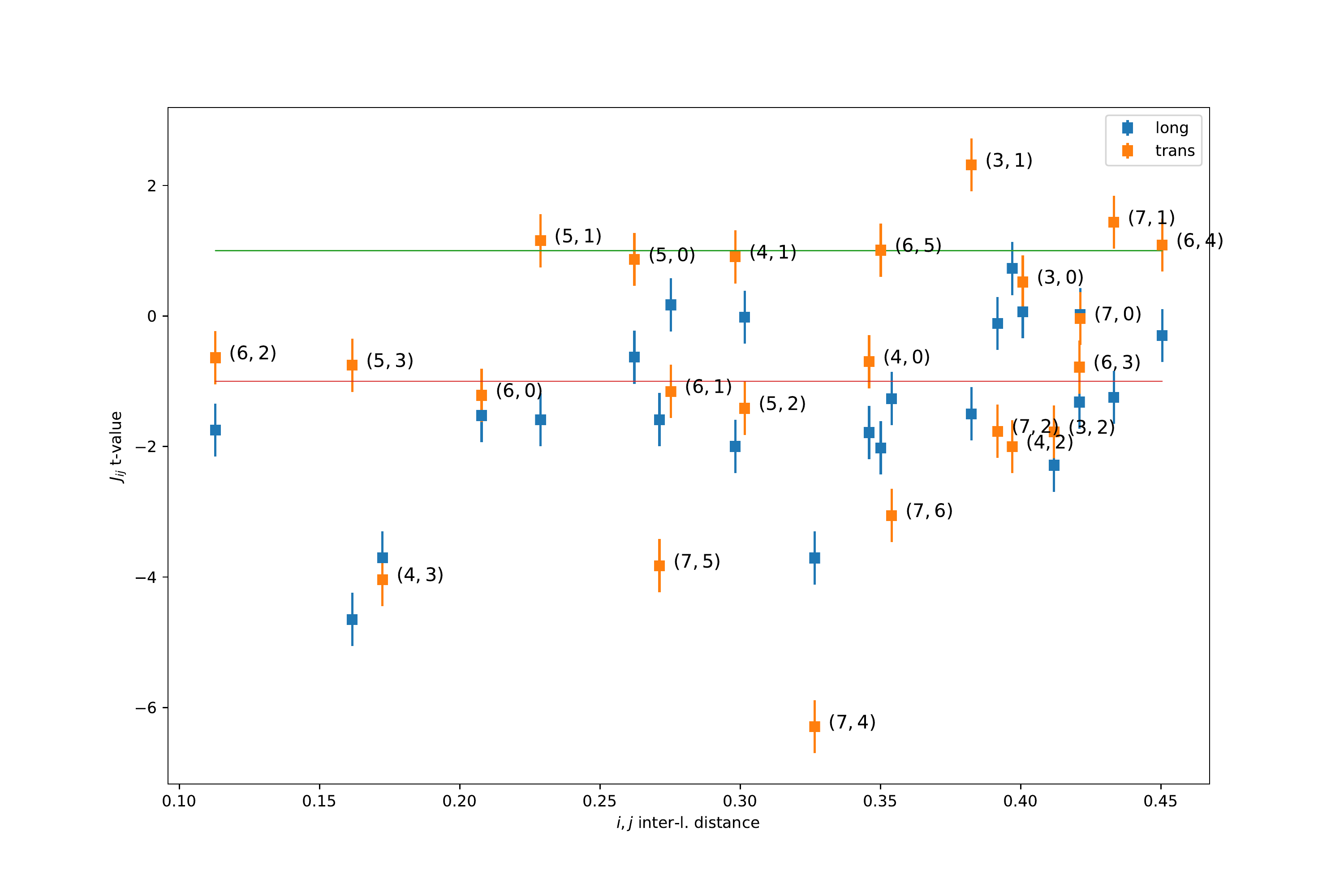}  
	\caption{The t-value corresponding to the matrix elements $J^{\parallel}_{ij}$, $J^{\perp}_{ij}$ versus the inter-landmark average distance $\<r_{ij}\>$. }
\label{fig:Jvsdistance}
\end{center}   
\end{figure}

\begin{figure}[t!]                        
\begin{center} 
\includegraphics[width=.8\columnwidth]{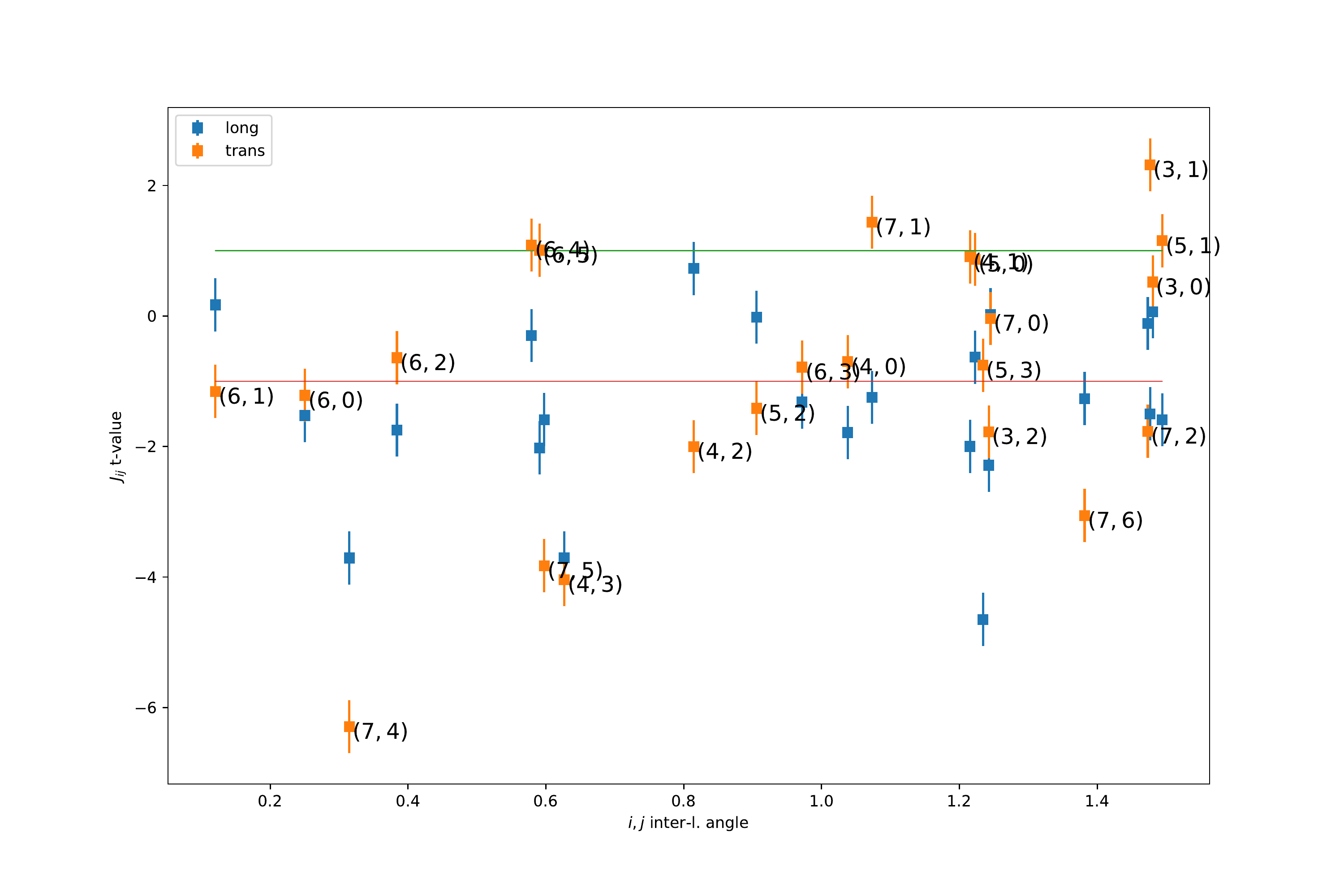}  
	\caption{The t-value corresponding to the matrix elements $J^{\parallel}_{ij}$, $J^{\perp}_{ij}$ versus the angle subtended by $\hat e_{(ij)}$ with the $\sx$-axis, $\alpha_{ij}$. }
\label{fig:Jvsangle}
\end{center}   
\end{figure}

We show in figures \ref{fig:Jvsdistance},\ref{fig:Jvsangle} the quantities $J^{\parallel}_{ij}$, $J^{\perp}_{ij}$ versus $\<r_{ij}\>$ and $\alpha_{ij}$, respectively (see the main article). Although no clear trend is observed, it is apparent a moderate decreasing trend of $|J^{\parallel}_{ij}|$ versus $\<r_{ij}\>$, as referred in the main article, and a slight decreasing trend of $|J^{\perp}_{ij}|$ versus $\alpha_{ij}$. 

We notice that, in the notation of the article, negative values of $J$ indicate the tendency to positive correlations (a ferromagnetic interaction, in the statistical-physical language).

\subsection{The Harmonic inference in the limit $\Cx,\Cy\gg \Cxy$ }

It can be shown that the solution of the inverse problem, at first order in the limit $\Cxy\ll\Cx,\Cy$), is:

 \begin{subequations}
\begin{eqnarray}
\Jx&=&\Cx^{-1}  \\
\Jy&=&\Cy^{-1}  \\
\Jxy&=&-\Jx\Cxy\Jy 
\end{eqnarray}
\label{eq:approxinverseproblem}
 \end{subequations}
and that, in this limit, it is:

\begin{equation}
Z = \frac{(2\pi)^n}{(\det \Jx\,\det \Jy\,\det \Jxy)^{1/2}}
\end{equation}
This can be shown by Gaussian integration, or approximating the inverse of the matrix 

\begin{equation}
J=\matriz{\Jx}{\Jxy}{\Jxy^\dag}{\Jy}
\end{equation}
by using the first-order (in $A$) matrix expansion: $[B(1+A)]^{-1}\simeq (1-A)B^{-1}$, with $B=\matriz{\Jx}{0}{0}{\Jy}$.  

\begin{figure}[t!]                        
\begin{center} 
\includegraphics[width=.8\columnwidth]{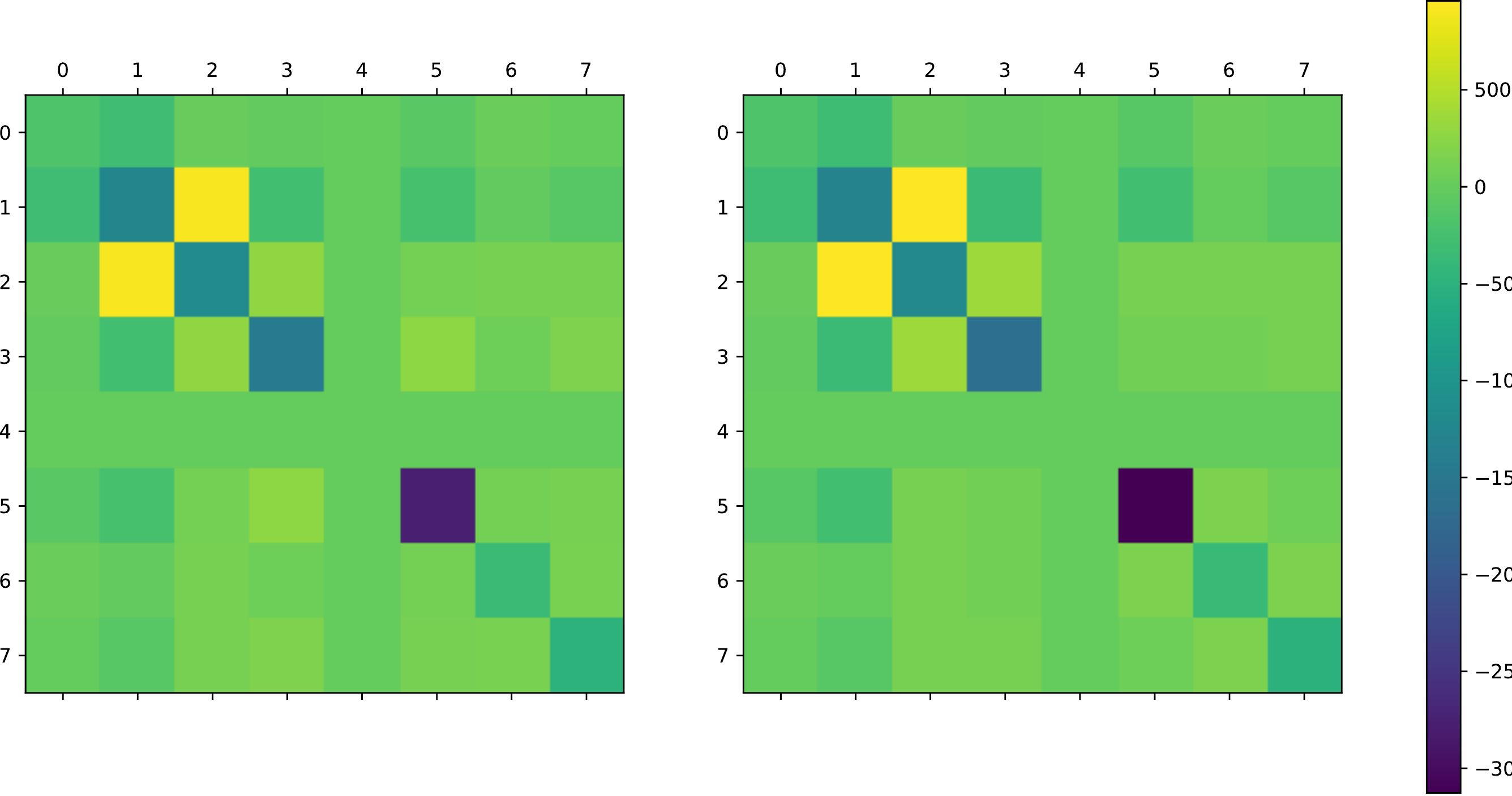}  \\
\includegraphics[width=.8\columnwidth]{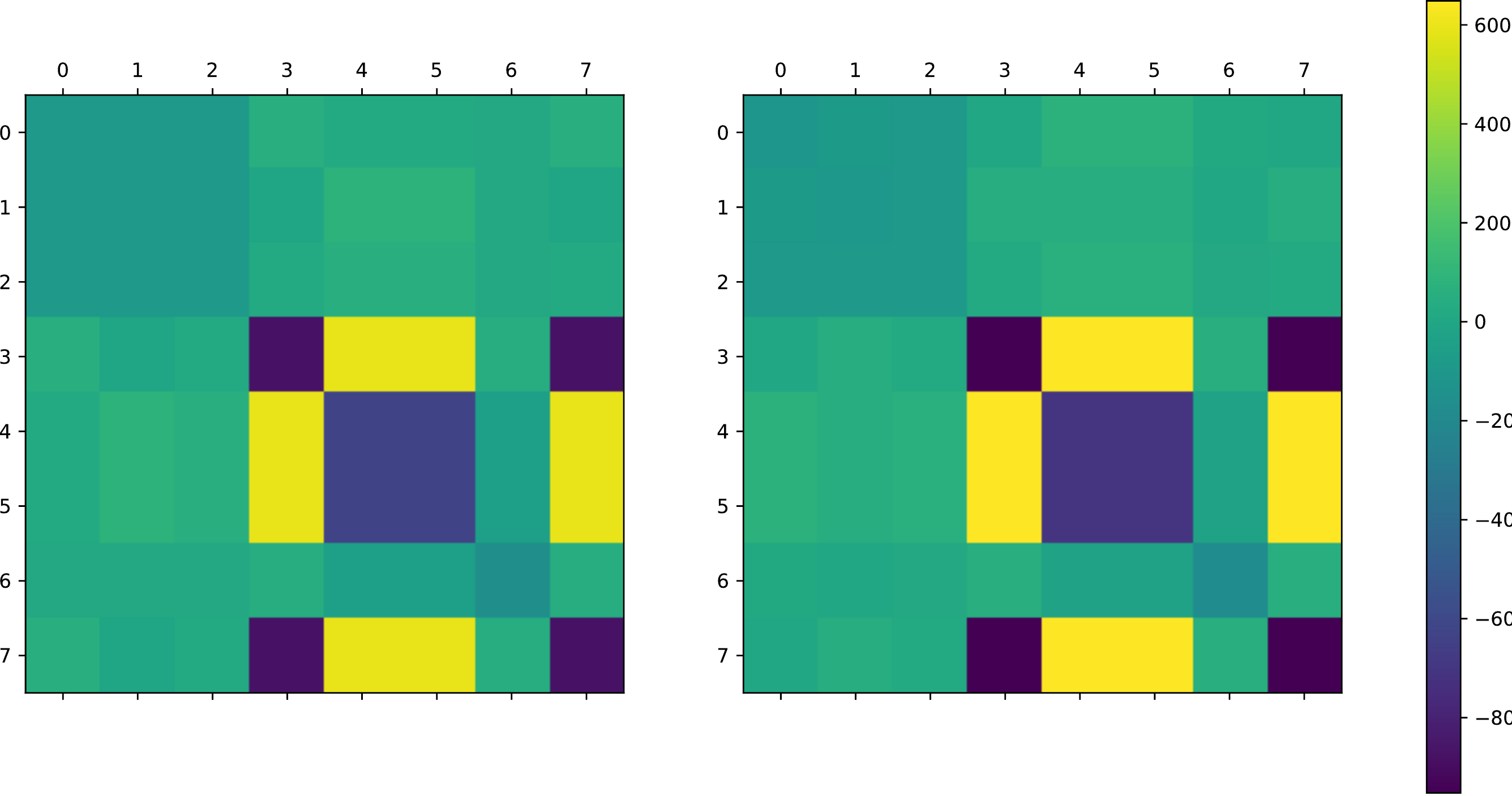}   \\
\includegraphics[width=.8\columnwidth]{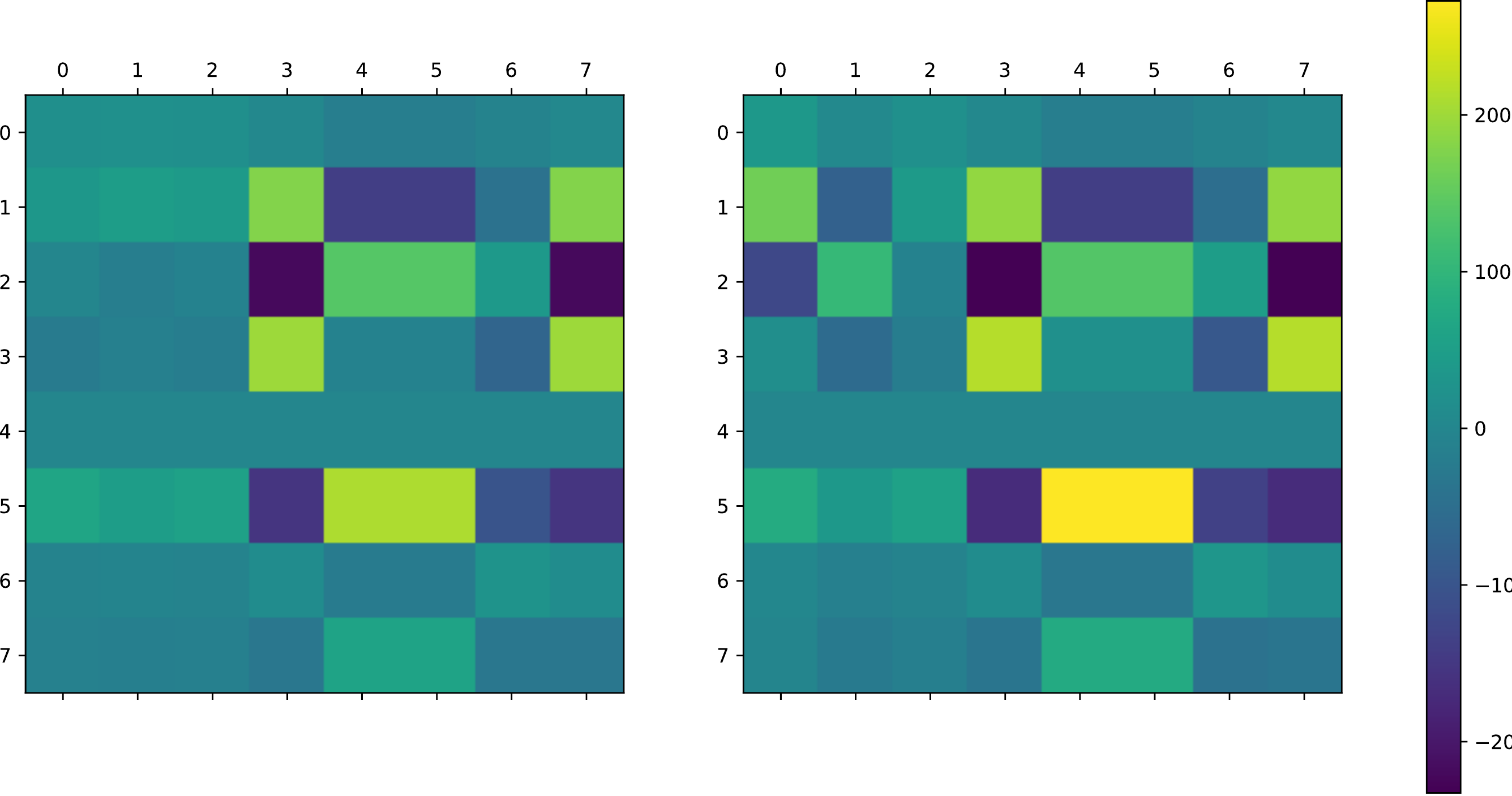}  
	\caption{The comparison between the exact $J=C^{-1}$ and the approximated $J$ computed as in equation \ref{eq:approxinverseproblem}. Left column: approximated $J$. Right column: exact $J$. First, second and third row: $\Jx$, $\Jy$ and $\Jxy$ respectively.}
\label{fig:Japproximation}
\end{center}   
\end{figure}

Indeed, the experimental matrices $\Cx$, $\Cx$ are larger than $\Cxy$. The approximated solution, equation \ref{eq:approxinverseproblem} is, consequently, a rather good approximation. In figure \ref{fig:Japproximation} we show this by comparing the exact $\Jx$, $\Jy$, $\Jxy$ as different blocks of $J=C^{-1}$, versus  the ones resulting from  equation (\ref{eq:approxinverseproblem}).

The relative influence of oblique correlations may be also assessed by defining a simpler model, that we will call the {\it null-$\sx\sy$ model}, consisting in neglecting oblique interaction terms (taking $\Jxy=0$). An even simpler model, that we will call {\it dot model}, consists in neglecting oblique interactions and supposing that the couplings $\Jx$ and $\Jy$ are equal: 

\begin{equation}
H_{\rm dot}=\frac{1}{2}\sum_{i,j} J^{(\rm dot)}_{ij}\,\D_i \cdot \D_j 
\end{equation}

In this case the probability distribution is simply:

\begin{equation}
{P}_{\rm dot}({\bf \Delta}_\sx,{\bf \Delta}_\sy | J^{({\rm dot})})=\frac{(2\pi)^{n}}{\det J^{({\rm dot})}} \exp\left(-H_{({\rm dot})}[{\bf \Delta}_\sx,{\bf \Delta}_\sy]\right)
\label{eq:dot}
\end{equation}
where $J^{({\rm dot})}$ is the inverse matrix of $C^{({\rm dot})}_{ij}=\<\D_i\cdot\D_j\>$.

We have assessed the efficiency of the dot and null-$\sx\sy$ models by evaluating their efficiency in the classification task. As we show in section \ref{sec:classification}, neglecting the oblique correlations (in the null-$\sx\sy$ model) and the anisotropy of vertical/horizontal correlations (in the dot model) leads to a poorer performance. This provides a quantitative assessment of the relative influence of these terms. We conclude that {\it the influence of oblique correlations is crucial, and not negligible, in the facial perception process}.

\subsection{Two ways of inferring with constraints in the database of facial modifications {\label{sec:inferring_constraints}}}

In section \ref{sec:constraints}, we have exposed a method of MaxEnt inference (from pairwise interactions) from a database exhibiting linear constraints. Within this method, all the $D$ components of the vectors are considered, and inferred from, despite they are redundant. The resulting experimental correlation matrix $C$ is singular as it exhibits $D-r$ null eigenvalues, each one corresponding to a constraint. However, the influence of the constraints on the inferred model is subtracted by defining a probability distribution in the subspace of the coordinates that are invariant under the linear operation associated to the constraint. Mathematically, this is done through the pseudo-inverse operation (see eq. \ref{2pointcorrelationtilde}), which discards the subspace expanded by the eigenvectors corresponding with null eigenvalue. The corresponding inferred probability distribution corresponds to a system which is invariant under rescaling of the constraints $c_j$, equation (\ref{eq:symmetry}). 

An alternative method to infer $P$ avoiding the influence of constraints consists in inferring only a subset of $r$ non-redundant, unconstrained variables, in terms of which the correlation matrix has rank equal to $r$. As mentioned before and in the main article, this method may lead to a matrix of effective interactions leading to a less clear {\it interpretation}. The  $J_{ij}$ elements will reflect in this case the influence of the constraints in the considered $r$ variables. Oppositely, with the null-mode subtraction method, the $J$ matrix represents a system which already satisfies the constraint (see section \ref{sec:CvsJ}) and, for this reason, the $J_{ij}$ matrix elements do not reflect its influence on the data. 

An illustration of these concepts is shown in the main article, where we compare matrices $C$ and $J$. The null-mode subtraction method provides a matrix $J$ which is actually sparser than matrix $C$. This does not occur when inferring from a reduced, non-redundant set of variables.

A further, particularly clear illustration is seen in terms of inter-landmark distances ${\bf d}=(d_{i})_{i=0}^{10}$, an alternative parametrization of the facial vectors $\bm \Delta$ (see the precise definition in \cite{ibanez2019} and in figure \ref{fig:key}) in terms of 11 vertical or horizontal distances separating couples of landmarks. The function that maps a vector of inter-landmark distances $\bf d$ into a vector of landmark coordinates $\bm \Delta$ is one-to-one (and depends on some distances of the reference portrait).  The distances $\bf d$ are subject to a constraint, reflecting the scale invariance of the problem \cite{ibanez2019}: $\sum_{i=1}^4 d_i=1$, which signifies that all the distances $d_i$ are in units of the total face length (see figure \ref{fig:key}). This constraint induces a null mode in the correlation matrix. 

\begin{figure}[t!]                        
\begin{center} 
\includegraphics[width=.4\columnwidth]{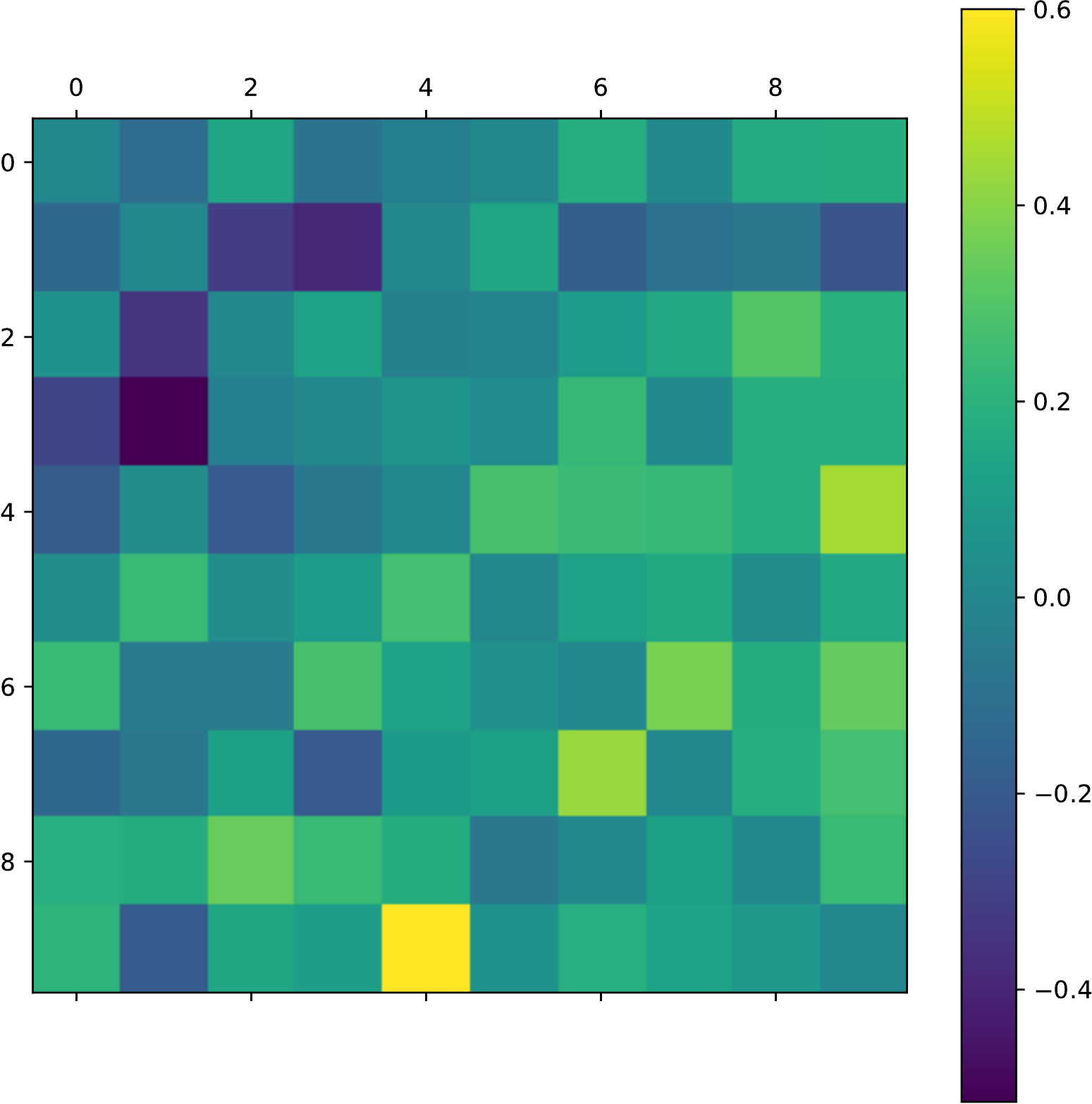}  
\includegraphics[width=.4\columnwidth]{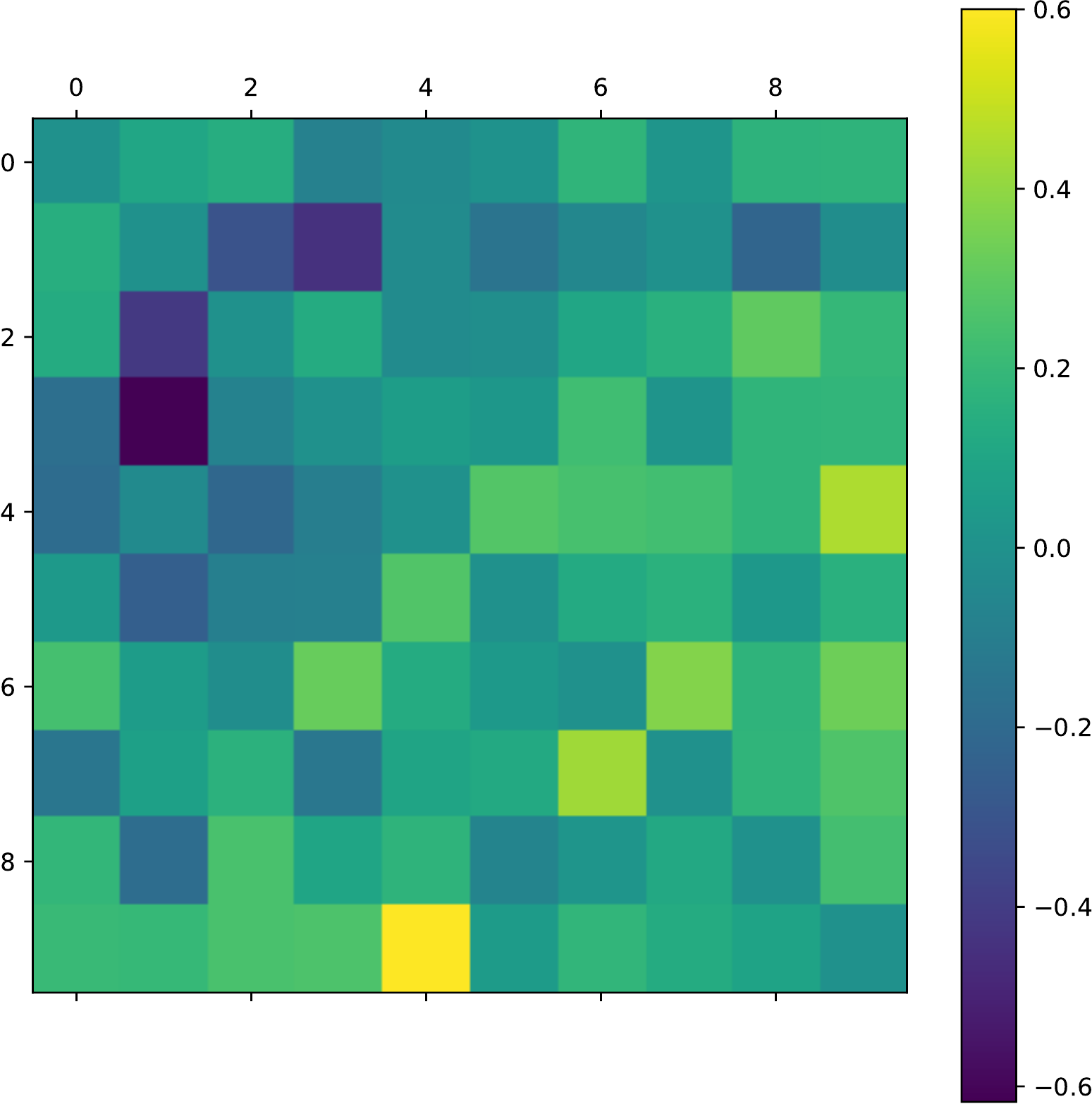}   \\
\includegraphics[width=.4\columnwidth]{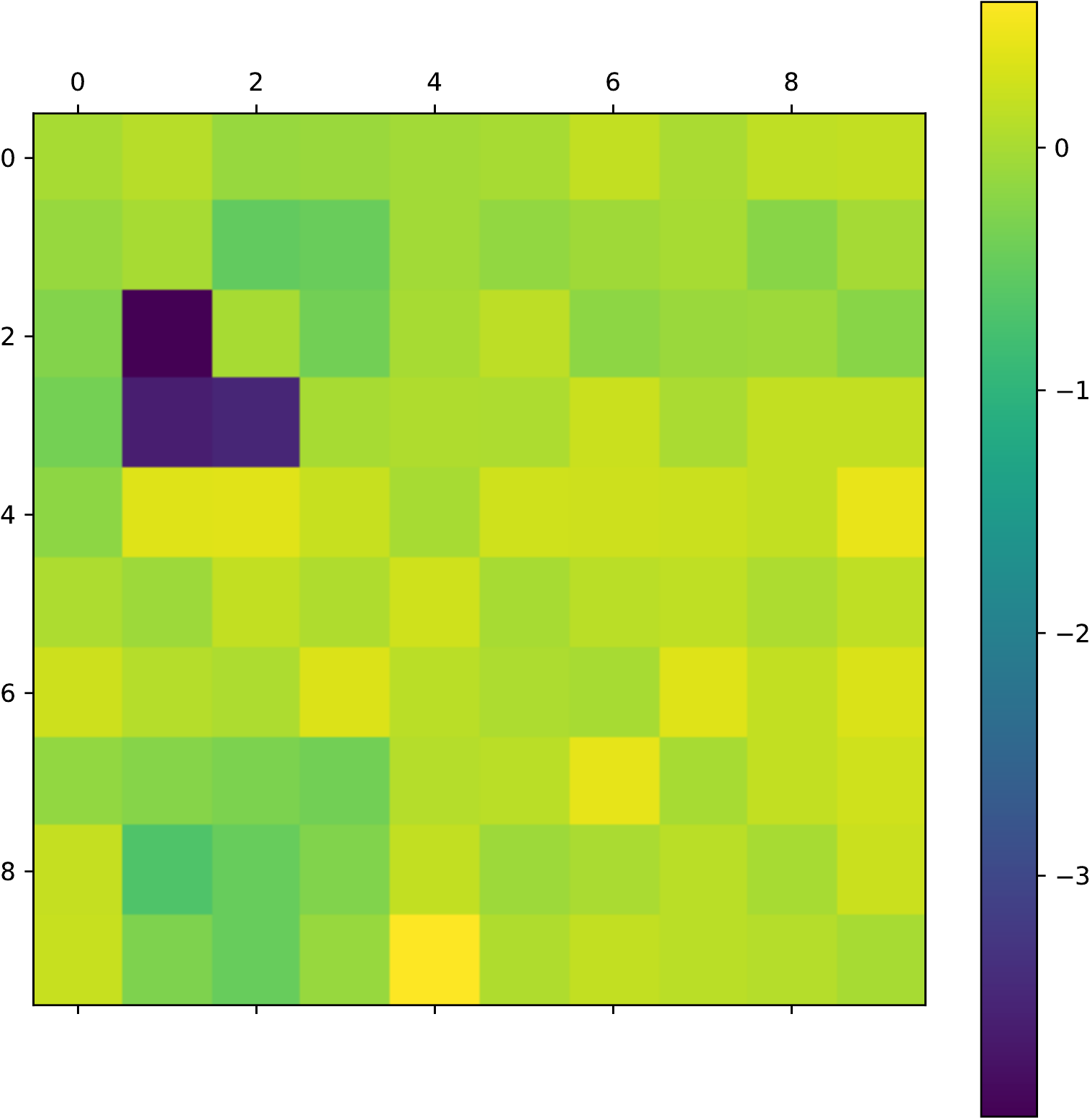}  
\includegraphics[width=.4\columnwidth]{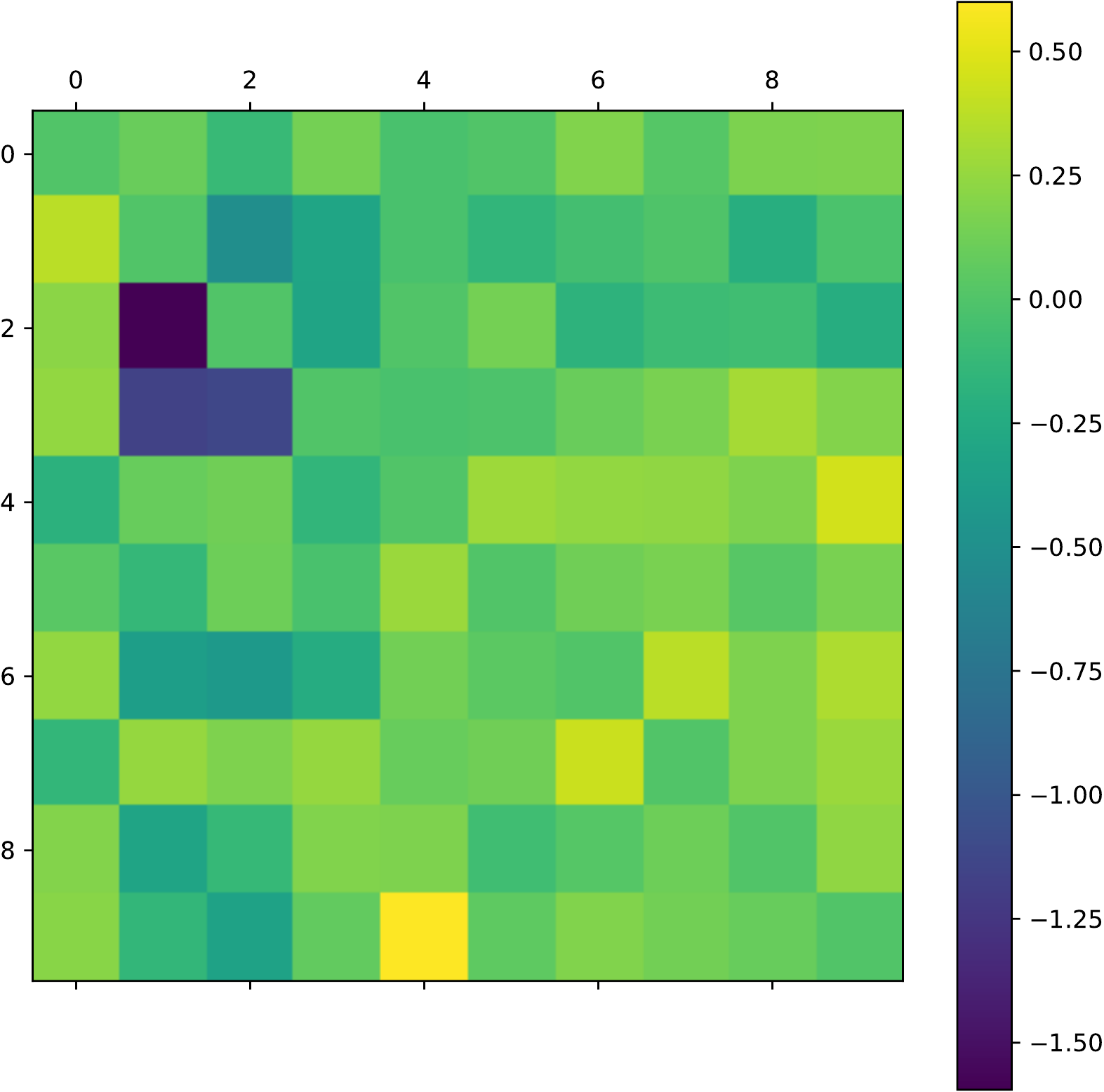}  
	\caption{$\sf t$-value corresponding to the matrices $C^{(-k)}$ (upper right triangle) and $J^{(-k)}$ (bottom left triangle), in terms of inter-landmark distances, and avoiding the $k$-th distance, $d_k$. Top left, top right, bottom left and bottom right correspond, respectively, to $k=1,2,3,4$. We can observe that, for $i,j=1,2,3,4$, all the elements of $J_{ij}$ are positive. The diagonals of all matrices have been set to zero for a clearer comparison.  }
\label{fig:J-k}
\end{center}   
\end{figure}

We now compare the effective interaction matrices corresponding to the two alternative ways of inference discussed before.  We first calculate, see figure \ref{fig:J-k}, the matrices $J^{(-k)}={C^{(-k)}}^{-1}$, the inverse of the $D-1 \times D-1$ correlation matrices $C^{(-k)}_{ij}=\<d_i d_j\>$ in which the $k$-th row and column have been removed, $i,j\ne k$. The matrices $J^{(-k)}$ are presented in figure \ref{fig:J-k} for $k=1,2,3,4$, compared with matrices ${C^{(-k)}}$. 

We observe that the variables involved in the constraint result to be anticorrelated, $C^{(-k)}_{ij}<0$ when both $i,j$ are in the set $1,2,3,4$, a fact fact may be attributed to the presence of the constraint (e.g., vectors with larger distances $d_1$ tend to exhibit lower $d_2$'s, since, for all vectors, $d_1+d_2+d_3+d_4=1$). Indeed, also  $-J^{(-k)}_{ij}<0$ for $1\le i,j \le 4$ and such that $i,j\ne k$ , $i\ne j$: it is necessary an {\it anti-ferromagnetic interaction}, or a statistical tendency of variable $i$ to decrease when variable $j$ increases, in order that the theoretical distribution associated to $J^{(k)}$ describes the statistics of the set of variables. Such statistical tendency is {\bf on the top of other statistical tendencies, of cognitive origin, not related to the constraint}. In other words, the matrices $J^{(-k)}$ describe the data statistics of two different origins: those associated to the constraint, and those of cognitive origin. 

\begin{figure}[t!]                        
\begin{center} 
\includegraphics[width=.4\columnwidth]{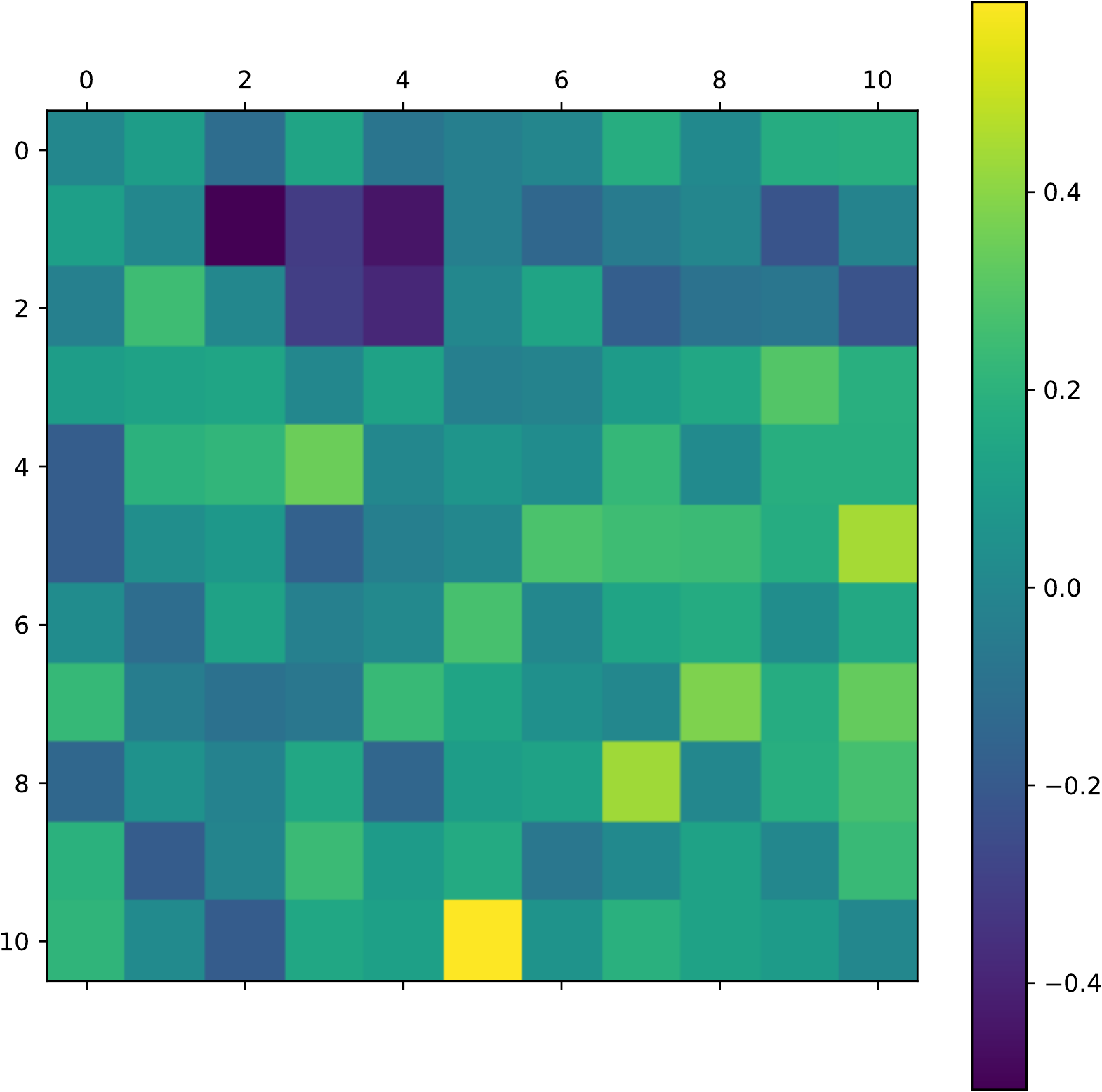}  
\includegraphics[width=.4\columnwidth]{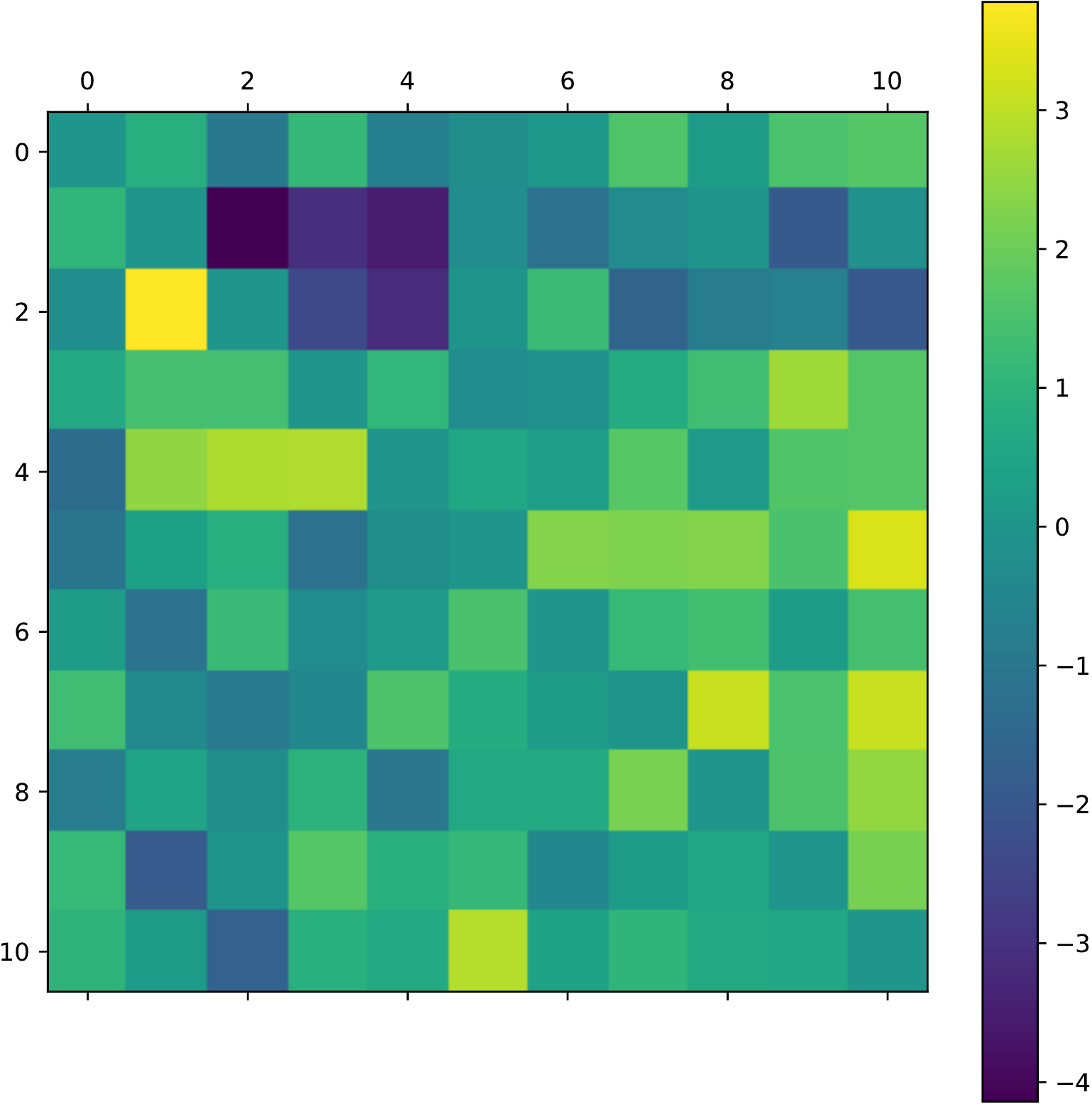}   \\
	\caption{Matrices $C$ (upper right triangle) and $J$ (bottom left triangle) in terms of inter-landmark distances, using the whole, redundant set of distances and the null-mode subtraction method. Left: the matrix elements. Right: their $\sf t$-values. We see, at opposite with figure \ref{fig:J-k}, that $J_{ij}<0$ for $i,j=1,2,3,4$. The diagonals have been set to zero for clearer comparison. }
\label{fig:Jnull}
\end{center}   
\end{figure}

Second, in figure \ref{fig:Jnull} we present the resulting effective interaction with the null-mode subtraction method, using as $J$ the pseudo-inverse of matrix $C$. We observe that, interestingly, all the couplings  $-J_{ij}>0$ when both $i,j$ belong to the set $\{1,2,3,4\}$, implying a {\it ferromagnetic effective interaction} in the physical language, or a {\it positive} tendency of $d_i$ to increase when $d_j$ increases. In the pseudo-inverse case, $J$ represents the statistical effective interaction {\it with the influence of the constraint subtracted}. We learn  information of {\it cognitive}, ``physical'' origin  from $J$, that was veiled in the matrices $J^{(-k)}$'s, influenced by the presence of the constraint. In particular, we learn that the experimental subjects tend to prefer higher eyes, higher $d_1$, in facial modifications with larger noses, larger $d_2$.

In conclusion, the method of null-mode removal that we have used in this work allows, in general, for a more faithful {\it interpretation} of the effective parameters with respect to the alternative method of inferring from a set of non-redundant variables. It is important to stress that the generative model obtained from $J$ and from $J^{(-k)}$ is expected to be equally faithful. In other words, the difference discussed in this section regards but the interpretation of $J_{ij}$ elements, and only in situations in which the parameters actually admit an interpretation, as it is the case in the present and in other problems in biophysics and neurophysics. Indeed, the efficiency of both generative models as a classifier in the two groups of subjects (male/female, ${\cal S}={\cal S}_A\cup {\cal S}_B$), results to be equivalent, see section \ref{sec:classification}.

\subsection{Average proportions and pairwise correlations}

Facial beauty has been related to proportions since the Renaissance \cite{langlois2000,naini2006}, and most modern machine learning studies pose the problem in terms of proportions too \cite{gunes2006,fan2012,chen2014,laurentini2014,grammer2016,shen2016}.

In the main text we have explained that the dataset is faithfully described by a MaxEnt probability distribution ${\cal L}(\x|J,{\bf h})$, whose sufficient statistics is the matrix of pairwise correlations. We have also argued that, for a complete statistical description of the database of facial modifications, a model based on pairwise correlations is not enough. This implies that {\it proportions}, or ratios among facial distances, contains most of the information present in the database, although there is significant information, of cognitive origin, beyond proportions.  We here justify such statement, making notice that the information regarding facial proportions is codified in the matrix of correlations among couples of facial distances.

Consider two facial coordinates, $r_\alpha$, $r_\beta$, referring to the $\sx$ or $\sy$ coordinates of two landmarks, say $i$ and $j$. We will consider their ratio, $r_\alpha/r_\beta$, which is the mathematical expression of a proportion. Calling $\bar r_\alpha=\<r_\alpha\>$ the experimental average value, one has $r_\alpha=\bar r_\alpha + \Delta_\alpha$, by definition of displacement $\Delta_\alpha$. The displacements around the average, $\Delta_\alpha$, are much lower than the averages $\bar r_\alpha$, for all coordinate, $\alpha$ (see \cite{ibanez2019}). This justifies a Taylor expansion of $r_\alpha/r_\beta$ for low $\Delta$'s. Indeed, to the second order in the $\Delta$'s:

\begin{equation}
\frac{r_\alpha}{r_\beta} = \frac{\bar r_\alpha}{\bar r_\beta}\left( 1- \frac{\Delta_\beta}{\bar r_{\beta}} \right) + \frac{\Delta_\alpha}{\bar r_\beta} -\frac{\Delta_\alpha\Delta_\beta}{\bar r_{\beta}} + O\left[\left(\frac{\Delta}{r}\right)^2\right]
\end{equation}

The experimental average of this expression $\<r_\alpha/r_\beta\>$ is, up to an additive constant, equal to $-(1/\bar r_\beta ^2)\<\Delta_\alpha \Delta_\beta\>$ (having used that $\<\Delta_\alpha\>=0$). Hence, the average proportions are completely determined, in the case of small displacements, by the pairwise correlations. 

\subsection{Harmonic interactions and elastic constants }


In the main article we have explained that the 2-MaxEnt model for vectors of facial distance displacements $\bm \Delta$ may be interpreted as the Maxwell-Boltzmann distribution corresponding to a set of particles in 2D interacting through a set of three anisotropic, couple-dependent springs, in the canonical ensemble. We will here justify such statement. We will focus in a couple of landmarks, say $i,j$.  We will call $x_i$, $x_j$ the components of the position of landmarks $i, j$ over two versors in the plane, $\hat e^{(i)}$, $\hat e^{(j)}$, respectively. In other words: $x_i=\vec r_i \cdot {\hat e^{(i)}}$ and the same for $j$, in the notation of the main article. Given the coordinates $x_i$, $x_j$, and if $\hat e^{(i)}=\hat e^{(j)}$, the quantity $\delta_{ij}=x_i-x_j-(\bar x_i - \bar x_j)$ is the change in the distance among $i$ and $j$ with respect to the average vector, and along the common axis $\hat e^{(i)}$. For example, if the  versor is the vertical axis, $\delta_{ij}$ indicates the shift of the vertical distance among landmarks $i,j$  with respect to the average distance among $i,j$. We will define {\it the elastic interaction energy as $(1/2) k_{ij} \delta_{ij}^2$}, which is minimum and equal to zero whenever the distance among $i,j$ is unchanged with respect to the average, $\delta_{ij}=0$, regardless on the single-landmark displacements $x_i$, $x_j$. We make notice that expanding, again, in $\delta_i=x_i-\bar x_i$ to the second order in $\delta_i$ and $\delta_{j}$, it is: $\delta_{ij}^2 = -2\delta_i\delta_j + b + O[\delta^2]$, where $b$ is a constant in $\delta_i$ and $\delta_j$, depending only on $\bar x_i$ and $\bar x_j$.
Henceforth, the elastic interaction energy, ${\cal E}=(1/2) k_{ij} \delta_{ij}^2$ is, up to a constant, and for small fluctuations around the average, $=-2\delta_i\delta_j k_{ij}$, which is the form of the interaction energy in the pairwise Hamiltonian model with the following relation among elastic constant and effective interaction matrix element: $k_{ij}=-(1/2) J_{ij}$. Fixing, for example, $\hat e^{(i)}=\hat e^{(j)}=\hat e_{ij}$, the versor joining the average position of both landmarks, $\hat e_{ij} = \< \vec r_i-\vec r_j \> /|\< \vec r_i-\vec r_j \> |$, we have that $k_{ij}^\parallel = -(1/2) J ^{\parallel}_{ij}$, and idem for the perpendicular components of $\vec r$, $\perp$, and for the vertical, horizontal and oblique components of  $\vec r$.

\subsection{Cognitive origin of non-linear correlations}

In the experiments presented in \cite{ibanez2019}, the subject sculpts her/his ideal facial modification through the interaction with a software called FACEXPLORE, based on genetic and image deformation algorithms. The sculpture process consists in a sequence of multiple left/right choices among couples of facial images, eventually leading to an estimation of the ideal modification according to the subject. Actually, the genetic algorithm performs the {\bf recombination} and {\bf mutation} steps, while the single experimental subject actually plays the role of the {\bf selection} step, through her/his choices. 

The genetic algorithm used (called Differential Evolution) processes different coordinates independently (see the SI of \cite{ibanez2019}). The only correlation among coordinates is expected to be induced by the {\bf selection} process, performed by the human subject. As a consequence, one should expect that the only origin of correlations among coordinates in the populations sculpted by subjects (by the same or by different subjects) are of cognitive nature. 

In fact, this is not the case: {\it part} of the correlations that one observes experimentally are due to an artifact of the algorithm. In a null-model experiment with a {\it random} sequence of left/right choices, the resulting database exhibits significant non-linear correlations of order $2$ and $3$ among facial coordinates. The correlations of order three, $\<\Delta_i\Delta_j\Delta_k\>$, are statistically compatible with the 3-order correlations observed in the human experiment \cite{ibanez2019}. 

The solution of this paradox is that, while the genetic algorithm does not introduces correlations in the recombination and mutation steps, it actually may  amplify the correlations among facial coordinates which are present in the initial condition of the null-model genetic population of facial vectors. Such initial populations are trivially correlated, since some constraints were imposed in the definition of the face space: mainly $\sum_{i=1}^4 d_i=1$ and $d_{10}<d_5$ (see section \ref{sec:inferring_constraints} and figure \ref{fig:key}).

In reference \cite{ibanez2019} we proposed a method to ``subtract'' the influence of the {\it a priori}, non-cognitive or artifact correlations present in the null model experiment, from the cognitive true correlations that we observed. The method revealed that the artifact {\bf pairwise} correlations did not have a significant impact in the results. We suspect that correlations of higher order may be, instead, significantly influenced by the artifact effect. 

In the main article text, we have explained that non-linear inference algorithms allow for a much better classification of the database according to the gender of the experimental subject. This fact implies that, quite interestingly, the differences between facial vectors sculpted by males and by females is encoded in non-Gaussian correlations, beyond proportions ($p=2$), beyond triplets and perhaps quadruplets of facial distances. In the main article, we have also explained that this {\bf may imply} that such differences codified in non-Gaussian correlations are of cognitive order, i.e., that male and female subjects' do prefer facial variations differing in non-Gaussian correlations and, in particular, that humans evaluate quantities that are much more complicated than proportions, when forming an impression about a face. The fact that the introduction of non-linearity helps in a gender classification task, which reflects real and well-known cognitive differences, may suggest so.

An alternative explanation is that the distinguishable differences among male and female preferred facial variations are all codified in pairwise effective interactions only (say, roughly speaking, that males and females differ only in the $J$ matrix, if it could be measured without bias). The non-linear interactions would turn anyway relevant for the classification, since the correlations propagated by the genetic algorithm are coupled to the ones induced by the subject: subjects differing only in $J$ would {\bf also} induce, by means of the artifact, differences in the correlations of higher order.  

Further experiments are needed to clarify this issue. 

\subsection{Generality of the MaxEnt models}

Crucially, the two models of unsupervised inference presented in the main text exhibit a wide generality, going beyond the particular database that we infer in this work. (1) First, the inferred set of facial vectors may be composed by facial images selected according to any criterion: selected by a pool of subjects among real facial images or by a single individual (in this case the distribution $\L(\cdot|\T)$ would probabilistically characterise the single subject's preferred region in face-space); selected according to a criterion different from attractiveness; even not having been selected by subjects but chosen according to some objective criterion as age or gender ($\L(\f|\T)$ would hence represent the probability that a facial image characterised by the facial vector $\f$ presents the desired feature). (2) Second, they can be used to infer any other database of images characterised by the geometric positions of facial (or, in general body) landmarks. (3) Third, these models may be immediately extended to process also non-geometric degrees of freedom (treating the texture and geometric degrees of freedom on the same footing \cite{chang2017}). 

\subsection{Learning in the non-linear MaxEnt model.}

The 3-MaxEnt model parameters are $n_{\rm p} = D+D(D+1)/2+D(D-1)(D-2)/6$ independent components of the interaction tensors $\T=({\bf h},J,Q)$ of order 1,2,3. Their value in the article is fixed by Maximum Likelihood, i.e., by the maximisation of the joint database likelihood, as in equation (\ref{eq:generalizedentropy}). In the case of the 3-MaxEnt model, we have estimated numerically the maximum likelihood value of the parameters $\T^*$ by means of a numerical maximisation of the joint database likelihood by deterministic gradient ascent, using an algorithm that will be presented in a dedicated publication. 

A discrete sequence of interaction tensors $\T(t)$ are recursively updated according to a deterministic gradient ascent rule: $\T(t+1) = \T(t) + \eta_{\T}\, \partial_{\T}\left[ \ln {\cal L}( {\cal S} | \T )\right]_{\T(t)}$, using a learning rate $\eta_{\T}$ depending on the tensor that is being update. We use $\eta_J=10^{-2}$ for the matrices and $\eta_{Q}=10^{-3}$ for the 3-order tensors. The rule is iterated until a quasi-stationary state of the associated test-set joint likelihood is achieved within its statistical error. In particular, the stopping criterion is that the joint test database likelihood increment is lower than $1\%$ in an interval of 10  epochs (an epoch is an iteration of all the tensor elements). As initial condition of the learning dynamics, we choose ${\bf h}={\bf 0}$, $J=\mathbb{I}_D$, $Q=0$.

At a given epoch, $t$, of the gradient ascent iteration, the gradient of the joint likelihood with respect to the effective interaction components involves a theoretical correlator (of order 1,2 or 3) according to the current value of the couplings. For instance: $\left[\partial_{J_{\alpha\beta}} \ln {\cal L}( {\cal S} | \T )\right]_{\T(t)}=\<\Delta_\alpha\Delta_\beta\>_{{\cal L}({\cdot}|\T(t))} -\<\Delta_\alpha\Delta_\beta\>$. Such theoretical correlator is in its turn estimated by means of a Markov Chain Monte Carlo (MCMC) Metropolis algorithm for the sampling of configurations from the theoretical distribution at the corresponding epoch, ${\cal L}(\cdot,\T(t))$. For such MCMC algorithm, we use a number of sweeps $=10^6$ in each epoch. The MCMC vectors $\DD$ are initialised as normal variables with variance equal to their empirical variance, and the Metropolis trials are chosen uniformly in the interval $\Delta_\alpha\in [-5\sigma_\alpha,5\sigma_\alpha]$, where $\sigma_\alpha$ is the empirical standard deviation of $\Delta_\alpha$. Finally, the evaluation of the joint database likelihood in every step of the gradient ascent maximisation algorithm requires the evaluation of the partition function $Z_3$. This is performed by means of a Mayer expansion of the energy around the Gaussian model, to the first order in  $J^{(3)}/J^{(2)}$.

\subsection{Learning the database with the Gaussian Restricted Boltzmann Machine}

\def\*#1{\bm{#1}}   
\def\<{\langle}
\def\>{\rangle}
\def\H{N_{\rm h}}
\def\V{N_{\rm v}}

{\bf Definition of the model.} The Gaussian Restricted Boltzmann Machine (GRBM) is a type of generative stochastic two-layered Artificial Neural Network \cite{Smolensky,Fischer_RBM,Hinton_1985,HintonPracticalGuide}. It is a generalisation of the Restricted Boltzmann Machine (RBM) model \cite{mehta2019}, that learns a probabilistic generative model for real-valued vectors: the visible neurons in the input layer, $\*v$, assume real values. The value of the hidden neurons $\*h$ is, instead, binary, $h_j=0,1$. The state of the $\V$ visible $\*v=(v_i)_{i=1}^{\V}$ and $\H$ hidden $\*h=(h_i)_{i=1}^{\H}$ neurons  is described by an energy-based probability density: 

\begin{equation}
	p(\*v, \*h|\*\theta) =  \frac{1}{Z_{\*\theta}} e^{-E(\*v, \*h|\*\theta)}
\end{equation}
in terms of the parameters $\*\theta =\{ W, \*b, \*c, \*\sigma \}$, to be inferred in the learning process. $W$ is a real ${\V\times \H}$ matrix coupling real and visible variables, while $\*c$, $\*\sigma$ are $\V$-dimensional real vectors representing the bias over the visible neurons and their standard deviation, respectively, while $\*b$ is a real $\H$-dimensional vector representing the bias over hidden neurons. ${Z_{\*\theta}}$ is a normalising constant, depending on the parameters. The function energy $E$ is defined so that the conditional probability distribution $p(\*v|\*h,\*\theta)$ results to be a normal, independent distribution over visible variables. It assumes the form:

\begin{align}
		E(\*v, \*h) = - \sum_{i=1}^{\V} \sum_{a=1}^{\H} \frac{W_{ia} v_i h_a}{\sigma_i^2} + \sum_{i=1}^{\V} \frac{\big(v_i - c_i\big)^2}{2\sigma_i^2} -  \sum_{a=1}^{\H} h_a b_a 
\end{align}

The GRBM probabilistic generative model is obtained through a marginalisation of the hidden variables: $p(\*v|\T)=\sum_{\*h} p(\*v,\*h|\*\theta)$. This model (as far as the hidden neurons are binary) is known to induce non-linear interactions among the visible variables, up to order $p=\V$ in the most general case \cite{cossu2019}.

\begin{figure}[h]
\centering
\includegraphics[width=.8\columnwidth]
	{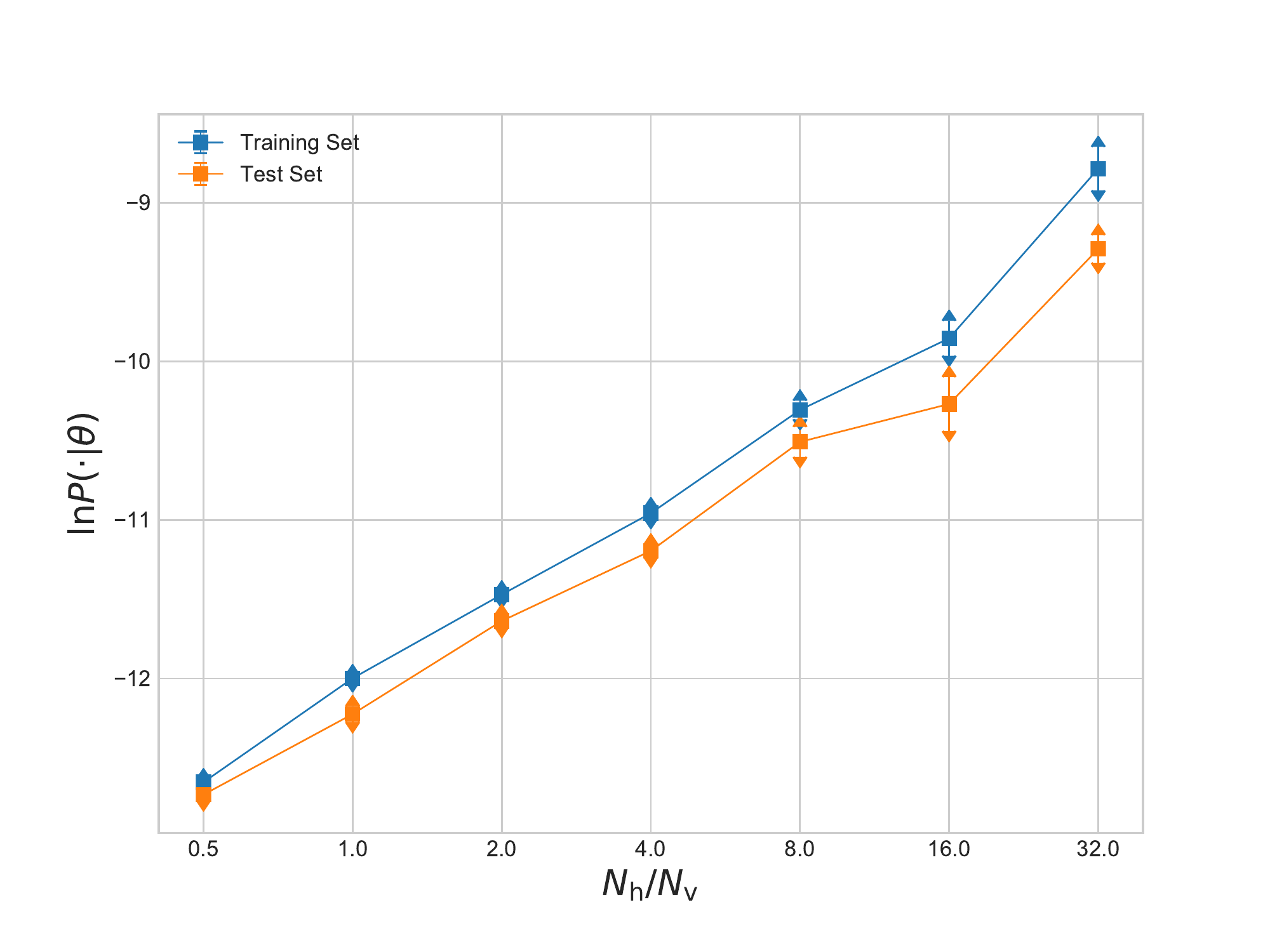}
\caption{Training and test-set log-likelihood as a function of the ratio $\H/\V$, each point has been obtained averaging over 5 realizations of the learning, once the stationary state of the train log-likelihood has been achieved.}
	\label{LLvsNh}
\end{figure}

{\bf Learning protocol.} We have trained the model over a set of redundant or non-redundant data, obtaining equivalent results. As a learning algorithm we have used gradient ascent through persistent contrastive divergence with ${\sf k}=1$ Monte Carlo step, along with mini-batch learning with batch size $B$ \cite{Fischer_RBM}, and an epoch-depending variable learning rate, $\eta$, increasing linearly with the number of epochs. We have set the value of the learning hyperparameters to: number of steps $n_{s} = 2\cdot 10^5$, batch size $B = 200$, momentum $\mu = 0$, initial learning rate $\eta_{0} = 2\cdot 10^{-3}$. The learning rate slope is set such that $\eta_{n_{s}} = 2\cdot 10^{-5}$. The parameters $W$, $\*b$ and $\*c$ are initialized following a standard procedure \cite{MELCHIOR_GRBM_Natural_images, Melchior_Thesis}:

 \begin{subequations}
\begin{align}
	W_{ia} = \chi_{ia} \sqrt{\frac{6}{N_V + \H}}  \ , \quad \chi_{ia} \in (-1,1) \quad \forall i,a
	\\
	b_a = - \frac{1}{2}( || \*W_{*,a} + \*c || + || \*c || ) + \log(0.1) \quad \forall a
	\\
	c_i = 0 \quad \forall i
	\\
	\sigma_i = 1/2 \quad \forall i
\end{align}
\end{subequations}
As equilibration test we have verified that the test-set joint likelihood is  stationary as a function of the number of epochs, within its associated standard deviation. 
We have performed an assessment of the algorithm efficiency as a function of the number of hidden neurons, $\H$. As shown in figure \ref{LLvsNh}, both the test and training-set joint likelihood exhibit a monotonous increasing behaviour vs $\H$, showing no sign of severe overfitting. The auROC score saturates at its maximum value for values $\H\gtrsim 8 \V$, with $\V=10$, confirming this picture. We have consequently considered, for the analysis performed in the main article, $\H=100$. 


Before GRBM learning, the data has been pre-processed eliminating $6$ redundant coordinates, subtracting the average (of the whole database, not of the $\{A,B\}\times\{{\rm train},{\rm test}\}$ datasets separately) and standardising, or dividing each vector component-wise by the vector of standard deviations along the whole set. We have learned the dataset with a variable number of hidden neurons $N_{\rm h}$, from  $N_{\rm h}=D$ to  $N_{\rm h}=16\,D$. 

Afterwards, for the sake of the classification, the model has been trained over the $A,B$ training databases separately, leading to two sets of parameters $\T_A$, $\T_B$ and, consequently, to a likelihood function ${\cal L}_{\rm RBM}(\cdot|\T_{A,B})$. Afterwords, the score ${\sf s}(\tilde{\bf r})=\ln \L(\tilde{\bf r}|\T_A)-\ln \L(\tilde {\bf r}|\T_B)$ is defined for every standardised and non-redundant vector $\tilde{\bf r}$ of the $A,B$ test-sets. Such score is used to construct the ROC curve and scores shown in the main article.

\subsection{Classification with the Random Forest algorithm}

In the random forest classification presented in the main article and in figure \ref{fig:ROCcurve}, we have used the Random Forest Classifier \cite{breiman2001,scikitlearn}, using $1000$ trees created from bootstrapped sub-sample and with nodes expanded until all leaves are pure. As an assessment of the single-split quality we have considered the Gini function. The number of random features considered in the best split choice is equal to ${\sf int}( {\sf sqrt}(D) )$, where $D=16$ is the number of features.

\subsection{ Detailed comparison among several classification methods {\label{sec:classification}}}

\begin{figure}[t!]                        
\begin{center} 
\includegraphics[width=.8\columnwidth]{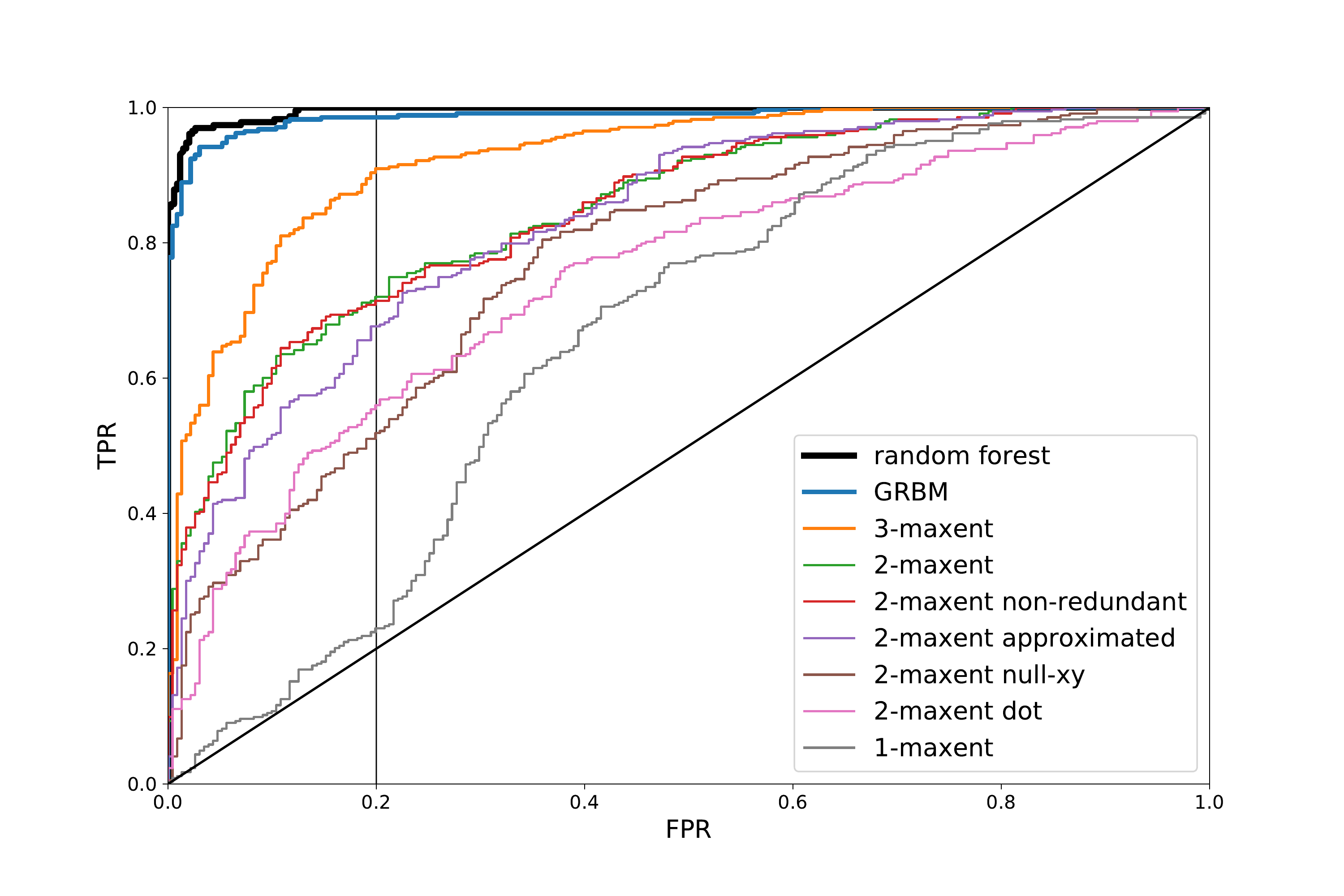}  
	\caption{ROC curves of all the models defined in the text. The order of the model in the legend (and in table \ref{tab:aurocs}) is also the order with which the corresponding curve crosses the vertical line at FPR=0.2.}
\label{fig:ROCcurve}
\end{center}   
\end{figure}

We now present a more detailed analysis of all the classification algorithms that we have considered for the classification of the database according to the subjects' gender. In table \ref{tab:aurocs} we present a systematic comparison of the auROC value \cite{murphy2012}, a standard estimator of the classification accuracy (the area under the corresponding ROC curves in figure \ref{fig:ROCcurve}), associated to the classification according the various algorithms. In particular, {\it 2-MaxEnt approximated} is the 2-Maxent model resulting from the approximation in equation (\ref{eq:approxinverseproblem}); {\it 2-MaxEnt null-${\sx\sy}$} is the model consisting neglecting the oblique interactions, $\Jxy=0$; {\it 2-MaxEnt dot} is defined in equation (\ref{eq:dot}); {\it 1-MaxEnt dot} is defined by inferring the external fields only (and taking the interaction matrix $J$, required for the normalisation of $P$, as a diagonal matrix whose diagonal is equal to the inverse variance of each variable).

The results of table \ref{tab:aurocs} and of figure \ref{fig:ROCcurve} confirm the picture presented in the main article. The value of the single facial distances (in units of the facial length) are not enough for an accurate description of the database of facial modifications. The introduction of pairwise effective interactions, which explain proportions, or ratios of facial coordinates,  induces a notable improvement in the statistical description. Moreover,  {\it oblique} effective interactions (coupling the $\sx$ coordinate of one landmark with the $\sy$ coordinate of another landmark) result a fundamental ingredient. Finally, a crucial role, at least for the sake of the classification according to the subjects' gender, is plaid by effective interactions of higher order: $p=3$ (3-MaxEnt) and $p>3$ (GRBM and random forest). We conclude that the classification is a valid method for the assessment of the assessment of the relative relevance of the various terms. 

Remarkably, an as we anticipated in section \ref{sec:inferring_constraints}, the algorithm used to avoid the constraints (inferring from a reduced, non-redundant set of variables, or using the null-mode subtraction method) do not change the efficiency of the classification. Indeed, the model that we call {\it 2-MaxEnt non-redundant} in table \ref{tab:aurocs} and in figure \ref{fig:ROCcurve} is identical to {\it 2-MaxEnt} but in terms of a subset of $10$ non-redundant variables. Its auROC estimator and ROC curve are statistically distinguishable from  {\it 2-MaxEnt} (with $16$ variables and null-mode subtraction).

\begin{table}
\begin{center}
\caption{\label{tab:aurocs}}
\begin{tabular}{ll}
 algorithm & auROC \\
\hline
	random forest & 0.995  \\
	GRBM & 0.988 \\
	3-MaxEnt & 0.930  \\
	2-MaxEnt & 0.848  \\
	2-MaxEnt non-redundant & 0.846  \\
	2-MaxEnt approximated & 0.830 \\
	2-MaxEnt null-${\sx\sy}$ & 0.770 \\
	2-MaxEnt dot & 0.745  \\
	1-MaxEnt & 0.654 \\
\end{tabular}
\end{center}
\end{table}

\subsection{Inter- and intra-subject correlations and errors.} The set of facial vectors sculpted by a single subject, $\{{\bf r}^{(v,i)}\}_{i=1}^{\cal N}$, are not the result of independent sculpting experiments. They are, rather, correlated as far as they are the outcome genetic {\it population} of facial vectors that evolved according to a stochastic evolutionary algorithm coupled to a sequence of choices performed by the experimental subject \cite{ibanez2019}.  Consequently, it is crucial to subtract the effect of intra-subject (or intra-genetic population) correlations among facial vector components from the inter-subject correlations. On the one hand, one may define the {\it bare correlation} matrix, accounting from both sources of correlations, defined by summing over both subject and population indices: $C_{\alpha\beta}=\<\Delta_{\alpha}\Delta_{\beta}\>$. On the other hand, the {\it inter-subject correlation matrix} accounts only for the inter-subject correlation, and is defined as $\bar C_{\alpha\beta}=\overline{(1/\Ns)\sum_{v'=1}^{\Ns} \Delta_\alpha^{v(v'),i(v')}\Delta_\beta^{v(v'),i(v')}}$, where $v(v')$ and $i(v')$ are random indices in the sets $1,\ldots,\Ns$ and $1,\dots,{\cal N}$ respectively, uncorrelated among them and on $v'$, and the overline $\overline{\cdot}$ means an average over a sufficiently high number of realisations of the set of indices $v(v')$, $i(v')$ for $v'=1,\ldots,\Ns$. The statistical uncertainty associated to the inter-subject correlation, $\sigma_{C_{\alpha\beta}}$, is the standard deviation of the overline argument under many realisations of the set of indices (in other words, a Bootstrap error using only one vector for subject in each Bootstrap sampling,  see the SI of ref. \cite{ibanez2019}). Consequently, the error associated to the inter-subject correlation is of order $\sim \Ns^{-1/2}$, and not of order $\sim S^{-1/2}$ as that of the bare correlation matrix. Analogously, we also define inter-subject and bare $3$-component correlations.

If the inferred model should describe the probability of a given facial vector to have been selected by any subject in the database, then it should be committed to reproduce by construction the inter-subject (not the {\it bare}) correlations. Otherwise, the probabilistic generative models may also simply describe the whole set of facial vectors in the \cite{ibanez2019} experiments, hence accounting also for the intra-subject correlations; the corresponding MaxEnt models would reproduce by construction the $2$ or $3$ bare correlations in this case. 
In our data analysis software one can specify whether the $2,3$ MaxEnt inferred model reproduce {\it bare} or {\it inter-subject} correlations. In this article, some results (the reproduction of angle histograms and the analysis of $J$ matrices) correspond to the inter-subject inference models. The classification tests have, instead, been done with the bare models. For the sake of classification, we have simply tested the ability of the algorithm to capture any useful correlation, regardless of its origin, cognitive or algorithmic. The bare inference models suffer less from the curse of dimensionality since the effective database size is $S={\cal N}\Ns$ instead of $\Ns$.

\section{Bibliography}

\printbibliography[heading=none] 

\end{document}